\documentclass{article}

\PassOptionsToPackage{numbers, compress}{natbib}
\usepackage[final]{neurips_2023}

\usepackage[utf8]{inputenc} 
\usepackage[T1]{fontenc}    
\usepackage{hyperref}
\usepackage{url}     
\usepackage{booktabs}
\usepackage{amsfonts}
\usepackage{nicefrac}
\usepackage{microtype}      
\usepackage{xcolor}  
\usepackage{algorithm}
\usepackage{algorithmic}
\usepackage{bm}
\usepackage{makecell}
\usepackage{amsthm}
\usepackage{amssymb}
\usepackage{amsmath}
\usepackage{bbm}
\newtheorem{definition}{Definition}

\newtheorem{remark}{Remark}

\newtheorem{proposition}{Proposition}

\newtheorem{condition}{Condition}
\usepackage{multirow}

\usepackage{adjustbox}

\usepackage{wrapfig}
\usepackage{graphicx}
\usepackage{subfigure}
\usepackage{lipsum}

\usepackage{colortbl}
\usepackage{color}
\definecolor{greyC}{RGB}{180,180,180}
\definecolor{greyL}{RGB}{235,235,235}
\definecolor{citeColor}{RGB}{0,20,115}
\hypersetup{colorlinks,linkcolor={citeColor},citecolor={citeColor},urlcolor={citeColor}}

\title{Out-of-distribution Detection Learning with Unreliable Out-of-distribution Sources}

\author{%
Haotian Zheng$^{1,2}$\thanks{Equal contributions.} \quad Qizhou Wang$^{1}$\footnotemark[1] \quad Zhen Fang$^{3}$ \quad Xiaobo Xia$^{4}$ \\ \textbf{Feng Liu$^{5}$ \quad Tongliang Liu$^{4}$ \quad Bo Han$^{1}\thanks{Correspondence to Bo Han (bhanml@comp.hkbu.edu.hk).}$} \\
  $^1$Department of Computer Science, Hong Kong Baptist University \\
  $^2$School of Electronic Engineering, Xidian University \\
  $^3$Australian Artificial Intelligence Institute, University of Technology Sydney\\
  $^4$Sydney AI Centre, The University of Sydney \\
  $^5$School of Computing and Information Systems, The University of Melbourne \\
  \textnormal{htzheng.xdu@gmail.com} \quad
  \textnormal{\{csqzwang, bhanml\}@comp.hkbu.edu.hk} \\ 
  \textnormal{zhen.fang@uts.edu.au} \quad 
  \textnormal{xiaoboxia.uni@gmail.com}\\
  \textnormal{fengliu.ml@gmail.com} \quad
  \textnormal{tongliang.liu@sydney.edu.au}
}

\begin{document}

\maketitle

\begin{abstract}
Out-of-distribution (OOD) detection discerns OOD data where the predictor cannot make valid predictions as in-distribution (ID) data, thereby increasing the reliability of open-world classification. However, it is typically hard to collect real out-of-distribution (OOD) data for training a predictor capable of discerning ID and OOD patterns. This obstacle gives rise to \emph{data generation-based learning methods}, synthesizing OOD data via data generators for predictor training without requiring any real OOD data. 
Related methods typically pre-train a generator on ID data and adopt various selection procedures to find those data likely to be the OOD cases. However, generated data may still coincide with ID semantics, i.e., mistaken OOD generation remains, confusing the predictor between ID and OOD data. To this end, we suggest that generated data (with mistaken OOD generation) can be used to devise an \emph{auxiliary OOD detection task} to facilitate real OOD detection. Specifically, we can ensure that learning from such an auxiliary task is beneficial if the ID and the OOD parts have disjoint supports, with the help of a well-designed training procedure for the predictor. Accordingly, we propose a powerful data generation-based learning method named \emph{Auxiliary Task-based OOD Learning} (ATOL) that can relieve the mistaken OOD generation. We conduct extensive experiments under various OOD detection setups, demonstrating the effectiveness of our method against its advanced counterparts. The code is publicly available at: \url{https://github.com/tmlr-group/ATOL}.
\end{abstract}

\section{Introduction}
\label{sec: intro}
Deep learning in the open world should not only make accurate predictions for in-distribution (ID) data meanwhile should detect out-of-distribution (OOD) data whose semantics are different from ID cases~\cite{hendrycks2016baseline,sun2021react,huang2021universal,zhang2022kgtuner,zhou2023combating,zhang2023adaprop}. It drives recent studies in OOD detection~\cite{MingYL22,HuangX00G0L23,li2023VAE,wang2023robustness,zhang2023sio,HuangZX00000L23}, which is important for many safety-critical applications such as autonomous driving and medical analysis. Previous works have demonstrated that well-trained predictors can wrongly take many OOD data as ID cases~\cite{BendaleB16,hendrycks2016baseline,huang2021importance,nguyen2015deep}, motivating recent studies towards effective OOD detection.

In the literature, an effective learning scheme is to conduct model regularization with OOD data, making predictors achieve low-confidence predictions on such data~\cite{atom2020, Hendrycks2019oe,energy2020,ming2022poem, oecc2021, Yang21UDG, mixupoe2021}. Overall, they directly let the predictor learn to discern ID and OOD patterns, thus leading to improved performance in OOD detection. However, it is generally hard to access real OOD data~\cite{du2022vos,fang2022out,wang2023DAL,wang2023doe}, hindering the practical usage scenarios for such a promising learning scheme.

\begin{figure}
    \minipage{0.56\textwidth}
    \subfigure[Mistaken Semantics]{
    \label{fig: mistaken images}
    \includegraphics[width=0.45\linewidth, trim=0 0 0 0,clip]{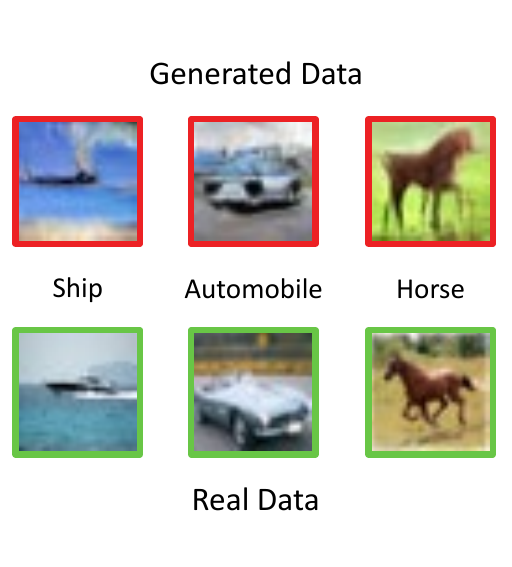}}
    \subfigure[Adverse Impacts]{
    \label{fig: mistaken impacts}
    \includegraphics[width=0.45\linewidth, trim=0 1 0 0,clip]{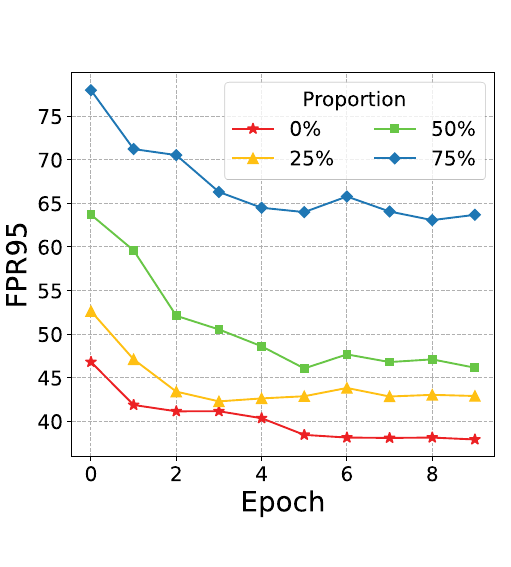}}
    \caption{Illustrations of mistaken OOD generation. Figure~\ref{fig: mistaken images} indicates the generated OOD data (colored in red) may wrongly possess semantics of the real ID data (colored in green), e.g., generated data with the ID semantics of ``Horse''. Figure~\ref{fig: mistaken impacts} demonstrates that the OOD performance of predictors are impaired by unreliable OOD data, described in Appendix~\ref{app: motivation exp}. With the proportion of OOD data possessing ID semantics increasing, the OOD performance (measured by FPR$95$) degrades.}
    \endminipage\hfill
    \minipage{0.38\textwidth}
    \includegraphics[width=0.9\linewidth, trim=0 -35 0 -10,clip]{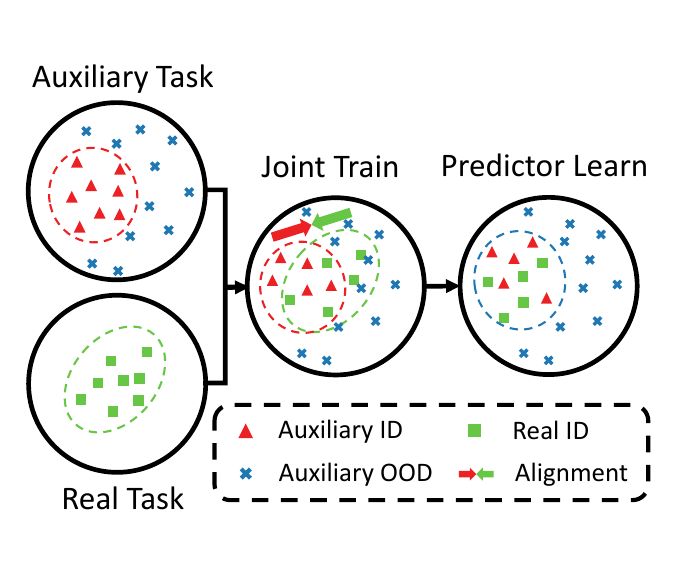}
    \caption{Illustration of our method. ATOL makes the predictor learn from the real task and our auxiliary task jointly via a well-designed learning procedure, leading to the separation between the OOD and ID cases. The predictor can benefit from such an auxiliary task for real OOD detection, relieving mistaken OOD generation.}\label{fig: ATOL}
    \endminipage\hfill
\end{figure}

Instead of collecting real OOD data, we can generate OOD data via data generators to benefit the predictor learning, motivating the {\emph{data generation-based methods}}~\cite{LeeLL2018, G2D2021, Sricharan18robust, Vernekar19via, CMG2022}. {Generally speaking}, existing works typically fit a data generator~\cite{Goodfellow2014gan} on ID data (since real OOD data are inaccessible). Then, these strategies target on finding the OOD-like cases from generated data by adopting various selection procedures. For instance, \citet{LeeLL2018} and \citet{Vernekar19via} take boundary data that lie far from ID boundaries as OOD-like data; \citet{G2D2021} select those data generated in the early stage of generator training. However, since the generators are trained on ID data and these selection procedures can make mistakes, one may wrongly select data with ID semantics as OOD cases (cf., Figure~\ref{fig: mistaken images}), i.e., \emph{mistaken OOD generation} remains. It will misguide the predictors to confuse between ID and OOD data, making the detection results unreliable, as in Figure~\ref{fig: mistaken impacts}. Overall, it is hard to devise generator training procedures or data selection strategies to overcome mistaken OOD generation, thus hindering practical usages of previous data generation-based methods.

To this end, we suggest that generated data can be used to devise an \emph{auxiliary task}, which can benefit real OOD detection even with data suffered from mistaken OOD generation. Specifically, such an auxiliary task is a crafted OOD detection task, thus containing ID and OOD parts of data (termed \emph{auxiliary ID} and \emph{auxiliary OOD data}, respectively). Then, two critical problems arise ---\emph{how to make such an auxiliary OOD detection task learnable w.r.t. the predictor} and \emph{how to ensure that the auxiliary OOD detection task is beneficial w.r.t. the real OOD detection}. For the first question, we refer to the advanced OOD detection theories~\cite{fang2022out}, suggesting that the auxiliary ID and the auxiliary OOD data should have the disjoint supports in the input space (i.e., without overlap w.r.t. ID and OOD distributions), cf., \textbf{C}\ref{con: supp} (Condition~\ref{con: supp} for short). For the second question, we justify that, for our predictor, if real and auxiliary ID data follow similar distributions, cf., \textbf{C}\ref{con: transfer}, the auxiliary OOD data are reliable OOD data w.r.t. the real ID data. In summary, if \textbf{C}\ref{con: supp} and \textbf{C}\ref{con: transfer} hold, we can ensure that learning from the auxiliary OOD detection task can benefit the real OOD detection, cf., Proposition~\ref{the: oe}.

Based on the auxiliary task, we propose \emph{Auxiliary Task-based OOD Learning} (ATOL), an effective data generation-based OOD learning method. For \textbf{C}\ref{con: supp}, it is generally hard to directly ensure the disjoint supports between the auxiliary ID and auxiliary OOD data due to the high-dimensional input space. Instead, we manually {craft} two disjoint regions in the low-dimensional latent space (i.e., the input space of the generator). Then, {with the generator having \emph{distance correlation} between input space and latent space}~\cite{wainwright2019high}, we can make the generated data that belong to two different regions in the latent space have the disjoint supports in the input space, assigning to the auxiliary ID and the auxiliary OOD parts, respectively. Furthermore, to fulfill \textbf{C}\ref{con: transfer}, we propose a distribution alignment risk for the auxiliary and the real ID data, alongside the OOD detection risk to make the predictor benefit from the auxiliary OOD detection task. Following the above learning procedure, we can relieve the mistaken OOD generation issue and achieve improved performance over previous works in data generation-based OOD detection, cf., Figure~\ref{fig: ATOL} for a heuristic illustration. 

To verify the effectiveness of ATOL, we conduct extensive experiments across representative OOD detection setups, demonstrating the superiority of our method over previous data generation-based methods, such as \citet{Vernekar19via}, with $12.02\%$, $18.32\%$, and $20.70\%$ improvements measured by average FPR$95$ on CIFAR-$10$, CIFAR-$100$, and ImageNet datasets. Moreover, compared with more advanced OOD detection methods (cf., Appendix~\ref{app: further}), such as \citet{knn2022}, our ATOL can also have promising improvements, which reduce the average FPR$95$ by $4.36\%$, $15.03\%$, and $12.02\%$ on CIFAR-$10$, CIFAR-$100$, and ImageNet datasets. 

{\textbf{Contributions. } We summarize our main contributions into three folds:
\begin{itemize}
    \item We focus on the mistaken OOD generation in data generation-based detection methods, which are largely overlooked in previous works. To this end, we introduce an auxiliary OOD detection task to combat the mistaken OOD generation and suggest a set of conditions in Section~\ref{sec: motivation} to make such an auxiliary task useful. 
    \item Our discussion about the auxiliary task leads to a practical learning method named ATOL. Over existing works in data generation-based methods, our ATOL introduces small costs of extra computations but is less susceptible to mistaken OOD generation. 
    \item We conduct extensive experiments across representative OOD detection setups, demonstrating the superiority of our method over previous data generation-based methods. Our ATOL is also competitive over advanced works that use real OOD data for training, such as outlier exposure~\cite{Hendrycks2019oe}, verifying that the data generation-based methods are still worth studying. 
\end{itemize}
}

\section{Preliminary}
Let $\mathcal{X}\subseteq\mathbb{R}^{n}$ the input space, $\mathcal{Y}=\{1, \ldots, c\}$ the label space, and $\mathbf{h}=\boldsymbol{\rho}\circ \boldsymbol{\phi}: \mathbb{R}^{n} \rightarrow \mathbb{R}^c$ the predictor, where $\boldsymbol\phi:\mathbb{R}^{n}\rightarrow\mathbb{R}^d$ is the feature extractor and $\boldsymbol\rho:\mathbb{R}^d\rightarrow\mathbb{R}^c$ is the classifier. We consider $\mathcal{P}^{\text{ID}}_{X,Y}$ and $\mathcal{P}^{\text{ID}}_{X}$ the joint and the marginal ID distribution defined over $\mathcal{X} \times \mathcal{Y}$  and $\mathcal{X}$, respectively. We also denote $\mathcal{P}^{\text{OOD}}_{X}$ the marginal OOD distribution over $\mathcal{X}$. Then, the main goal of OOD detection is to find the proper predictor $\mathbf{h}(\cdot)$ and the scoring function $s(\cdot ;\mathbf{h}):\mathbb{R}^{n}\rightarrow \mathbb{R}$ such that the OOD detector
\begin{equation}\label{Eq::OOD detector}
  f_{\beta}(\mathbf{x}) = \left \{
  \begin{aligned}
   &~~~~~~~~~{\rm ID},~~\textnormal{ if}~s(\mathbf{x};\mathbf{h})\geq \beta,\\ 
   & ~~~~ {\rm OOD},~~\textnormal{ if}~s(\mathbf{x};\mathbf{h})< \beta,
    \end{aligned}
    \right.
\end{equation}
can detect OOD data with a high success rate, where $\beta$ is a given threshold. To get a proper predictor $\mathbf{h}(\cdot)$, outlier exposure~\cite{Hendrycks2019oe} proposes regularizing the predictor to produce low-confidence predictions for OOD data. {Assuming the real ID data $\mathbf{x}_\text{ID}$ and label $y_\text{ID}$ are randomly drawn from $\mathcal{P}^{\text{ID}}_{X,Y}$,} then the learning objective of outlier exposure can be written as:
\begin{equation}
    \min_{\mathbf{h}}\mathbb{E}_{\mathcal{P}^{\text{ID}}_{X,Y}}\left[\ell_\text{CE}(\mathbf{x}_{\text{ID}},{y}_{\text{ID}};\mathbf{h})\right] + \lambda \mathbb{E}_{\mathcal{P}^{\text{OOD}}_{X}}\left[\ell_\text{OE}(\mathbf{x};\mathbf{h})\right], \label{eq: oe}
\end{equation}
where $\lambda$ is the trade-off hyper-parameter, $\ell_\text{CE}(\cdot)$ is the cross-entropy loss, and $\ell_\text{OE}(\cdot)$  is the Kullback-Leibler divergence of softmax predictions to the uniform distribution. Although outlier exposure remains one of the most powerful methods in OOD detection, its critical reliance on OOD data hinders its practical applications.

Fortunately, data generation-based OOD detection methods can overcome the reliance on real OOD data, meanwhile making the predictor learn to discern ID and OOD data. Overall, existing works ~\cite{LeeLL2018, Vernekar19via, CMG2022} leverage the \emph{generative adversarial network} (GAN)~\cite{Goodfellow2014gan, DCGAN15} to generate data, collecting those data that are likely to be the OOD cases for predictor training. As a milestone work, \citet{LeeLL2018} propose selecting the ``boundary'' data that lie in the low-density area of $\mathcal{P}^{\text{ID}}_{X}$, with the following learning objective for OOD generation:
\begin{equation}
    \min_G \max_D~~ \underbrace{\mathbb{E}_{\mathcal{G}_{X}}[\ell_{\text{OE}}(\mathbf{x};\mathbf{h})]}_{\text{(a) generating~boundary~data}} 
      + \underbrace{\mathbb{E}_{\mathcal{P}^{\text{ID}}_{X}}[\log D(\mathbf{x}_{\text{ID}})] + \mathbb{E}_{\mathcal{G}_{X}}[\log(1-D(\mathbf{x}))]}_{\text{(b) training~data~generator}},
\end{equation}
where $\mathcal{G}_{X}$ denotes the OOD distribution of the generated data w.r.t. the generator $G:\mathbb{R}^m\rightarrow\mathbb{R}^n$ ($m$ is the dimension of the {latent space}). Note that the term (a) forces the generator $G(\cdot)$ to generate low-density data and the term (b) is the standard learning objective for GAN training~\cite{Goodfellow2014gan}. Then, with the generator trained for OOD generation, the generated data are used for predictor training, following the outlier exposure objective, given by
\begin{equation}
    \label{eq: generation}
    \min_{\mathbf{h}}\mathbb{E}_{\mathcal{P}_{X,Y}^\text{ID}}\left[\ell_\text{CE}(\mathbf{x}_{\text{ID}},y_{\text{ID}};\mathbf{h})\right] + \lambda \mathbb{E}_{\mathcal{G}_{X}}\left[\ell_\text{OE}(\mathbf{x};\mathbf{h})\right].
\end{equation}

\textbf{Drawbacks.} Although a promising line of works, the generated data therein may still contain many ID semantics (cf., Figure~\ref{fig: mistaken images}). It stems from the lack of knowledge about OOD data, where one can only access ID data to guide the training of generators. Such mistaken OOD generation is common in practice, and its negative impacts are inevitable in misleading the predictor (cf., Figure~\ref{fig: mistaken impacts}). Therefore, {how to overcome the mistaken OOD generation} is one of the key challenges in data generation-based OOD detection methods. 

\section{Motivation}
\label{sec: motivation}

Note that getting generators to generate reliable OOD data is difficult, since there is no access to real OOD data in advance and the selection procedures can make mistakes. Instead, given that generators suffer from mistaken OOD generation, we aim to make the predictors alleviate the negative impacts raised by those unreliable OOD data. 

Our key insight is that these unreliable generated data can be used to devise a reliable \emph{auxiliary task} which can benefit the predictor in  real OOD detection. In general, such an auxiliary task is a crafted OOD detection task, of which the generated data therein are separated into the ID and the OOD parts. Ideally, the predictor can learn from such an auxiliary task and \emph{transfer} the learned knowledge (from the auxiliary task) to benefit the real OOD detection, thereafter relieving the mistaken OOD generation issue. Therein, two issues require our further study:
\begin{itemize}
    \item Based on unreliable OOD sources, how to devise a learnable auxiliary OOD detection task?

    \item Given the well-designed auxiliary task, how to make it benefit the real OOD detection?
\end{itemize}

For the first question, we refer the advanced work~\cite{fang2022out} in OOD detection theories, which provides necessary conditions for which the OOD detection task is learnable. In the context of the auxiliary OOD detection task, we will separate the generated data into the \emph{auxiliary ID} parts and \emph{auxiliary OOD} parts, following the distributions defined by $\mathcal{G}^{\text{ID}}_{X}$ and $\mathcal{G}^{\text{OOD}}_{X}$, respectively.  Then, we provide the following condition for the \emph{separable property} of the auxiliary OOD detection task. 
\begin{condition}
    [\textbf{Separation for the Auxiliary Task}] \label{con: supp}
    There is no overlap between the auxiliary ID 
    distribution and the auxiliary OOD distribution, i.e., $\texttt{supp}~\mathcal{G}^{\text{ID}}_{X} \cap \texttt{supp}~\mathcal{G}^{\text{OOD}}_{X} = \emptyset$, where \texttt{supp} denotes the support set of a distribution.
\end{condition}
Overall, \textbf{C}\ref{con: supp} states that the auxiliary ID data distribution $\mathcal{G}^{\text{ID}}_{X}$ and the auxiliary OOD data distribution $\mathcal{G}^{\text{OOD}}_{X}$ should have the disjoint supports. Then, according to \citet{fang2022out} (Theorem 3), such an auxiliary OOD detection might be learnable, in that the predictor can have a high success rate in discerning the auxiliary ID and the auxiliary OOD data. In general, the separation condition is indispensable for us to craft a valid OOD detection task. Also, as in the proof of Proposition~\ref{the: oe}, the separation condition is also important to benefit the real OOD detection.  

For the second question, we need to first formally define if a data point can benefit the real OOD detection. Here, we generalize the separation condition in \citet{fang2022out}, leading to the following definition for the \emph{reliability} of OOD data. 
\begin{definition}[\textbf{Reliability of OOD Data}] \label{def: reliable OOD}
    An OOD data point $\mathbf{x}$ is reliable (w.r.t. real ID data) to a mapping function $\boldsymbol{\phi}'(\cdot)$, if $\boldsymbol{\phi}'(\mathbf{x})\notin\texttt{supp}\boldsymbol{\phi}'_{\#}\mathcal{P}_{X}^{\text{ID}}$, where $\boldsymbol{\phi}'_{\#}$ is the distribution transformation associated with $\boldsymbol{\phi}'(\cdot)$.
\end{definition}

With identity mapping function $\boldsymbol{\phi}'(\cdot)$, the above definition states that $\mathbf{x}$ is a reliable OOD data point if it is not in the support set of the real ID distribution w.r.t. the transformed space. Intuitively, if real ID data are representative almost surely~\cite{fang2022out}, $\boldsymbol{\phi}'(\cdot)$ far from the ID support also indicates that $\boldsymbol{\phi}'(\cdot)$ will not possess ID semantics, namely, reliable. Furthermore, $\boldsymbol{\phi}'(\cdot)$ is not limited to identity mapping, which can be defined by any transformed space. The feasibility of such a generalized definition is supported by many previous works~\cite{du2022vos,mehra2022certifying} in OOD learning, demonstrating that reliable OOD data defined in the embedding space (or in our context, the transformed distribution) can also benefit the predictor learning in OOD detection. 

Based on Definition~\ref{def: reliable OOD}, we want to make auxiliary OOD data reliable w.r.t. real ID data, such that the auxiliary OOD detection task is ensured to be beneficial, which motivates the following condition. 
\begin{condition}[\textbf{Transferability of the Auxiliary Task}] \label{con: transfer}
    The auxiliary OOD detection task is transferable w.r.t. the real OOD detection task if $\boldsymbol{\phi}'_{\#}\mathcal{P}_{X}^{\text{ID}}\approx \boldsymbol{\phi}'_{\#}\mathcal{G}^{\text{ID}}_{X}$.
\end{condition}
{Overall, \textbf{C}\ref{con: transfer} states that auxiliary ID data approximately follow the same distribution as that of the real ID data in the transformed space, i.e., $\boldsymbol{\phi}'(\mathcal{X})$. Note that, Auxiliary ID and OOD data can arbitrarily differ from real ID and OOD data in the data space, i.e., \textit{auxiliary data are unreliable OOD sources}. However, if \textbf{C}\ref{con: transfer} is satisfied, the model "\textit{believes}" the auxiliary ID data and the real ID data are the same.} Then, we are ready to present our main proposition, demonstrating why \textbf{C}\ref{con: transfer} can ensure the reliability of auxiliary OOD data.

\begin{proposition}
    \label{the: oe}
    Assume the predictor $\mathbf{h}=\boldsymbol{\rho} \circ \boldsymbol{\phi}$ can separate ID and OOD data, namely, 
    \begin{equation}
    \label{eq: predictor theorem}
 \mathbb{E}_{\mathcal{G}^{\text{ID}}_{X,Y}}[\ell_{\text{CE}}(\mathbf{x}_\text{ID},y_{\text{ID}};\mathbf{h})] + \lambda \mathbb{E}_{\mathcal{G}^{\text{OOD}}_X}[\ell_\text{OE}(\mathbf{x};\mathbf{h})]
    \end{equation}
    approaches 0. Under \textbf{C}\ref{con: supp} and \textbf{C}\ref{con: transfer},  auxiliary OOD data are reliable w.r.t. real ID data given $\boldsymbol{\phi}'=\boldsymbol{\phi}$.
\end{proposition}
The proof can be found in Appendix~\ref{app: proof}. In heuristics, Eq.~\eqref{eq: predictor theorem} states that the predictor should excel at such an auxiliary OOD detection task, where the model can almost surely discern the auxiliary ID and the auxiliary OOD cases. Then, if we can further align the real and the auxiliary ID distributions (i.e., \textbf{C}\ref{con: transfer}), the predictor will make no difference between the real ID data and the auxiliary ID data. Accordingly, since the auxiliary OOD data are reliable OOD cases w.r.t. the auxiliary ID data, the auxiliary OOD data are also reliable w.r.t. the real ID data. Thereafter, one can make the predictor learn to discern the real ID and auxiliary OOD data, where no mistaken OOD generation affects. Figure~\ref{fig: ATOL} summarizes our key concepts. As we can see, learning from the auxiliary OOD detection task and further aligning the ID distribution ensure the separation between ID and OOD cases, benefiting the predictor learning from reliable OOD data and improving real OOD detection performance.

\begin{remark}
There are two parallel definitions for real OOD data~\cite{wang2023doe}, related to different goals towards effective OOD detection. One definition is that an exact distribution of real OOD data exists, and the goal is to mitigate the distribution discrepancy w.r.t. the real OOD distribution for effective OOD detection. Another definition is that all those data whose true labels are not in the considered label space are real OOD data, and the goal is to make the predictor learn to see as many OOD cases as possible. Due to the lack of real OOD data for supervision, the latter definition is more suitable than the former one in our context, where we can concentrate on mistaken OOD generation. 
\end{remark}

\section{Learning Method}

Section~\ref{sec: motivation} motivates a new data generation-based method named \emph{Auxiliary Task-based OOD Learning} (ATOL), overcoming the mistaken OOD generation via the auxiliary OOD detection task. Referring to Theorem~\ref{the: oe}, the power of the auxiliary task is built upon \textbf{C}\ref{con: supp} and \textbf{C}\ref{con: transfer}. It makes ATOL a two-staged learning scheme, related to auxiliary task crafting (crafting the auxiliary task for \textbf{C}\ref{con: supp}) and predictor training (applying the auxiliary task for \textbf{C}\ref{con: transfer}), respectively. Here, we provide the detailed discussion.

\subsection{Crafting the Auxiliary Task} 
\label{sec: craft}
We first need to construct the auxiliary OOD detection task, defined by auxiliary ID and auxiliary OOD data generated by the generator $G(\cdot)$. In the following, we assume the auxiliary ID data $\hat{\mathbf{x}}_\text{ID}$ drawn from $\mathcal{G}^{\text{ID}}_{X}$ and the auxiliary OOD data $\hat{\mathbf{x}}_\text{OOD}$ drawn from $\mathcal{G}^{\text{OOD}}_{X}$. Then, according to \textbf{C}\ref{con: supp}, we need to fulfill the condition that $\mathcal{G}^{\text{ID}}_{X}$ and $\mathcal{G}^{\text{OOD}}_{X}$ should have the disjoint supports. 

Although all distribution pairs $(\mathcal{G}^{\text{ID}}_{X}, \mathcal{G}^{\text{OOD}}_{X})$ with disjoint supports can make \textbf{C}\ref{con: supp} hold, it is hard to realize such a goal due to the complex data space $\mathbb{R}^n$ of images. Instead, we suggest crafting such a distribution pair in the latent space $\mathbb{R}^m$ of the generator, of which the dimension $m$ is much lower than $n$. We define the latent ID and the latent OOD data as $\mathbf{z}_\text{ID}$ and $\mathbf{z}_\text{OOD}$, respectively. Then, we assume the latent ID data are drawn from the high-density region of a Mixture of Gaussian (MoG):
\begin{equation}
\label{eq: latent ID}
    \mathbf{z}_\text{ID}\in \mathcal{Z}^\text{ID}~\text{with}~\mathcal{Z}^\text{ID}=\{\mathbf{z}\sim\mathcal{M}_Z|\mathcal{M}(\mathbf{z})>\tau\},
\end{equation}
where $\tau$ is the threshold and $\mathcal{M}_Z$ is a pre-defined MoG, {given by the density function of MoG $\mathcal{M}(\mathbf{z}) = \sum_{i=1}^c \frac{1}{c} \mathcal{N}(\mathbf{z}|\boldsymbol{\mu}_i, \boldsymbol{\sigma}_i)$, with the mean $\boldsymbol{\mu}_i$ and the covariance $\boldsymbol{\sigma}_i$ for the $i$-th sub-Gaussian}. Furthermore, as we demonstrate in Section~\ref{sec: apply}, we also require the specification of labels for latent ID data. Therefore, we assume that $\mathcal{M}_Z$ is constructed by $c$ sub-distributions of Gaussian, and data belong to each of these $c$ sub-distributions should have the same label $\hat{y}_\text{ID}$. 

\begin{remark}
    {MoG provides a simple way to generate data, yet the key point is to ensure that auxiliary ID/OOD data should have the disjoint support, i.e, \textbf{C}\ref{con: supp}. Therefore, if \textbf{C}\ref{con: supp} is satisfied properly, other noise distributions, such as the beta mixture models\cite{ma2009beta} and the uniform distribution, can also be used. Further, our ATOL is different from previous data generation-based methods in that we do not require generated data to be reliable in the data space. ATOL does not involve fitting the MoG to real ID data, where the parameters can be pre-defined and fixed. Therefore, we do not consider overfitting and accuracy of MoG in our paper.}
\end{remark}

Furthermore, for the disjoint support, the latent OOD data are drawn from a uniform distribution except for the region with high MoG density, i.e.,
\begin{equation}
\label{eq: latent OOD}
    \mathbf{z}_\text{OOD}\in \mathcal{Z}^\text{OOD}~\text{with}~\mathcal{Z}^\text{OOD}=\{\mathbf{z}\sim\mathcal{U}_Z|\mathcal{M}(\mathbf{z})\le\tau\},
\end{equation}
with $\mathcal{U}_Z$ a pre-defined uniform distribution. For now, we can ensure that $\mathcal{Z}^\text{ID}\cap\mathcal{Z}^\text{OOD}=\emptyset$.

However, $\mathcal{Z}^\text{ID}\cap\mathcal{Z}^\text{OOD}=\emptyset$ does not imply $G(\mathcal{Z}^\text{ID})\cap G(\mathcal{Z}^\text{OOD})=\emptyset$, i.e., \textbf{C}\ref{con: supp} is not satisfied given arbitrary $G(\cdot)$. To this end, we suggest further ensuring the generator to be a distance-preserving function~\cite{Distance22}, which is a sufficient condition to ensure that the disjoint property can be transformed from the latent space into the data space. For distance preservation, we suggest a regularization term for the generator regularizing, which is given by
\begin{equation}
\label{eq: map}
\ell_{\text{reg}}(\mathbf{z_1}, \mathbf{z_2};G) = -\frac{\mathbb{E}_{{\mathcal{U}_Z}\times{\mathcal{U}_Z}}\left[d(\mathbf{z}_1,\mathbf{z}_2) \cdot d(G(\mathbf{z}_1),G(\mathbf{z}_2))\right]}{\sqrt{\mathbb{E}_{{\mathcal{U}_Z}\times{\mathcal{U}_Z}}\left[d(\mathbf{z}_1,\mathbf{z}_2)\right] \mathbb{E}_{{\mathcal{U}_Z}\times{\mathcal{U}_Z}}\left[d(G(\mathbf{z}_1),G(\mathbf{z}_2))\right]}},
\end{equation}
where $d(\mathbf{z}_1,\mathbf{z}_2)$ is the centralized distance given by $\left \| \mathbf{z}_1 - \mathbf{z}_2\right \|_2-\mathbb{E} \left \| \mathbf{z}_1 - \mathbf{z}_2\right \|_2$. Overall, Eq.~\eqref{eq: map} is the correlation for the distances between data pairs measured in the latent and the data space, stating that the closer distances in the latent space should indicate the closer distances in the data space. 

In summary, to fulfill \textbf{C}\ref{con: supp}, we need to specify the regularizing procedure for the generator, overall summarized by the following two steps. First, we regularize the generator with the regularization in Eq.~\eqref{eq: map} to ensure the distance-preserving property. Then, we specify two disjoint regions in the latent space following Eqs.~\eqref{eq: latent ID}-\eqref{eq: latent OOD}, of which the data after passing through the generator are taken as auxiliary ID and auxiliary OOD data, respectively.  We also summarize the algorithm details in crafting the auxiliary OOD detection task in Appendix~\ref{app: overall alg}. 

\subsection{Applying the Auxiliary Task} 
\label{sec: apply}

We have discussed a general strategy to construct the auxiliary OOD detection task, which can be learned by the predictor following an outlier exposure-based learning objective, similar to Eq.~\eqref{eq: oe}. Now, we fulfill \textbf{C}\ref{con: transfer}, transferring model capability from the auxiliary task to real OOD detection. 

In general, \textbf{C}\ref{con: transfer} states that we should align distributions between the auxiliary and the real ID data such that the auxiliary OOD data can benefit real OOD detection, thus overcoming the negative effects of mistaken OOD generation. Following \citet{VHL2022}, we align the auxiliary and the real ID distribution via supervised contrastive learning~\cite{supcon20}, pulling together data belonging to the same class in the embedding space meanwhile pushing apart data from different classes. Accordingly, we suggest the alignment loss for the auxiliary ID data $\hat{\mathbf{x}}_{\text{ID}}$, following,
\begin{equation}
    \ell_{\text{align}}(\hat{\mathbf{x}}_{\text{ID}}, \hat{y}_{\text{ID}}; \boldsymbol{\phi})=  -\log \frac{1}{|\mathcal{N}_l^{\hat{y}_{\text{ID}}}|} \sum_{\mathbf{x}^p_{\text{ID}}\in \mathcal{N}_l^{\hat{y}_{\text{ID}}}}^{} \frac{\exp[\boldsymbol{\phi}(\hat{\mathbf{x}}_{\text{ID}})\cdot \boldsymbol{\phi}({\mathbf{x}^p_{\text{ID}}})]}{\sum_{\mathbf{x}^a_{\text{ID}}\in \mathcal{N}_l}^{}  \exp[\boldsymbol{\phi}(\hat{\mathbf{x}}_{\text{ID}})\cdot \boldsymbol{\phi}(\mathbf{x}^a_{\text{ID}})]}.
\end{equation}
$\mathcal{N}_l$ is a set of data of size $l$ drawn from the real ID distribution, namely, $\mathcal{N}_l=\{\mathbf{x}^i_\text{ID}|(\mathbf{x}^i_\text{ID}, y^i_\text{ID})\sim\mathcal{P}_{X,Y}^\text{ID},~\text{for}~i\in\{1,\dots,l\}\}$. Furthermore, $\mathcal{N}_l^y$ is a subset of $\mathcal{N}_l$ with ID data that below to the label of $y$, namely,  $\mathcal{N}_l^y=\{\mathbf{x}^i_\text{ID}|(\mathbf{x}^i_\text{ID}, y^i_\text{ID})\sim\mathcal{P}_{X,Y}^\text{ID}~\text{, and}~y^i_\text{ID}=\hat{y}_{\text{ID}},~\text{for}~i\in\{1,\dots,l\}\}$. 
Accordingly, based on Theorem~\ref{the: oe}, we can ensure that learning from the auxiliary OOD detection task, i.e., small Eq.~\eqref{eq: predictor theorem}, can benefit real OOD detection, freeing from the mistaken OOD generation.

\subsection{Overall Algorithm}

We further summarize the training and the inferring procedure for the predictor of ATOL, given the crafted auxiliary ID and the auxiliary OOD distributions that satisfy \textbf{C}\ref{con: supp} via Eqs.~\eqref{eq: latent ID}-\eqref{eq: latent OOD}. The pseudo codes are further summarized in Appendix~\ref{app: overall alg} due to the space limit. 

\textbf{Training Procedure.} The overall learning objective consists of the real task learning, the auxiliary task learning, and the ID distribution alignment, which is given by:
\begin{equation}
    \begin{aligned}
    \label{eq: overall}
    \min_{\mathbf{h}=\boldsymbol{\rho}\circ\boldsymbol{\phi}}~& \overbrace{\mathbb{E}_{\mathcal{P}^{\text{ID}}_{X,Y}} \left [\ell_{\text{CE}}(\mathbf{x}_{\text{ID}},y_{\text{ID}}; \mathbf{h}) \right]}^{\text{(a) real task learning}}+ \overbrace{{\mathbb{E}_{\mathcal{G}^{\text{ID}}_{X,Y}} \left [\ell_{\text{CE}}(\hat{\mathbf{x}}_{\text{ID}},\hat{y}_{\text{ID}}; \mathbf{h}) \right]} + \lambda  \mathbb{E}_{\mathcal{G}^{\text{OOD}}_{X}} \left [\ell_{\text{OE}}(\hat{\mathbf{x}}_{\text{OOD}}; \mathbf{h}) \right]}^{\text{(b) auxiliary task learning}} + \\ &  \underbrace{\alpha \mathbb{E}_{\mathcal{G}^{\text{ID}}_{X,Y}} \left [\ell_{\text{align}}(\hat{\mathbf{x}}_{\text{ID}},\hat{y}_{\text{ID}};\boldsymbol{\phi}) \right]}_{\text{(c) ID distribution alignment}},
    \end{aligned}
\end{equation}
where $\alpha, \lambda \ge 0$ are the trade-off parameters. By finding the predictor that leads to the minimum of the Eq.~\eqref{eq: overall}, our ATOL can ensure the improved performance of real OOD detection. Note that, Eq.~\eqref{eq: overall} can be realized in a stochastic manner, suitable for deep model training.

\textbf{Inferring Procedure.} We adopt the MaxLogit scoring~\cite{Hendrycks22maxlogit} in OOD detection. Given a test input $\mathbf{x}$, the MaxLogit score is given by:
\begin{equation}
    s_{\text{ML}}(\mathbf{x};\mathbf{h})=\max_k \mathbf{h}_k(\mathbf{x}),
\end{equation}
where $\mathbf{h}_k(\cdot)$ denotes the $k$-th logit output. In general, the MaxLogit scoring is better than other commonly-used scoring functions, such as maximum softmax prediction~\cite{hendrycks2016baseline}, when facing large semantic space. Therefore, we choose MaxLogit instead of MSP for OOD scoring in our realization. 
\vspace{-5pt}

\section{Experiments}
This section conducts extensive experiments for ATOL in OOD detection. In Section~\ref{sec: setup}, we describe the experiment setup. In Section~\ref{sec: main results}, we demonstrate the main results of our method against the data generation-based counterparts on both the CIFAR~\cite{krizhevsky2009learning} and the ImageNet~\cite{deng2009imagenet} benchmarks. In Section~\ref{sec: ablation}, we further conduct ablation studies to comprehensively analyze our method.

\subsection{Setup}
\label{sec: setup}
\textbf{Backbones setups. } {For the CIFAR benchmarks, we employ the WRN-40-2~\cite{zagoruyko2016wide} as the backbone model. Following~\cite{energy2020}, models have been trained for $200$ epochs via empirical risk minimization, with a batch size $64$, momentum $0.9$, and initial learning rate $0.1$. The learning rate is divided by $10$ after $100$ and $150$ epochs. 
For the ImageNet, we employ pre-trained ResNet-50~\cite{he2016deep} on ImageNet, downloaded from the PyTorch official repository.}

\textbf{Generators setups. } For the CIFAR benchmarks, following~\cite{LeeLL2018}, we adopt the generator from the Deep Convolutional (DC) GAN~\cite{DCGAN15}). Following the vanilla generation objective, DCGAN has been trained on the ID data, where the batch size is $64$ and the initial learning rate is $0.0002$. For the ImageNet benchmark, we adopt the generator from the {BigGAN}~\cite{biggan19} model, designed for scaling generation of high-resolution images, where the pre-trained model {biggan-256} can be downloaded from the TensorFlow hub. Note that our method can work on various generators except those mentioned above, even with a random-parameterized one, later shown in Section~\ref{sec: ablation}.

\textbf{Baseline methods. } We compare our ATOL with advanced data generation-based methods in OOD detection, including (1) BoundaryGAN~\cite{LeeLL2018}, (2) ConfGAN~\cite{Sricharan18robust}, (3) ManifoldGAN~\cite{Vernekar19via}, (4) G2D~\cite{G2D2021} (5) CMG~\cite{CMG2022}. For a fair comparison, all the methods use the same generator and pre-trained backbone without regularizing with outlier data. 

\textbf{Auxiliary task setups. } {Hyper-parameters are chosen based on the OOD detection performance on validation datasets}. In latent space, the auxiliary ID distribution is the high density region of the Mixture of Gaussian (MoG). Each sub-Gaussian in the MoG has the mean $\boldsymbol{\mu}=(\mu_1, \mu_2,\dots, \mu_m)$ and the same covariance matrix $\sigma \cdot \boldsymbol{I}$, where $m$ is the dimension of latent space, $\mu_i$ is randomly selected from the set $\{-\mu, \mu\}$, and $\boldsymbol{I}$ is the identity matrix. The auxiliary OOD distribution is the uniform distribution except for high MoG density region, where each dimension has the same space size $u$. For the CIFAR benchmarks, the value of $\alpha$ is set to $1$, $\mu$ is $5$, $\sigma$ is $0.1$, and $u$ is $8$. For the ImageNet benchmarks, the value of $\alpha$ is set to $1$, $\mu$ is $5$, $\sigma$ is $0.8$, and $u$ is $8$.

\textbf{Training details. } For the CIFAR benchmarks, ATOL is run for $10$ epochs and uses SGD with an initial learning rate $0.01$ and the cosine decay~\cite{LoshchilovH17}. The batch size is $64$ for real ID cases, $64$ for auxiliary ID cases, and $256$ for auxiliary OOD cases. For the ImageNet benchmarks, ATOL is run for $10$ epochs using SGD with an initial learning rate $0.0003$ and a momentum $0.9$. The learning rate is decayed by a factor of $10$ at $30\%, 60\%$, and $90\%$ of the training steps. Furthermore, the batch size is fixed to $32$ for real ID cases, $32$ for auxiliary ID cases, and $128$ for auxiliary OOD cases.

\textbf{Evaluation metrics. } The OOD detection performance of a detection model is evaluated via two representative metrics, which are both threshold-independent~\cite{DavisG06}: the false positive rate of OOD data when the true positive rate of ID data is at $95\%$ (FPR$95$); and the \emph{area under the receiver operating characteristic curve} (AUROC), which can be viewed as the probability of the ID case having greater score than that of the OOD case.

\subsection{Main Results}
\label{sec: main results}

We begin with our main experiments on the CIFAR and ImageNet benchmarks. Model performance is tested on commonly-used OOD datasets. For the CIFAR cases, we employed Texture~\cite{cimpoi2014describing}, SVHN~\cite{netzer2011reading}, Places$365$~\cite{ZhouLKO018}, LSUN-Crop~\cite{yu2015lsun}, LSUN-Resize~\cite{yu2015lsun}, and iSUN~\cite{xu2015turkergaze}. For the ImageNet case, we employed iNaturalist~\cite{HornASCSSAPB18}, SUN~\cite{xu2015turkergaze}, Places$365$~\cite{ZhouLKO018}, and Texture~\cite{cimpoi2014describing}. In Table~\ref{tab: cifar&imagenet}, we report the average performance (i.e., FPR$95$ and AUROC) regarding the OOD datasets mentioned above. Please refer to Tables~\ref{tab: cifar10 full}-\ref{tab: cifar100 full} and \ref{tab: imagenet full} in Appendix~\ref{app: full exp} for the detailed reports. 

\textbf{CIFAR benchmarks. } Overall, our ATOL can lead to effective OOD detection, which generally demonstrates better results than the rivals by a large margin. In particular, compared with the advanced data generation-based method, e.g., ManifoldGAN~\cite{Vernekar19via}, ATOL significantly improves the performance in OOD detection, revealing $12.02\%$ and $2.96\%$ average improvements w.r.t. FPR$95$ and AUROC on the CIFAR-$10$ dataset and $18.32\%$ and $9.84\%$ of the average improvements on the CIFAR-$100$ dataset. This highlights the superiority of our novel data generation strategy, relieving mistaken OOD generation as in previous methods.

\textbf{ImageNet benchmark. } We evaluate a large-scale OOD detection task based on ImageNet dataset. Compared to the CIFAR benchmarks above, the ImageNet task is more challenging due to the large semantic space and large amount of training data. As we can see, previous data generation-based methods reveal poor performance due to the increased difficulty in searching for the proper OOD-like data in ImageNet, indicating the critical mistaken OOD generation remains (cf., Appendix~\ref{app: generated images}). In contrast, our ATOL can alleviate this issue even for large-scale datasets, e.g., ImageNet, thus leading to superior results. In particular, our ATOL outperforms the best baseline ManifoldGAN~\cite{Vernekar19via} by $20.70\%$ and $8.09\%$ average improvement w.r.t. FPR$95$ and AUROC, which demonstrates the advantages of our ATOL to relieve mistaken OOD generation and benefit to real OOD detection.

\begin{table}[t]
\caption{Comparison in OOD detection on the CIFAR and ImageNet benchmarks. $\downarrow$ (or $\uparrow$) indicates smaller (or larger) values are preferred; a bold font indicates the best results in a column.} \label{tab: cifar&imagenet}
\centering
\small
\begin{tabular}{c|cc|cc|cc}
\specialrule{0em}{1pt}{1pt}
\toprule[1.5pt]
\multicolumn{1}{c|}{\multirow{2}{*}{Methods}} & \multicolumn{2}{c|}{CIFAR-$10$} & \multicolumn{2}{c|}{CIFAR-$100$} & \multicolumn{2}{c}{ImageNet} \\ \specialrule{0em}{1pt}{0pt} \cline{2-7} \specialrule{0em}{2pt}{0pt}
\multicolumn{1}{c|}{} & FPR$95$~$\downarrow$ & \multicolumn{1}{c|}{AUROC~$\uparrow$} & FPR$95$~$\downarrow$ & \multicolumn{1}{c|}{AUROC~$\uparrow$} & FPR$95$~$\downarrow$ & AUROC~$\uparrow$ \\ \midrule[1pt]
\multicolumn{1}{c|}{BoundaryGAN~\cite{LeeLL2018}} & 55.60& \multicolumn{1}{c|}{86.46} & 76.72& \multicolumn{1}{c|}{75.79} & 85.48& 66.78 \\
\multicolumn{1}{c|}{ConfGAN~\cite{Sricharan18robust}} & 31.57& \multicolumn{1}{c|}{93.01} & 74.86& \multicolumn{1}{c|}{77.67} & 74.88& 77.03 \\
\multicolumn{1}{c|}{ManifoldGAN~\cite{Vernekar19via}}& 26.68& \multicolumn{1}{c|}{94.09} & 73.54& \multicolumn{1}{c|}{77.40} & 72.50& 77.73 \\
\multicolumn{1}{c|}{G2D~\cite{G2D2021}}   & 31.83& \multicolumn{1}{c|}{91.74} & 70.73& \multicolumn{1}{c|}{79.03} & 74.93& 77.16 \\ 
\multicolumn{1}{c|}{CMG~\cite{CMG2022}} & 39.83& \multicolumn{1}{c|}{92.83} & 79.60& \multicolumn{1}{c|}{77.51} & 72.95& 77.63 \\
\midrule[0.6pt]
\multicolumn{1}{c|}{\textbf{ATOL} (ours)} & \textbf{14.66}& \multicolumn{1}{c|}{\textbf{97.05}} & \textbf{55.22}& \multicolumn{1}{c|}{\textbf{87.24}} & \textbf{51.80}& \textbf{85.82} \\ 

\bottomrule[1.5pt]
\end{tabular}
\end{table}

\subsection{Ablation Study}
\label{sec: ablation}

To further demonstrate the effectiveness of our ATOL, we conduct extensive ablation studies to verify the contributions of each proposed component.

\textbf{Auxiliary task crafting schemes. } In Section~\ref{sec: craft}, we introduce the realization of the auxiliary OOD detection task in a tractable way. Here, we study two components of generation: the disjoint supports for ID and OOD distributions in the latent space and the distance-preserving generator. In particular, we compare ATOL with two variants: 1) {ATOL} $w/o$ $\tau$, where the ID and OOD distributions may overlap in latent space without separation via the threshold $\tau$; 2) {ATOL} $w/o$ $\ell_{\text{reg}}$, where we directly use the generator to generate the auxiliary data without further regularization. As we can see in Table~\ref{tab: ablation}, the overlap between the auxiliary ID and the auxiliary OOD distributions in latent space will lead to the failure of the auxiliary OOD detection task, demonstrating the catastrophic performance (e.g., $92.20\%$ FPR$95$ and $51.34\%$ AUROC on CIFAR-$100$). Moreover, without the regularized generator as a distance-preserving function, the ambiguity of generated data will compromise the performance of ATOL, which verifies the effectiveness of our generation scheme. 

\textbf{Auxiliary task learning schemes. }
In Section~\ref{sec: apply}, we propose ID distribution alignment to transfer model capability from auxiliary OOD detection task to real OOD detection. In Table~\ref{tab: ablation}, we conduct experiments on a variant, namely, {ATOL} $w/o$ $\ell_{\text{align}}$, where the predictor directly learns from the auxiliary task without further alignment. Accordingly, {ATOL} $w/o$ $\ell_{\text{align}}$ can reveal better results than {ATOL} $w/o$ $\ell_{\text{reg}}$, with $18.84\%$ and $3.70\%$ further improvement w.r.t. FPR$95$ and AUROC on CIFAR-$10$ and $8.60\%$ and $6.61\%$ on CIFAR-$100$. However, the predictor learned on the auxiliary task cannot fit into the real OOD detection task. Thus, we further align the distributions between the auxiliary and real ID data in the transformed space, significantly improving the OOD detection performance with $7.75\%$ on CIFAR-$10$ and $19.14\%$ on CIFAR-$100$ w.r.t. FPR$95$.

\begin{table}[t]
\caption{Effectiveness of auxiliary task crafting and learning scheme of ATOL. $\downarrow$ (or $\uparrow$) indicates smaller (or larger) values are preferred; a bold font indicates the best results in a row.}
\label{tab: ablation}
\centering
\small
\begin{tabular}{cc|c|c|c|c}
\toprule[1.5pt]
\multicolumn{2}{c|}{Ablation}                           & ATOL w/o $\tau$ & ATOL w/o $\ell_{\text{reg}}$ & ATOL w/o $\ell_{\text{align}} $ & ~~~~~ATOL~~~~~  \\ \midrule[1pt]
\multicolumn{1}{c|}{\multirow{2}{*}{CIFAR-10}}  & FPR$95$~$\downarrow$ & 87.64                 & 41.13                   & 22.29            & \textbf{14.54} \\
\multicolumn{1}{c|}{}                           & AUROC~$\uparrow$ & 59.32                 & 90.38                   & 94.08            & \textbf{97.01} \\ \midrule[0.6pt]
\multicolumn{1}{c|}{\multirow{2}{*}{CIFAR-100}} & FPR$95$~$\downarrow$ & 92.90                 & 82.80                   & 74.20            & \textbf{55.06} \\
\multicolumn{1}{c|}{}                           & AUROC~$\uparrow$ & 51.34                 & 73.95                   & 80.56            & \textbf{87.26} \\ \bottomrule[1.5pt]
\end{tabular}
\end{table}

\begin{table}[t]
\small
\caption{Performance comparisons with different setups of the generator on CIFAR benchmarks; $\downarrow$ (or $\uparrow$) indicates smaller (or larger) values are preferred; a bold font indicates the best results in a row.}
\label{tab: generators}
\centering
\begin{tabular}{cc|c|c|c|c}
\toprule[1.5pt]
\multicolumn{2}{c|}{Generators}                           & ~~~DCGAN~~~ & Rand-DCGAN & ~~~~StyleGAN~~~~ & ~~~BigGAN~~~  \\ \midrule[1pt]
\multicolumn{1}{c|}{\multirow{2}{*}{CIFAR-10}}  & FPR$95$~$\downarrow$ & 14.66                & 20.78                   & 13.65            & \textbf{8.61} \\
\multicolumn{1}{c|}{}                           & AUROC~$\uparrow$ & 97.05                & 95.57                   & 96.63            & \textbf{97.90} \\ \midrule[0.6pt]
\multicolumn{1}{c|}{\multirow{2}{*}{CIFAR-100}} & FPR$95$~$\downarrow$ & 55.52                & 65.69                   & 42.62            & \textbf{36.72} \\
\multicolumn{1}{c|}{}                           & AUROC~$\uparrow$ & 86.21                & 83.26                   & 89.17            & \textbf{91.33} \\ \bottomrule[1.5pt]
\end{tabular}
\end{table}

\textbf{Generator Setups. }
On CIFAR benchmarks, we use the generator of the DCGAN model fitting on the ID data in our ATOL. We further investigate whether ATOL remains effective when using generators with other setups, summarized in Table~\ref{tab: generators}. First, we find that even with a random-parameterized DCGAN generator, i.e., {Rand-DCGAN}, our ATOL can still yield reliable performance surpassing its advanced counterparts. Such an observation suggests that our method requires relatively low training costs, a property not shared by previous methods. Second, we adopt more advanced generators, i.e., BigGAN~\cite{biggan19} and StyleGAN-v2~\cite{Karras2019stylegan2}. As we can see, ATOL can benefit from better generators, leading to large improvement $18.80\%$ and $5.12\%$ w.r.t. FPR$95$ and AUROC on CIFAR-$100$ over the DCGAN case. Furthermore, even with complicated generators, our method demonstrates promising performance improvement over other methods with acceptable computational resources. Please refer to the Appendix~\ref{app: generators} and Appendix~\ref{app: efficient} for more details. 


\begin{wraptable}{r}{0.4\linewidth}
\small
\centering
\caption{Computational comparison with other counterparts. We report the per-epoch training time (measured by seconds) on CIFAR benchmarks.}
\vspace{0.5em}
  \label{tab: time}
    \begin{tabular}{c|c}
        \toprule[1.5pt]
        {Methods} & Training time (s)  \\
        \midrule[1pt]
        BoundaryGAN  & $106.98 \pm 7.82$     \\
        ConfGAN      & $111.37 \pm 5.72$     \\
        ManifoldGAN  & $159.23 \pm 9.23$     \\
        G2D          & $73.29 \pm 4.09$        \\
        CMG          & $254.22 \pm 2.42$        \\
        \midrule[0.6pt]
        ATOL          & $\textbf{57.22} \pm 2.28$   \\ \bottomrule[1.5pt]
    \end{tabular}
\end{wraptable}

\textbf{Quantitative analysis on computation. } Data generation-based methods are relatively expensive in computation, so the consumed training time is also of our interest. In Table~\ref{tab: time}, we compare the time costs of training for the considered data generation-based methods. Since our method does not need a complex generator training procedure during the OOD learning step, we can largely alleviate the expensive cost of computation inherited from the data generation-based approaches. Overall, our method requires minimal training costs compared to the advanced counterparts, which further reveals the efficiency of our method. Therefore, our method can not only lead to improved performance over previous methods, but it also requires fewer computation resources.

\section{Conclusion}
Data generation-based methods in OOD detection are a promising way to make the predictor learn to discern ID and OOD data without real OOD data. However, mistaken OOD generation significantly challenges their performance. To this end, we propose a powerful data generation-based learning method by the proposed auxiliary OOD detection task, largely relieving mistaken OOD generation. Extensive experiments show that our method can notably improve detection performance compared to advanced counterparts. Overall, our method can benefit applications for data generation-based OOD detection, and we intend further research to extend our method beyond image classification. We also hope our work can draw more attention from the community toward data generation-based methods, and we anticipate our auxiliary task-based learning scheme can benefit more directions in OOD detection, such as wild OOD detection~\cite{Katz-SamuelsNNL22}.

\begin{ack}
    HTZ, QZW and BH were supported by the NSFC Young Scientists Fund No. 62006202, NSFC General Program No. 62376235, Guangdong Basic and Applied Basic Research Foundation No. 2022A1515011652, HKBU Faculty Niche Research Areas No. RC-FNRA-IG/22-23/SCI/04, and HKBU CSD Departmental Incentive Scheme. XBX and TLL were partially supported by the following Australian Research Council projects: FT220100318, DP220102121, LP220100527, LP220200949, and IC190100031. FL was supported by Australian Research Council (ARC) under Award No. DP230101540, and by NSF and CSIRO Responsible AI Program under Award No. 2303037.
\end{ack}

{
\bibliographystyle{plainnat}
\bibliography{reference}
}


\clearpage

\appendix
\section{Notations}
In this section, we summarize the adopted notations in Table~\ref{tab: notation}.

\begin{table}[htbp]
\centering
\small
\caption{Main notations and their descriptions.}
\begin{tabular}{cl}
\toprule[1.5pt]
\multicolumn{1}{c}{~~~~Notation~~~~} & \multicolumn{1}{c}{Description} \\
\midrule[1pt]
\multicolumn{2}{c}{\cellcolor{greyL} Spaces} \\ 
$\mathcal{X}$ and $\mathcal{Y}$                                                 & the data space and the label space $\{1, \cdots, c\}$                                       \\
$n$ and $c$                                                                            & the dimension of the data space and the dimension of the label space                                                           \\
$d$ and $m$                                                                            & the dimension of the embedding space and the dimension of the latent space                                                       \\

\multicolumn{2}{c}{\cellcolor{greyL} Distributions and Sets}                                                                                                                                    \\
$\mathcal{P}_{\text{X,Y}}^{\text{ID}}$ and $\mathcal{P}_{\text{X}}^{\text{ID}}$ & the joint and the marginal real ID distribution                                             \\
$\mathcal{P}_{\text{X}}^{\text{OOD}}$                                           & the marginal real OOD distribution                                                          \\
$\mathcal{G}_{\text{X,Y}}^{\text{ID}}$ and $\mathcal{G}_{\text{X}}^{\text{ID}}$ & the joint and the marginal auxiliary ID distribution                                        \\
$\mathcal{G}_{\text{X}}^{\text{OOD}}$                                           & the marginal auxiliary OOD distribution                                                     \\
$\mathcal{M}_{\text{Z}}$                                                        & the specified MoG distribution                                                              \\
$\mathcal{U}_{\text{Z}}$                                                        & the specified uniform distribution                                                          \\
\multicolumn{2}{c}{\cellcolor{greyL} Data}                                                                                                                                                      \\
$\mathbf{x}_{\text{ID}}$ and $y_{\text{ID}}$                                    & the real ID data and label                                                                  \\
$\hat{\mathbf{x}}_{\text{ID}}$ and $\hat{y}_{\text{ID}}$                        & the auxiliary ID data and label                                                             \\
$\hat{\mathbf{x}}_{\text{OOD}}$                                                 & the auxiliary OOD data                                                                      \\
$\mathbf{z}_{\text{ID}}$ and $\mathbf{z}_{\text{OOD}}$                          & the latent ID and the latent OOD data                                                       \\
~~~~$\mathcal{Z}^{\text{ID}}$ and $\mathcal{Z}^{\text{OOD}}$~~~~                        & the latent ID and the latent OOD data sets                                                  \\
\multicolumn{2}{c}{\cellcolor{greyL} Models}                                                                                                                                                    \\
$\mathbf{h}$                                                                    & the predictor: $\mathbb{R}^{n} \rightarrow \mathbb{R}^c$                                    \\
$\boldsymbol\phi$ and $\boldsymbol\rho$                                         & the feature extractor and the classifier                                                    \\
$s(\cdot;\mathbf{h})$                                                           & the scoring function:$\mathbb{R}^{n}\rightarrow \mathbb{R}$                                 \\
$f_{\beta}(\cdot)$                                                              & the OOD detector:$\mathcal{X}\rightarrow \{\text{ID}, \text{OOD}\}$, with threshold $\beta$~~~~~~ \\
$G$                                                                             & the generator: $\mathbb{R}^m\rightarrow\mathbb{R}^n$                                        \\
\multicolumn{2}{c}{\cellcolor{greyL} Loss and Function}                                                                                                                                         \\
$\ell_{\text{CE}}$ and $\ell_{\text{OE}}$                                         & ID loss function and OOD loss function                                                      \\
$\ell_{\text{reg}}$                                                             & the generator regularization loss function                                                  \\
$\ell_{\text{align}}$                                                           & the alignment loss                                                                          \\
$\boldsymbol{\phi}'(\cdot)$                                          & the general mapping function                                                                \\
$\mathcal{M}(\cdot)$                                                            & the density function of MoG                                                                 \\ 
\bottomrule[1.5pt]
\end{tabular}
\label{tab: notation}
\end{table}

\section{Related Works}
\label{app: related works}
{\bfseries OOD Scoring Functions}. To discern OOD data from ID data, many works study how to devise effective \emph{OOD scoring functions}, typically assuming a well-trained model on ID data with its fixed parameters~\cite{hendrycks2016baseline,energy2020}. As a baseline method, \citet{hendrycks2016baseline} takes the \emph{maximum softmax prediction} (MSP) as the OOD score, where, in expectation, the MSP should be low for OOD data since their true labels are not in the considered label space. However, MSP frequently makes mistakes due to the over-confidence issue~\cite{energy2020}. Therefore, recent works devise improved scoring strategies~\cite{energy2020,MOS2021,Hendrycks22maxlogit,vim2022,knn2022}; integrate gradient information~\cite{igoe2022useful, huang2021importance, LiangLS18} or embedding features~\cite{lee2018simple,sastry2019detecting,djurisic2022ash, morteza2021provable}; make further adjustments on specific tasks~\cite{sun2021react,wang2022watermarking,Yadan-1,Yadan-2}. {In Appendix~\ref{app: different scoring}, we test ATOL with different scoring strategies, demonstrating that proper scoring functions can lead to improved performance for data generation-based OOD detection. Therefore, OOD scoring functions are typically orthogonal to data generation-based approaches.}

{\bfseries OOD Training Strategies}. OOD detection can also be improved by model fine-tuning, motivating advanced works studying OOD training strategies. As one of the most potent approaches, \emph{outlier exposure}~\cite{Hendrycks2019oe} train to make the perdictor produce low-confidence predictions for OOD data. Based on outlier exposure, a set of improved methods have been proposed, from the perspective of data re-sampling~\cite{mixupoe2021,ming2022poem}, background classification~\cite{LiV20,atom2020, MohseniPYW20}, data transformation~\cite{wang2023doe}, adversarial robust learning~\cite{LiV20, 0001AB19, chen2022anomman}, meta learning~\cite{JeongK20} and energy scoring~\cite{energy2020,Katz-SamuelsNNL22}. However, real OOD data are hard to be accessed, largely hindering their practical validity. 

Therefore, a set of learning strategies has been proposed, considering situations where real OOD data are unavailable. Therein, improved representation~\cite{Tack20CSI,ssd2021,WangZZZLSW22, zaeemzadeh2021ood, logitnorm22,ming2022cider}, extra-class learning~\cite{MOS2021,openmatch2021}, and pseudo features learning~\cite{du2022vos, tao2023nonparametric} have demonstrated their strengths for improved detection. However, these methods can hardly beat the outlier exposure-based methods. Therefore, several works~\cite{LeeLL2018,Vernekar19via,Sricharan18robust,G2D2021,CMG2022} adopt data generators to synthesize OOD data for model training. They can make the predictor learn to discern ID and OOD patterns without tedious OOD data acquisition. However, the data generator may wrongly generate unreliable data containing ID semantics, confusing the predictor between ID and OOD cases.

\section{Proof of Proposition~\ref{the: oe}}
\label{app: proof}

\begin{proof}
    Following Theorem 3 in \citet{fang2022out}, if Eq.~\eqref{eq: predictor theorem} approaches $0$ and \textbf{C}\ref{con: supp} holds, the predictor $\mathbf{h} = \boldsymbol{\rho} \circ \boldsymbol{\phi}$ can separate the auxiliary ID and the auxiliary OOD cases well, i.e., $\texttt{supp}\boldsymbol{\phi}_{\#}\mathcal{G}^{\text{ID}}_{X}\cap \texttt{supp}\boldsymbol{\phi}_{\#}\mathcal{G}^{\text{OOD}}_{X}=\emptyset$. Then, by aligning the real and the auxiliary ID distributions in transformed space (cf. \textbf{C}\ref{con: transfer}), we can conclude that the auxiliary OOD data are almost reliable w.r.t. the real ID data.
\end{proof}

\section{Overall Algorithm}
\label{app: overall alg}
In this section, we discuss our learning framework in detail. Our ATOL consists of two stages: 1) generator regularization and 2) auxiliary task OOD learning.

For the generator regularization, the overall training framework is summarized in Algorithm~\ref{alg: ATOL generator}, regularizing in a stochastic manner with \texttt{num\_step\_g} iterations. We have the generator $G: \mathbb{R}^m\rightarrow \mathbb{R}^n$ and the pre-defined latent distribution $\mathcal{U}_{Z}$ in the latent space. In each training step, we sample a set of latent data from $\mathcal{U}_{Z}$, assuming be of the size $b$ as that of the mini-batch. With the regularization term, we update the generator via one step of mini-batch gradient descent. 

\begin{algorithm}[h]  
\caption{Generator Reguralization.} 
\label{alg: ATOL generator}
\begin{algorithmic}[1]
\STATE {\bfseries Inputs:} initialized generator $G(\cdot)$ and the pre-defined latent distribution $\mathcal{U}_{Z}$.
\FOR{$t=1$ \textbf{to} \texttt{num\_step\_g}}
    \STATE Sample latent data $\{\mathbf{z}\}^{b}_{i=1}$ from $\mathcal{U}_{Z}$;
    \STATE Compute regularization risk $\ell_{\text{reg}}(\mathbf{z};G)$;
    \STATE Update generator $\min_{G}\ell_{\text{reg}}(\mathbf{z};G)$;
\ENDFOR
\STATE {\bfseries Output:} regularized generator $G(\cdot)$.
\end{algorithmic}
\end{algorithm}

For the auxiliary task OOD learning, the overall training framework is summarized in Algorithm~\ref{alg: ATOL predictor}, optimizing the predictor in a stochastic manner with \texttt{num\_step\_p} iterations. In each training step, with the regularized generator, the auxiliary ID data $\hat{\mathbf{x}}_{\text{ID}}$ and the auxiliary OOD data $\hat{\mathbf{x}}_{\text{OOD}}$ can be generated from the latent ID data $\mathbf{z}_{\text{ID}}$ and the latent OOD data $\mathbf{z}_{\text{OOD}}$ respectively, where the latent ID and OOD data follow the crafted distribution $\mathcal{Z}_{\text{ID}}$, $\mathcal{Z}_{\text{OOD}}$\footnote{With abuse of notation, we denote the distribution in latent space as $\mathcal{Z}$ for simplicity.} in Eqs~\eqref{eq: latent ID}-\eqref{eq: latent OOD}. We assume the size $b$ as that of the mini-batch regarding the ID samples and $b'$ for the OOD samples\footnote{Note that the mini-batch size of the real ID data and the auxiliary ID data have no need to be equal in the Algorithm~\ref{alg: ATOL predictor}. In this paper we make them equal for convenience}. Then, the risk for the auxiliary and the real data are jointly computed and update the predictor $\mathbf{h}$ parameters using Eq.~\eqref{eq: overall}. After training, we apply the MaxLogit scoring in discerning ID and OOD cases.

\begin{algorithm}[h]  
\caption{Auxiliary Task OOD Learning.} 
\label{alg: ATOL predictor}
\begin{algorithmic}[1]
\STATE {\bfseries Inputs:} predictor $\mathbf{h}=\boldsymbol{\rho} \circ \boldsymbol{\phi}$, regularized generator $G(\cdot)$, real ID distribution $\mathcal{P}^{\text{ID}}_{X,Y}$, crafted latent distributions $\mathcal{Z}_{\text{ID}}$ and $\mathcal{Z}_{\text{OOD}}$;
\FOR{$t=1$ \textbf{to} \texttt{num\_step\_p}}
    \STATE Sample latent data $\{\mathbf{z}_{\text{ID}}\}^{b}_{i=1}$ and $\{\mathbf{z}_{\text{OOD}}\}^{b'}_{i=1}$ from $ \mathcal{Z}_{\text{ID}}$ and $\mathcal{Z}_{\text{OOD}}$, resp;
    \STATE Generate auxiliary data $\hat{\mathbf{x}}_{\text{ID}}=G(\mathbf{z}_{\text{ID}})$ and $\hat{\mathbf{x}}_{\text{OOD}}=G(\mathbf{z}_{\text{OOD}})$ by regularized generator $G(\cdot)$ and sample real ID data $\{(\mathbf{x}_{\text{ID}}, y_{\text{ID}})\}_{i=1}^{b}$ from $\mathcal{P}^{\text{ID}}_{X,Y}$;
    \STATE Compute risk $\ell(\mathbf{h})=\ell_{\text{CE}}(\mathbf{x}_{\text{ID}}, y_{\text{ID}};\hat{\mathbf{x}}_{\text{ID}}, \hat{y}_{\text{ID}};\mathbf{h}) + \ell_{\text{OE}}(\hat{\mathbf{x}}_{\text{OOD}};\mathbf{h})+\ell_{\text{align}}(\hat{\mathbf{x}}_{\text{ID}}, \hat{y}_{\text{ID}};\boldsymbol{\phi}) $;
    \STATE Update predictor $\min_{\mathbf{h}}\ell(\mathbf{h})$;
\ENDFOR
\STATE {\bfseries Output:} predictor $\mathbf{h}(\cdot)$.
\end{algorithmic}
\end{algorithm}

\section{Details of Experiment Configuration}
\subsection{Hyper-parameters}
\label{app: hyper}
We study the effect of hyper-parameters on the final performance of our ATOL, where we consider the trade-off parameter $\alpha$, mean value $\mu$, covariance matrix scale $\sigma$ for \emph{Mixture of Gaussian}(MoG), the sampling space size $u$ in latent space, and the dimension of latent space $m$. We also study the impact of transformed data from different layers of WRN-40-2. All the above experiments are conducted on the CIFAR benchmarks.

\textbf{Trade-off parameter $\alpha$. } Table~\ref{tab: trade-off} demonstrates the performance of ATOL with varying $\alpha$ values that trade-off the OOD detection task learning loss and the alignment losses. $\alpha$ is selected from $\{0.01, 0.1, 0.5, 1, 3, 5, 10\}$. We can observe that on CIFAR-10, the best performance is obtained at $\alpha = 1$, and the best performance on CIFAR-100 is obtained at $\alpha = 5$. In general, a large $\alpha$ is advantageous for the accuracy of the predictor since ID distribution alignment not only enables the transfer of the ability to discern ID and OOD data but also facilitates the ID classification of the predictor. However, when $\alpha$ is large, the predictor tends to align the ID data rather than discern ID and OOD cases at the early stage of training. Hence, a relatively small $(\alpha \le 5)$ usually leads to good performance than a much larger value. 

\begin{table}[htbp]
\caption{Performance of ATOL with varying $\alpha$ on CIFAR-$10$ and CIFAR-$100$}
\label{tab: trade-off}
\centering
\small
\begin{tabular}{c|ccc|ccc}
\toprule[1.5pt]
\multirow{2}{*}{Trade-off parameter $\alpha$} & \multicolumn{3}{c|}{CIFAR-$10$}              & \multicolumn{3}{c}{CIFAR-$100$}                  \\ \cline{2-7} 
& FPR$95$ $\downarrow$  & AUROC $\uparrow$      & Acc ID $\uparrow$    & FPR$95$ $\downarrow$ & AUROC $\uparrow$ & Acc ID $\uparrow$ \\ \midrule[1pt]
$\alpha = 0.01$ & 22.99       & 92.66       & 92.89       & 71.07      & 80.45  & 72.67  \\
$\alpha = 0.1$  & 20.02       & 94.66       & 93.43       & 65.99      & 82.92  & 73.44  \\
$\alpha = 0.5$  & 15.56       & 96.05       & 94.03       & 60.23      & 84.88  & 73.89  \\
$\alpha = 0.8$  & 15.51       & 96.40       & 94.15       & 60.23 & 84.93      & 73.99      \\
$\alpha = 1$    & \textbf{14.11}       & \textbf{96.94}       & 94.17       & 59.89& 84.77      & 73.99      \\
$\alpha = 3$    & 16.01       & 96.56       & 93.96       & 56.53 & 86.37      & 73.89      \\
$\alpha = 5$    & 16.28       & 96.57       & 94.17       & \textbf{55.38}& \textbf{86.41}     & 73.98      \\
$\alpha = 10$   & 17.08       & 96.50       & \textbf{94.57} & 61.82& 84.97      & \textbf{74.58}     \\ \bottomrule[1.5pt]
\end{tabular}
\end{table}

\textbf{Mean values for MoG $\mu$. } We show the effect of the mean value $\mu$ of sub-Gaussian from the Mixture of Gaussian $\mathcal{M}$ in the latent space. The mean value $\mu$ decides the latent distribution for the auxiliary ID data, and the results are listed in Table~\ref{tab: mean}. We conduct experiments with two realizations of the Gaussian mean: case $1$ is to choose at random from a uniform distribution, and case $2$ (used in ATOL) is to obtain the values from the vertices of a high-dimensional hypercube (i.e., a mean vector with $m$ elements, where each element is randomly selected from between $\{-\mu, \mu\}$ as the mean for MoG).

Specifically, the performance of case $1$ is stable, since the randomness of value sampling. In contrast, in case $2$, the mean value plays an important role in the performance of ATOL, where the mean for each sub-distribution is fixed at a set value. In general, a relatively large mean value for the MoG usually leads to good performance than a much smaller value. When the mean value is small, the MoG will concentrate on a limited region, leading to the great overlapping among sub-Gaussians. Such an overlapping can result in confusion of the semantics of the auxiliary ID data. Moreover, a proper choice value for the case $2$ is superior to the case $1$, reflecting that our strategy is useful for our proposed ATOL.

\begin{table}[!htbp]
\caption{Performance of ATOL with varying $\mu$ on CIFAR-$10$ and CIFAR-$100$}
\label{tab: mean}
\centering
\resizebox{\linewidth}{!}{
\begin{tabular}{cc|ccccccc|ccccccc}
\toprule[1.5pt]
\multicolumn{2}{c|}{\multirow{2}{*}{Mean $\mu$}}              & \multicolumn{7}{c|}{Case $1$}                             & \multicolumn{7}{c}{Case $2$}                              \\ 
\multicolumn{2}{c|}{} & 0.1   & 0.5   & 1     & 1.5   & 2     & 3     & 5     & 0.1   & 0.5   & 1     & 1.5   & 2     & 3     & 5     \\ \midrule[1pt]
\multicolumn{1}{c|}{\multirow{2}{*}{CIFAR-$10$}}  & FPR$95$ $\downarrow$ & 17.60 & 16.26 & 16.26 & 16.20 & 16.30 & 16.29 & 16.31        & 23.29 & 21.32 & 18.62 & 16.08 & 15.14 & 14.69 & \textbf{14.52} \\
\multicolumn{1}{c|}{}                           & AUROC $\uparrow$ & 96.13 & 96.59 & 96.56 & 96.55 & 96.56 & 96.55 & 96.54      & 93.44 & 94.61 & 95.85 & 96.37 & 96.75 & 97.02 & \textbf{97.05} \\ \midrule[0.6pt]
\multicolumn{1}{c|}{\multirow{2}{*}{CIFAR-$100$}} & FPR$95$ $\downarrow$ & 63.93 & 64.04 & 63.42 & 62.55 & 63.05 & 63.84 & 63.56        & 62.08 & 63.17 & \textbf{55.24} & 59.49 & 60.06 & 61.80 & 62.54 \\
\multicolumn{1}{c|}{}                           & AUROC $\uparrow$ & 83.26 & 83.37 & 83.54 & 84.42 & 84.14 & 83.80 & 84.08      & 84.19 & 82.28 & 86.35 & \textbf{86.98} & 85.72 & 85.22 & 84.85 \\ 
\bottomrule[1.5pt]
\end{tabular}}
\end{table}

\textbf{Covariance matrix for MoG $\sigma$. } We show the effect of $\sigma$ for the Mixture of Gaussian $\mathcal{M}$ in the latent space. The value $\sigma$ decides the covariance matrix of the latent distribution for the auxiliary ID data. We vary $\sigma \in \{0.05, 0.1, 0.3, 0.5, 1, 1.5\}$, and the results are listed in Table~\ref{tab: std}. As we can see, too large $\sigma$ leads to inferior performance since too large $\sigma$ will result in largely overlapping with other Gaussian in latent space, as we discussed earlier. On CIFAR-$10$ dataset, the best performance is obtained at $\sigma = 0.1$, while on CIFAR-$100$, the best performance is obtained at $\sigma = 0.5$. We suppose that the semantics and scale of the varying datasets differ, necessitating different $\sigma$.

\begin{table}[htbp]
\caption{Performance of ATOL with varying $\sigma$ on CIFAR-$10$ and CIFAR-$100$}
\label{tab: std}
\centering
\small
\begin{tabular}{cc|cccccc}
\toprule[1.5pt]
\multicolumn{2}{c|}{Covariance matrix scale $\sigma$}                & $\sigma = 0.01$ & $\sigma = 0.1$ & $\sigma = 0.3$ & $\sigma = 0.5$ & $\sigma = 1$ & $\sigma = 1.5$ \\ 
\midrule[1pt]
\multicolumn{1}{c|}{\multirow{2}{*}{CIFAR-$10$}}  & FPR$95$ $\downarrow$ 
& 14.67     & \textbf{14.62}    & 14.88    & 15.99  & 20.09  & 21.31   \\
\multicolumn{1}{c|}{}                           & AUROC $\uparrow$ 
& 96.99     & \textbf{97.02}    & 96.99    & 96.37  & 95.45  & 94.61   \\ \midrule[0.6pt]
\multicolumn{1}{c|}{\multirow{2}{*}{CIFAR-$100$}} & FPR$95$ $\downarrow$ 
& 63.97     & 62.82    & 60.44    & \textbf{56.06}  & 63.72  & 64.42   \\
\multicolumn{1}{c|}{}                           & AUROC $\uparrow$ 
& 84.50     & 84.85    & 85.70    & \textbf{86.49}  & 82.95  & 83.58   \\ 
\bottomrule[1.5pt]
\end{tabular}
\end{table}

\textbf{Space size for latent space $u$. } We show the effect of $u$ for the latent distribution $\mathcal{U}_Z$, which identifies the space size of $\mathcal{U}_Z$ in latent space for the auxiliary OOD data. We vary $u \in \{0.5, 1, 4, 8, 16, 32\}$, and the results are listed in Table~\ref{tab: uniform}. Generally, a small latent space leads to the limited information in the latent space, which results in the inferior performance (e.g., the ATOL performance when $u=0.5$). As we set a relatively large value $(u \ge 4)$, the performances are stable on both datasets.

\begin{table}[htbp]
\caption{Performance of ATOL with varying $u$ on CIFAR-$10$ and CIFAR-$100$}
\label{tab: uniform}
\centering
\small
\begin{tabular}{cc|cccccc}
\toprule[1.5pt]
\multicolumn{2}{c|}{Latent space size $u$}  & $u = 0.5$ & $u = 1$ & $u = 4$ & $u = 8$ & $u = 16$ & $u = 32$ \\ 
\midrule[1pt]
\multicolumn{1}{c|}{\multirow{2}{*}{CIFAR-$10$}}  & FPR$95$ $\downarrow$ 
& 18.15     & 14.90    & 13.96    & 13.90  & 14.12  & 14.02   \\
\multicolumn{1}{c|}{}                           & AUROC $\uparrow$ 
& 93.96     & 95.97    & 96.96    & 96.96  & 96.95  & 96.96   \\ \midrule[0.6pt]
\multicolumn{1}{c|}{\multirow{2}{*}{CIFAR-$100$}} & FPR$95$ $\downarrow$ 
& 60.18     & 58.02    & 56.25    & 56.42  & 56.09  & 55.97   \\
\multicolumn{1}{c|}{}                           & AUROC $\uparrow$ 
& 83.42     & 85.43    & 86.64    & 86.57  & 86.38  & 86.44   \\ 
\bottomrule[1.5pt]
\end{tabular}
\end{table}

\textbf{Dimension of latent space $m$. } To study the impact of dimension $m$ of latent space (the input space of the data generator), we conduct experiments based on different dimensions of a random-parameterized generator of DCGAN. We vary $m \in \{8, 16, 32, 64, 128, 256\}$, and the results are listed in Table~\ref{tab: dimension}. We find that the generator with $m=64$ has the best performance. We suppose that a data generator with high dimensional input space may be more intractable, while low dimensional input space may not contain sufficient information. Therefore, a reasonably large $m$ helps achieve a better result. 

\begin{table}[htbp]
\caption{Performance of ATOL with varying $m$ on CIFAR-$10$ and CIFAR-$100$}
\label{tab: dimension}
\centering
\small
\begin{tabular}{cc|cccccc}
\toprule[1.5pt]
\multicolumn{2}{c|}{Dimension $m$}  & $m = 8$ & $m = 16$ & $m = 32$ & $m = 64$ & $m = 128$ & $m = 256$ \\ 
\midrule[1pt]
\multicolumn{1}{c|}{\multirow{2}{*}{CIFAR-$10$}}  & FPR$95$ $\downarrow$ 
& 24.98     & 22.71    & 22.62    & 20.78  & 28.39  & 29.55   \\
\multicolumn{1}{c|}{}                           & AUROC $\uparrow$ 
& 94.74     & 94.84    & 94.81    & 95.57  & 94.11  & 93.12   \\ \midrule[0.6pt]
\multicolumn{1}{c|}{\multirow{2}{*}{CIFAR-$100$}} & FPR$95$ $\downarrow$ 
& 70.62    & 70.03    & 68.33    & 65.69  & 65.93  & 71.83   \\
\multicolumn{1}{c|}{}                           & AUROC $\uparrow$ 
& 80.45     & 80.65    & 82.30    & 83.26  & 82.86  & 78.90   \\ 
\bottomrule[1.5pt]
\end{tabular}
\end{table}

The above experiments about hyper-parameters only serve to support the validity of our method. However, a proper choice of hyper-parameters can truly induce improved results in effective OOD detection, reflecting that all the introduced hyper-parameters are useful in our proposed ATOL. 

\subsection{{Effect of Mistaken OOD Generation}}
\label{app: motivation exp}
{In section~\ref{sec: intro}, we revisit the common baseline approach~\cite{Hendrycks2019oe}, which uses real OOD data, for OOD detection. We investigate the effect of mistaken OOD generation on the OOD detection performance. In particular, we use a WRN-40-2 architecture trained on CIFAR-100 with varying proportions of real ID data mixed in the real OOD data, reflecting the severity of wrong OOD data during training. As shown in Figure~\ref{fig: mistaken impacts}, the performance (FPR95) degrades rapidly from 35.70\% to 64.21\% as the proportion of unreliable OOD data increases from 0\% to 75\%. This trend signifies that current OOD detection methods are indeed challenged by mistaken OOD generation, which motivates our work.}

\subsection{Realizations of Distance} As described in \ref{sec: craft}, we suppose that the centralized distance between two data points that are measured in the latent space should be positively correlated to that of the distance measured in the data space, namely, distance-preserving. In this ablation, we contrast the performance of different realizations used for the normalized distance in Eq.~\ref{eq: map}. We empirically test three realizations: 1) \emph{Cosine similarity}-based: $d(\mathbf{z}_1, \mathbf{z}_2) = \frac{\mathbf{z}_1^\top \mathbf{z}_2}{\left \|\mathbf{z}_1\right \| \left \|\mathbf{z}_2\right \|}$; 2) \emph{Taxicab distance}-based: $d(\mathbf{z}_1,\mathbf{z}_2)=\left \| \mathbf{z}_1 - \mathbf{z}_2\right \|_1-\mathbb{E} \left \| \mathbf{z}_1 - \mathbf{z}_2\right \|_1$; 3) \emph{Euclidean distance}-based~\cite{xia2023moderate}: $d(\mathbf{z}_1,\mathbf{z}_2)=\left \| \mathbf{z}_1 - \mathbf{z}_2\right \|_2-\mathbb{E} \left \| \mathbf{z}_1 - \mathbf{z}_2\right \|_2$. As one can see from the Table~\ref{tab: distance}, all the forms of distance can lead to reliable OOD detection performance, indicating that ATOL is general to various realizations.

\begin{table}[htbp]
\small
\caption{Performance comparisons with different realizations of centralized distance}
\label{tab: distance}
\centering
\begin{tabular}{c|cc|cc}
\toprule[1.5pt]
\multirow{2}{*}{Centralized distance $d$} & \multicolumn{2}{c|}{CIFAR-10} & \multicolumn{2}{c}{CIFAR-100} \\ \cline{2-5} & FPR$95$ $\downarrow$ & AUROC $\uparrow$ & FPR$95$ $\downarrow$ & AUROC $\uparrow$ \\ \midrule[1pt]
Cosine similarity-based     & 19.73 & 95.77 & 60.15 & 85.41 \\
Taxicab distance-based     & 18.40 & 94.78 & 58.76 & 85.83 \\
Euclidean distance-based    & \textbf{14.62}  & \textbf{97.02} & \textbf{55.52} & \textbf{86.32} \\ 
\bottomrule[1.5pt]
\end{tabular}
\end{table}

\subsection{Class-agnostic Auxiliary ID Data} 
ATOL adopts a class-conditional way to generate auxiliary ID data~\cite{VHL2022, XiaPICMM22}, then aligns transformed distribution with real ID data based on labels. In this ablation, we contrast with a class-agnostic implementation, i.e., we generate the auxiliary ID data from one single Gaussian and randomly assign labels to the auxiliary ID data. The parameter of the single Gaussian is similar to the sub-Gaussian in the MoG, except that the mean value is the zero vector. Under the same training setting, the class-agnostic way for auxiliary ID data leads to a worse result, which may be caused by random labels and unavailable alignment. Even if class-agnostic approaches struggle due to the homogenization of auxiliary data, the predictor can still learn to discern ID and OOD data.
\begin{table}[htbp]
\small
\caption{Performance comparisons with different implementations of auxiliary ID data}
\label{tab: class}
\centering
\begin{tabular}{c|cc|cc}
\toprule[1.5pt]
\multirow{2}{*}{Methods} & \multicolumn{2}{c|}{CIFAR-10} & \multicolumn{2}{c}{CIFAR-100} \\ \cline{2-5} & FPR$95$ $\downarrow$ & AUROC $\uparrow$ & FPR$95$ $\downarrow$ & AUROC $\uparrow$ \\ \midrule[1pt]
Class-agnostic     & 21.72 & 93.93 & 61.54 & 84.34 \\
Class-conditional  & \textbf{14.62}  & \textbf{97.02} & \textbf{55.52} & \textbf{86.32} \\ 
\bottomrule[1.5pt]
\end{tabular}
\end{table}

\subsection{Sample More Auxiliary Data}
We note that with a powerful generation-based learning scheme, we can generate sufficient data to make the predictor learn the knowledge of OOD without tedious OOD data acquisition. Therefore, we consider increasing the number of generated auxiliary data, making the predictor see more auxiliary data to strengthen the ability to discern ID and OOD data. To verify the effect of generating more data, we conduct ATOL on CIFAR-$10$ and CIFAR-$100$ with different batch size $b$ and $b'$ w.r.t. the ID and OOD data, respectively. The experiment results are shown in Table~\ref{tab: batch size}, demonstrating that generating more auxiliary data could strengthen the effect of ATOL. However, larger batch size means the extra calculation cost, which may be improved in the future.

\begin{table}[htbp]
\caption{Performance of ATOL with varying batch size of ID and OOD data}
\label{tab: batch size}
\centering
\small
\begin{tabular}{ll|cc|cc}
\toprule[1.5pt]
\multicolumn{2}{c|}{Batch size}                                                                      & \multicolumn{2}{c|}{CIFAR-$10$} & \multicolumn{2}{c}{CIFAR-$100$} \\ \cline{3-6} 
\multicolumn{1}{c}{$b$}      & \multicolumn{1}{c|}{$b'$}   & FPR$95$ $\downarrow$ & AUROC $\uparrow$ & FPR$95$ $\downarrow$ & AUROC $\uparrow$ \\ \midrule[1pt]

\multicolumn{1}{l|}{$b=32$}  & $b'=32$& 22.21& 95.04    & 65.94     & 83.67 \\
\multicolumn{1}{l|}{$b=32$}  & $b'=128$       & 15.43& 96.83    & 58.77 & 83.98 \\
\multicolumn{1}{l|}{$b=64$}  & $b'=64$& 19.60 & 95.38    & 63.26     & 84.12 \\
\multicolumn{1}{l|}{$b=64$}  & \multicolumn{1}{l|}{$b'=256$}  & 14.21& 96.97    & 55.03     & 86.16 \\
\multicolumn{1}{l|}{$b=128$} & $b'=128$       & 20.39& 95.43    & 63.62     & 82.33 \\
\multicolumn{1}{l|}{$b=128$} & \multicolumn{1}{l|}{$b'=512$}  & 13.99 & 97.09    & 54.96    & 86.17 \\
\multicolumn{1}{l|}{$b=256$} & \multicolumn{1}{l|}{$b'=256$}  & 19.25 & 95.72    & 62.94     & 83.97 \\
\multicolumn{1}{l|}{$b=256$} & \multicolumn{1}{l|}{$b'=1024$} & \textbf{13.68} & \textbf{97.18}    & \textbf{54.71}     & \textbf{86.22} \\ \bottomrule[1.5pt]
\end{tabular}
\end{table}

\subsection{Using Embedding of Different Layers} 
To study the impact of embedding spaces from different layers, we conduct ID distribution alignment on different output layers of WRN-40-2. In Table~\ref{tab: layers}, we find that using the embedding space from the penultimate layer of WRN-40-2 achieves the best performance. We suppose that alignment based on a too-shallow layer may not be enough to impact the embedding of subsequent layers. However, the calibration based on the last layer may interrupt the normal classification between the real data and auxiliary data, since that they have completely different high-level semantics.

\begin{table}[htbp]
\small
\caption{Performance comparisons with different layers on CIFAR-$10$ and CIFAR-$100$}
\label{tab: layers}
\centering
\begin{tabular}{c|cc|cc}
\toprule[1.5pt]
\multirow{2}{*}{Different embedding layers} & \multicolumn{2}{c|}{CIFAR-10} & \multicolumn{2}{c}{CIFAR-100} \\ \cline{2-5} & FPR$95$ $\downarrow$ & AUROC $\uparrow$ & FPR$95$ $\downarrow$ & AUROC $\uparrow$ \\ \midrule[1pt]
Block $1$ ($h_{\text{shallow}}$)     & 31.78 & 91.01 & 72.88 & 80.15 \\
Block $2$ ($h_{\text{middle}}$)      & 23.43     & 93.62     & 70.04     & 81.52     \\
Block $3$ ($h_{\text{deep}}$)   & \textbf{13.68}  & \textbf{97.18} & \textbf{55.52} & \textbf{86.32} \\ 
Last layer ($h_{\text{last}}$)   & 27.11     & 91.85   & 65.12     & 83.28     \\
\bottomrule[1.5pt]
\end{tabular}
\end{table}

\subsection{{Using Different Network Architectures}} 
{In the main paper, we have shown that ATOL is competitive on WRN-40-2. The following experimental results on CIFAR benchmarks can support our claims~\ref{the: oe}, where using a more complex model (i.e., DenseNet-121~\cite{huang2017densely}) can lead to better performance in OOD detection. All the numbers are reported over OOD test datasets described in Section~\ref{sec: setup}.}

\begin{table}[htbp]
\caption{Performance comparisons with different network architectures DenseNet-121 on CIFAR-$10$ and CIFAR-$100$ $\downarrow$ (or $\uparrow$) indicates smaller (or larger) values are preferred.}\label{tab: backbone}
\vspace{0.5em}
\resizebox{\linewidth}{!}{
\begin{tabular}{c|cccccccccccccc}
\toprule[1.5pt]
\multicolumn{1}{c|}{\multirow{2}{*}{Method}} & \multicolumn{2}{c}{SVHN}   & \multicolumn{2}{c}{LSUN-Crop}         & \multicolumn{2}{c}{LSUN-Resize}       & \multicolumn{2}{c}{iSUN}   & \multicolumn{2}{c}{Texture}& \multicolumn{2}{c|}{Places365}         & \multicolumn{2}{c}{\textbf{Average}}\\ \cline{2-15} 
& FPR95 $\downarrow$ & AUROC $\uparrow$ & FPR95 $\downarrow$ & AUROC $\uparrow$ & FPR95 $\downarrow$ & AUROC $\uparrow$ & FPR95 $\downarrow$ & AUROC $\uparrow$ & FPR95 $\downarrow$ & AUROC $\uparrow$ & FPR95 $\downarrow$ & \multicolumn{1}{c|}{AUROC $\uparrow$} & FPR95 $\downarrow$ & AUROC $\uparrow$ \\ \midrule[1pt]
\multicolumn{15}{c}{CIFAR-$10$} \\ 
\midrule[0.6pt]
\multicolumn{1}{c|}{WRN-40-2} & 20.60 & 96.03  & 1.48  & 99.59  & 5.20  & 98.78  & 5.00  & 98.76  & 26.05 & 95.03  & 27.55 & \multicolumn{1}{c|}{94.33}  & 14.31 & 97.09  \\
\multicolumn{1}{c|}{DenseNet-121} & 18.05 & 96.27  & 1.45  & 99.58  & 4.35  & 99.88	  & 4.45  & 98.86  & 20.90 & 95.70  & 25.85 & \multicolumn{1}{c|}{94.39}  & 12.51 & 97.28  \\

\midrule[0.6pt]
\multicolumn{15}{c}{CIFAR-$100$} \\ 
\midrule[0.6pt]

\multicolumn{1}{c|}{WRN-40-2} & 70.85 & 84.70  & 13.45  & 97.52  & 51.85  & 90.12  & 55.80  & 89.02  & 63.10 & 83.37  & 75.30 & \multicolumn{1}{c|}{78.86}  & 55.06 & 87.26  \\
\multicolumn{1}{c|}{DenseNet-121} & 70.30 & 85.96  & 61.55  & 87.93  & 22.15  & 95.87  & 22.30  & 95.59  & 48.30 & 87.87  & 77.15 & \multicolumn{1}{c|}{79.28}  & 50.29 & 88.75  \\
\bottomrule[1.5pt]
\end{tabular}}
\end{table}

\subsection{How to best perform data generation-based methods for OOD detection}
\label{app: generators}
BoundaryGAN~\cite{LeeLL2018}, ConfGAN~\cite{Sricharan18robust} and ManifoldGAN~\cite{Vernekar19via} use DCGAN~\cite{DCGAN15} to generate OOD data to benefit the predictor for OOD detection. As we show in Section~\ref{sec: main results}, even with the DCGAN, ATOL has shown promising OOD detection performance. Moreover, we argue for ATOL can profit from a delicate data generators~\cite{chen2023humanmac}, which can generate more diverse data to further benefit the predictor learning from generated data. To this end, we use the generator of StyleGAN-v2 and BigGAN as the data generator for ATOL, namely, \emph{ATOL-StyleGAN} and \emph{ATOL-BigGAN} (\emph{ATOL-S} and \emph{ATOL-B} for short), which are one of the most popular generative models.

\begin{table}[htbp]
\caption{Performance comparisons with different data generators on CIFAR-$10$ and CIFAR-$100$ $\downarrow$ (or $\uparrow$) indicates smaller (or larger) values are preferred; a bold font indicates the best results. }\label{tab: ATOL+}
\vspace{0.5em}
\resizebox{\linewidth}{!}{
\begin{tabular}{c|cccccccccccccc}
\toprule[1.5pt]
\multicolumn{1}{c|}{\multirow{2}{*}{Method}} & \multicolumn{2}{c}{SVHN}   & \multicolumn{2}{c}{LSUN-Crop}         & \multicolumn{2}{c}{LSUN-Resize}       & \multicolumn{2}{c}{iSUN}   & \multicolumn{2}{c}{Texture}& \multicolumn{2}{c|}{Places365}         & \multicolumn{2}{c}{\textbf{Average}}\\ \cline{2-15} 
& FPR95 $\downarrow$ & AUROC $\uparrow$ & FPR95 $\downarrow$ & AUROC $\uparrow$ & FPR95 $\downarrow$ & AUROC $\uparrow$ & FPR95 $\downarrow$ & AUROC $\uparrow$ & FPR95 $\downarrow$ & AUROC $\uparrow$ & FPR95 $\downarrow$ & \multicolumn{1}{c|}{AUROC $\uparrow$} & FPR95 $\downarrow$ & AUROC $\uparrow$ \\ \midrule[1pt]
\multicolumn{15}{c}{CIFAR-$10$} \\ 
\midrule[0.6pt]
\multicolumn{1}{c|}{ATOL} & 20.60 & 96.03  & 1.45  & 99.59  & 5.20  & 98.78  & 5.00  & 98.76  & 26.05 & 95.03  & 27.55 & \multicolumn{1}{c|}{94.33}  & 14.31 & 97.09  \\
\multicolumn{1}{c|}{ATOL-S} & 21.40 & 94.66  & 1.95  & 99.43  & 2.20  & 99.44  & 2.15  & 99.45  & 23.85 & 93.96  & 30.35 & \multicolumn{1}{c|}{92.84}  & 13.65 & 96.63  \\
\multicolumn{1}{c|}{ATOL-B} & 12.75 & 96.92  & 4.60  & 98.92  & 0.65  & 99.78  & 0.55  & 99.83  & 10.25 & 97.12  & 22.85 & \multicolumn{1}{c|}{94.80}  & \textbf{8.61}  & \textbf{97.90} \\

\midrule[0.6pt]
\multicolumn{15}{c}{CIFAR-$100$} \\ 
\midrule[0.6pt]

\multicolumn{1}{c|}{ATOL} & 70.85 & 84.70  & 13.45  & 97.52  & 51.85  & 90.12  & 55.80  & 89.02  & 63.10 & 83.37  & 75.30 & \multicolumn{1}{c|}{78.86}  & 55.06 & 87.26  \\
\multicolumn{1}{c|}{ATOL-S} & 69.95 & 80.95  & 17.00  & 96.89  & 16.35  & 96.95  & 22.75  & 95.11  & 58.55 & 83.35  & 74.45 & \multicolumn{1}{c|}{77.35}  & 43.18 & 88.43  \\
\multicolumn{1}{c|}{ATOL-B} & 54.65 & 89.69  & 43.95  & 91.27  & 7.80  & 98.57  & 9.60  & 98.13  & 37.45 & 89.51  & 66.90 & \multicolumn{1}{c|}{80.82}  & \textbf{36.72}  & \textbf{91.33} \\ 
\bottomrule[1.5pt]
\end{tabular}}
\end{table}

\subsection{Visualization of Embedding Features} 

\begin{figure}
    \centering
    \subfigure[Mistaken OOD Generation]{
    \label{fig: mistaken tsne}
    \includegraphics[width=0.32\linewidth]{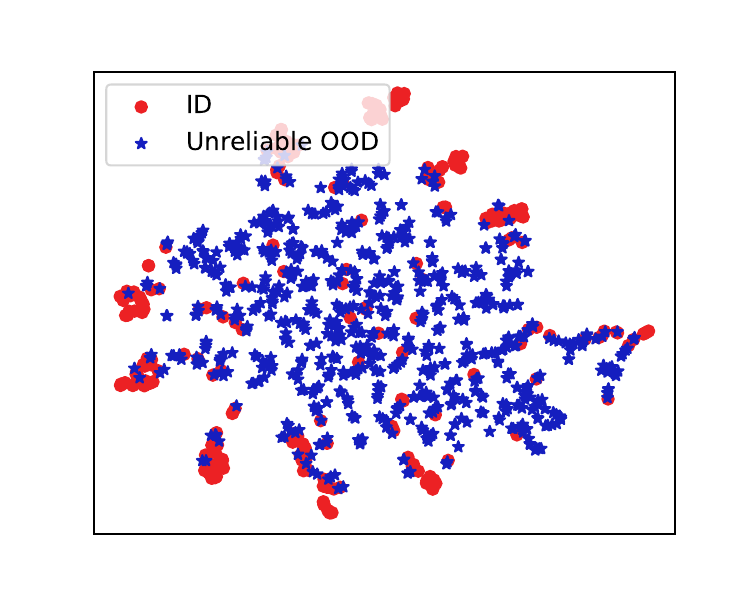}}
    \subfigure[ATOL without Alignment]{			
    \label{fig: before algin}
    \includegraphics[width=0.32\linewidth]{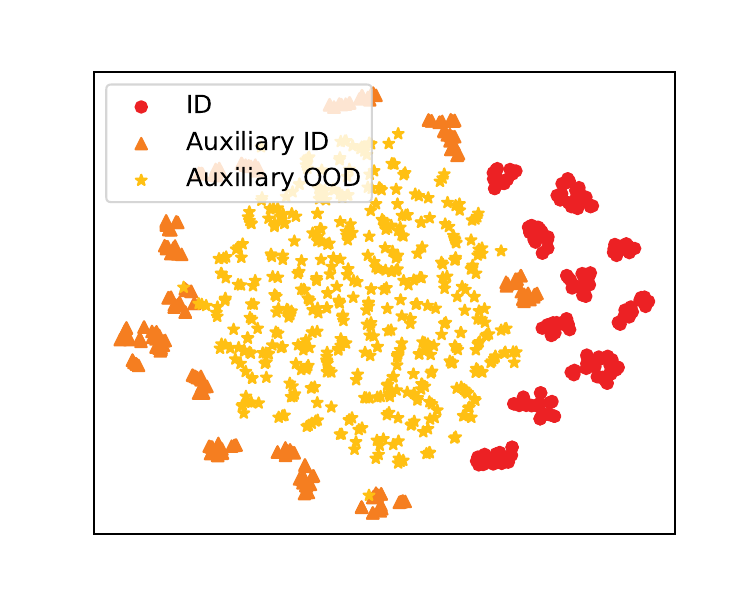}}
    \subfigure[ATOL]{			
    \label{fig: after algin}
    \includegraphics[width=0.32\linewidth]{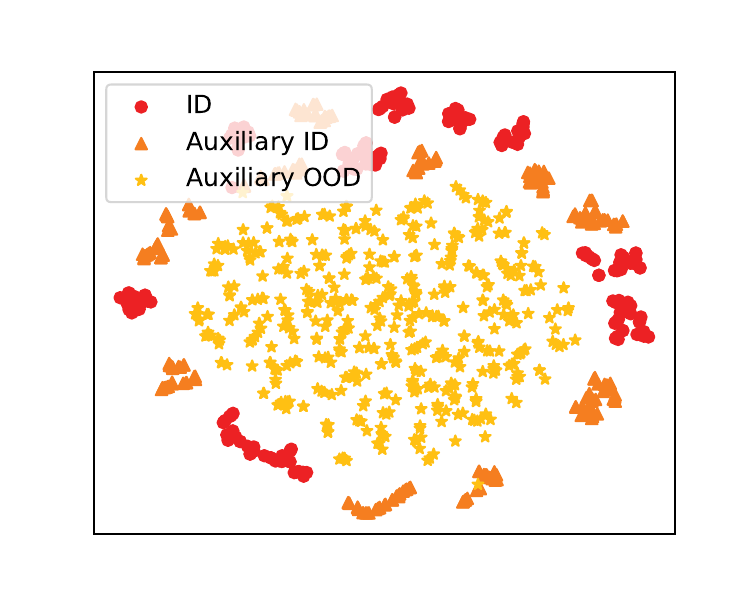}}
    \caption{The t-SNE Visualization of empirical embedding feature distribution of ATOL training on CIFAR-$10$ dataset. The red circle represents the real ID data, the blue star represents the unreliable OOD generated data, the orange triangle represents auxiliary ID data, and the yellow star represents auxiliary OOD data. We qualitatively illustrate the results about mistaken OOD generation, ATOL without alignment and our ATOL.} 
\end{figure}

Qualitatively, to understand how our method helps the predictor to learn, we exploit t-SNE~\cite{van2008visualizing} to show the embedding distributions of real and auxiliary data. The embedding features are extracted from the penultimate layer of a WRN-40-2 model trained on CIFAR-$10$. In Figure~\ref{fig: mistaken tsne}, the mistaken OOD generation leads to an overlap between ID samples and the unreliable OOD samples in the embedding space. In contrast, our ATOL can make the predictor learn to discern the ID and OOD cases as in~\ref{fig: before algin}, relieving the mistaken OOD generation by a large margin. However, the auxiliary ID distribution is far from the real ID distribution, which indicates that the OOD detection capability in the auxiliary task cannot benefit the predictor in real OOD detection. Such a distribution shift results in poor OOD detection performance in the real task (cf. Section~\ref{sec: ablation}). As shown in Figure~\ref{fig: after algin}, we further align the real ID and auxiliary ID distribution in embedding space, where the features of auxiliary ID data are consistent with real ID data. Further, the auxiliary OOD data is observably separate from the real ID data, which proves that ATOL can benefit the predictor learning from the reliable OOD data, thereby providing strong flexibility and generality.

\subsection{Hardware Configurations} 

All experiments are realized by Pytorch $1.11$ with CUDA $12.0$, using machines equipped with NVIDIA Tesla A100 GPUs.

\section{Further Experiments}
\label{app: further}

\subsection{CIFAR Benchmarks}
\label{app: full exp}
In this section, We compare our method with advanced OOD detection methods besides the data generation-based methods, including MSP~\cite{hendrycks2016baseline}, ODIN~\cite{LiangLS18}, Mahalanobis~\cite{lee2018simple}, Free Energy~\cite{energy2020}, MaxLogit~\cite{Hendrycks22maxlogit}, ReAct~\cite{sun2021react}, ViM~\cite{vim2022}, KNN~\cite{knn2022}, 
Watermark~\cite{wang2022watermarking}, ASH~\cite{djurisic2022ash}, BoundaryGAN~\cite{LeeLL2018}, ConfGAN~\cite{Sricharan18robust}, ManifoldGAN~\cite{Vernekar19via},CSI~\cite{Tack20CSI}, G2D~\cite{G2D2021}, CMG~\cite{CMG2022}, LogitNorm~\cite{logitnorm22}, VOS~\cite{du2022vos},   NPOS~\cite{tao2023nonparametric} and CIDER~\cite{ming2022cider}. For clarity, we divide the baseline methods into two categories: OOD scoring functions and OOD training strategies, referring to Appendix~\ref{app: related works}. 

We summarize the main experiments in Table~\ref{tab: cifar10 full}-\ref{tab: cifar100 full} on CIFAR benchmarks for common OOD detection compared with the advanced OOD detection methods. For a fair comparison, all the methods only use ID data without using real OOD datasets. We show that ATOL can achieve superior OOD detection performance on average for the evaluation metrics of FPR$95$ and AUROC, outperforming the competitive rivals by a large margin. Specifically, We incorporate the auxiliary OOD detection task to benefit the predictor learn the OOD knowledge without accessing to the real OOD data, which significantly improve the OOD detection performance. 

Compared with the best baseline KNN+~\cite{knn2022}, ATOL reduces the FPR$95$ from $33.12\%$ to $8.61\%$ on CIFAR-10 and from $51.75\%$ to $8.61\%$ on CIFAR-100. Moreover, for the previous works that adopt similar concepts in synthesize the boundary samples as outliers in the embedding space, i.e., VOS~\cite{du2022vos} and NPOS~\cite{tao2023nonparametric}, our ATOL also reveals better results, with $28.76\%$ and $17.31\%$ improvements on the CIFAR-$10$ dataset and $11.27\%$ and $7.20\%$ improvements on the CIFAR-$100$ dataset w.r.t. FPR$95$. It indicates that our data generation strategy can lead a better OOD learning compared with the synthesizing features strategies. Note that, we can further benefit our ATOL from the latest progress in OOD scoring, improving the performance of our method in OOD detection (cf. Appendix~\ref{app: different scoring}).

\begin{table*}[t]
\caption{Comparison of ATOL and advanced methods on CIFAR-$10$ dataset. All methods are trained on ID data only, without using outlier data. $\downarrow$ (or $\uparrow$) indicates smaller (or larger) values are preferred; a bold font indicates the best results. }\label{tab: cifar10 full}
\vspace{0.5em}
\resizebox{\linewidth}{!}{
\begin{tabular}{c|cccccccccccccc}
\toprule[1.5pt]
\multicolumn{1}{c|}{\multirow{2}{*}{Method}} & \multicolumn{2}{c}{SVHN}   & \multicolumn{2}{c}{LSUN-Crop}         & \multicolumn{2}{c}{LSUN-Resize}       & \multicolumn{2}{c}{iSUN}   & \multicolumn{2}{c}{Texture}& \multicolumn{2}{c|}{Places365}         & \multicolumn{2}{c}{\textbf{Average}}\\ \cline{2-15} 
& FPR95 $\downarrow$ & AUROC $\uparrow$ & FPR95 $\downarrow$ & AUROC $\uparrow$ & FPR95 $\downarrow$ & AUROC $\uparrow$ & FPR95 $\downarrow$ & AUROC $\uparrow$ & FPR95 $\downarrow$ & AUROC $\uparrow$ & FPR95 $\downarrow$ & \multicolumn{1}{c|}{AUROC $\uparrow$} & FPR95 $\downarrow$ & AUROC $\uparrow$ \\ \midrule[1pt]
\multicolumn{15}{c}{OOD Scoring Functions} \\ 
\midrule[0.6pt]
\multicolumn{1}{c|}{MSP} & 49.45 & 91.46  & 26.40 & 96.27  & 51.80 & 91.56  & 54.40 & 90.41  & 59.60 & 88.41  & 58.50 & \multicolumn{1}{c|}{88.32}  & 50.03 & 91.07  \\
\multicolumn{1}{c|}{ODIN}& 29.16 & 91.52  & 12.84 & 96.29  & 32.53 & 90.19  & 42.27 & 89.79  & 44.51 & 88.74  & 34.54 & \multicolumn{1}{c|}{90.25}  & 32.64 & 91.13  \\
\multicolumn{1}{c|}{Mahalanobis}& 13.33 & 97.57  & 39.56 & 94.10  & 43.21 & 93.14  & 43.55 & 92.80  & 15.46 & 97.18  & 68.23 & \multicolumn{1}{c|}{84.69}  & 37.23 & 93.25  \\
\multicolumn{1}{c|}{Free Energy}& 37.00 & 90.29  & 5.85  & 98.93  & 30.35 & 93.88  & 33.00 & 92.61  & 52.15 & 85.72  & 42.15 & \multicolumn{1}{c|}{89.25}  & 33.42 & 91.78  \\
\multicolumn{1}{c|}{MaxLogit}& 36.60 & 90.36  & 6.55  & 98.82  & 30.50 & 93.80  & 33.15 & 92.54  & 51.75 & 85.84  & 42.55 & \multicolumn{1}{c|}{89.21}  & 33.52 & 91.76  \\
\multicolumn{1}{c|}{ReAct}& 50.73 & 86.36  & 6.48  & 98.70  & 29.23 & 94.59  & 36.53 & 92.92  & 59.08 & 85.16  & 42.76 & \multicolumn{1}{c|}{90.24}  & 37.47 & 91.33  \\
\multicolumn{1}{c|}{ViM}& 56.97 & 89.77  & 49.96  & 91.43  & 63.54 & 88.15  & 62.20 & 88.47  & 45.20 & 91.76  & 47.86 & \multicolumn{1}{c|}{91.45}  & 54.29	 & 90.17  \\
\multicolumn{1}{c|}{KNN} & 31.29 & 95.01  & 26.84 & 95.33  & 25.89 & 95.36  & 29.48 & 94.28  & 41.21 & 92.08  & 44.02 & \multicolumn{1}{c|}{90.47}  & 33.12 & 93.76  \\
\multicolumn{1}{c|}{KNN+} & 8.98 & 98.33  & 7.94 & 97.90  & 18.67 & 96.92  & 23.55 & 94.65  & 16.57 & 96.72  & 36.33 & \multicolumn{1}{c|}{93.98}  & 18.67 & 96.42  \\
\multicolumn{1}{c|}{Watermark} & 16.80 & 96.89  & 13.30 & 97.74  & 12.50 & 97.86  & 12.95 & 97.73  & 32.20 & 93.87  & 34.20 & \multicolumn{1}{c|}{93.63}  & 20.33 & 96.29  \\
\multicolumn{1}{c|}{ASH}& 50.73 & 86.36  & 6.48  & 98.70  & 29.23 & 94.59  & 36.53 & 92.92  & 59.08 & 85.16  & 42.76 & \multicolumn{1}{c|}{90.24}  & 37.47 & 91.33  \\
\midrule[0.6pt]
\multicolumn{15}{c}{OOD Training Strategies} \\ 
\midrule[0.6pt]
\multicolumn{1}{c|}{BoundaryGAN}& 86.15 & 79.48  & 19.05 & 96.50  & 41.75 & 92.18  & 46.35 & 90.63  & 70.15 & 78.71  & 70.15 & \multicolumn{1}{c|}{81.25}  & 55.60 & 86.46  \\
\multicolumn{1}{c|}{ConfGAN}& 56.75 & 87.56  & 7.95 & 98.26  & 14.70 & 97.02  & 17.65 & 96.72  & 40.25 & 90.25  & 52.10 & \multicolumn{1}{c|}{88.23}  & 31.57 & 93.01  \\
\multicolumn{1}{c|}{ManifoldGAN}& 26.20 & 94.51  & 5.05 & 98.84  & 27.25 & 95.22  & 30.70 & 93.94  & 32.05 & 91.06  & 38.85 & \multicolumn{1}{c|}{90.95}  & 26.68 & 94.09  \\
\multicolumn{1}{c|}{CSI} & 20.48 & 96.63  & 1.88  & 99.55  & 6.18 & 98.78 & 5.49 & 98.99  & 21.07 & 96.27  & 33.73 & \multicolumn{1}{c|}{93.68}  & 14.80 & 97.31  \\
\multicolumn{1}{c|}{G2D}& 6.10 & 98.63  & 8.30 & 98.41  & 45.35 & 89.48  & 45.25 & 88.96  & 36.80 & 88.63  & 49.20 & \multicolumn{1}{c|}{86.34}  & 31.83 & 91.74 \\
\multicolumn{1}{c|}{LogitNorm} & 6.84 & 98.58  & 1.58 & 99.53  & 18.85 & 96.94  & 20.79 & 96.58  & 26.64 & 94.87  & 30.38 &\multicolumn{1}{c|}{93.85}  & 17.51 & 96.73  \\ 
\multicolumn{1}{c|}{CMG}& 41.70 & 92.48  & 50.35 & 89.35  & 37.80 & 94.24  & 35.60 & 94.75  & 38.70 & 92.12  & 34.95 & \multicolumn{1}{c|}{94.07}  & 39.83 & 92.83  \\
\multicolumn{1}{c|}{VOS} & 46.15 & 93.69  & 3.30  & 99.11  & 41.80 & 94.20  & 48.10 & 93.31  & 57.85 & 88.33  & 61.25 & \multicolumn{1}{c|}{87.54}  & 43.08 & 92.03  \\
\multicolumn{1}{c|}{NPOS} & 36.55 & 93.30  & 9.98  & 98.03  & 21.87 & 95.60  & 28.93 & 94.25  & 52.83 & 85.74  & 39.56 & \multicolumn{1}{c|}{89.71}  & 31.62 & 92.77  \\
\multicolumn{1}{c|}{{CIDER}} & 5.86 & 98.36  & 7.35  & 98.50  & 47.58 & 93.64  & 47.15 & 93.60  & 28.04 & 94.79  & 41.10 & \multicolumn{1}{c|}{91.03}  & 29.51 & 94.99  \\

\midrule[0.6pt]
\multicolumn{1}{c|}{ATOL} & 20.60 & 96.03  & 1.45  & 99.59  & 5.20  & 98.78  & 5.00  & 98.76  & 26.05 & 95.03  & 27.55 & \multicolumn{1}{c|}{94.33}  & 14.31 & 97.09  \\
\multicolumn{1}{c|}{ATOL-S} & 21.40 & 94.66  & 1.95  & 99.43  & 2.20  & 99.44  & 2.15  & 99.45  & 23.85 & 93.96  & 30.35 & \multicolumn{1}{c|}{92.84}  & 13.65 & 96.63  \\
\multicolumn{1}{c|}{ATOL-B} & 12.75 & 96.92  & 4.60  & 98.92  & 0.65  & 99.78  & 0.55  & 99.83  & 10.25 & 97.12  & 22.85 & \multicolumn{1}{c|}{94.80}  & \textbf{8.61}  & \textbf{97.90} \\ 
\bottomrule[1.5pt]
\end{tabular}}
\end{table*}

\begin{table*}[t]
\caption{Comparison of ATOL and advanced methods on CIFAR-$100$ dataset. All methods are trained on ID data only, without using outlier data. $\downarrow$ (or $\uparrow$) indicates smaller (or larger) values are preferred; a bold font indicates the best results. }\label{tab: cifar100 full}
\vspace{0.5em}
\resizebox{\linewidth}{!}{
\begin{tabular}{c|cccccccccccccc}
\toprule[1.5pt]
\multicolumn{1}{c|}{\multirow{2}{*}{Method}} & \multicolumn{2}{c}{SVHN}   & \multicolumn{2}{c}{LSUN-Crop}         & \multicolumn{2}{c}{LSUN-Resize}       & \multicolumn{2}{c}{iSUN}   & \multicolumn{2}{c}{Texture}& \multicolumn{2}{c|}{Places365}         & \multicolumn{2}{c}{\textbf{Average}}\\ \cline{2-15} 
& FPR95 $\downarrow$ & AUROC $\uparrow$ & FPR95 $\downarrow$ & AUROC $\uparrow$ & FPR95 $\downarrow$ & AUROC $\uparrow$ & FPR95 $\downarrow$ & AUROC $\uparrow$ & FPR95 $\downarrow$ & AUROC $\uparrow$ & FPR95 $\downarrow$ & \multicolumn{1}{c|}{AUROC $\uparrow$} & FPR95 $\downarrow$ & AUROC $\uparrow$ \\ \midrule[1pt]
\multicolumn{15}{c}{OOD Scoring Functions} \\ 
\midrule[0.6pt]
\multicolumn{1}{c|}{MSP} & 83.95 & 72.80  & 59.30 & 85.33  & 80.45 & 76.51  & 81.85 & 76.19  & 83.00 & 74.01  & 82.10 & \multicolumn{1}{c|}{74.39}  & 78.44 & 76.54  \\
\multicolumn{1}{c|}{ODIN}& 70.75 & 72.57  & 46.20 & 85.31  & 63.38 & 76.55  & 60.23 & 74.83  & 60.31 & 76.96  & 61.61 & \multicolumn{1}{c|}{77.72}  & 60.41 & 77.32  \\
\multicolumn{1}{c|}{Mahalanobis}& 48.66 & 87.40  & 98.30 & 57.28  & 38.27 & 90.77  & 41.91 & 89.38  & 45.37 & 87.65  & 95.11 & \multicolumn{1}{c|}{65.45}  & 61.26 & 79.65  \\
\multicolumn{1}{c|}{Free Energy}& 85.55 & 74.53  & 23.90  & 95.53  & 77.60 & 80.22  & 80.30 & 79.07  & 79.95 & 76.24  & 79.15 & \multicolumn{1}{c|}{76.37}  & 71.07 & 80.33  \\
\multicolumn{1}{c|}{MaxLogit}& 84.40 & 74.72  & 28.20  & 94.77  & 77.25 & 80.19  & 79.45 & 79.14  & 79.05 & 76.34  & 79.25 & \multicolumn{1}{c|}{76.47}  & 71.27 & 80.27  \\
\multicolumn{1}{c|}{ReAct}& 51.22 & 77.77  & 21.91  & 95.67  & 68.54 & 77.95  & 66.82 & 78.06  & 58.81 & 79.54  & 69.22 & \multicolumn{1}{c|}{76.27}  & 56.09  & 80.88  \\
\multicolumn{1}{c|}{ViM}& 72.32	 & 82.92  & 74.03  & 81.57  & 84.89	 & 77.03  & 84.15 & 76.69  & 33.51	 & 91.72  & 64.17 & \multicolumn{1}{c|}{79.57}  & 68.78  & 81.58  \\
\multicolumn{1}{c|}{KNN} & 42.39 & 92.26  & 59.46 & 79.88  & 59.46 & 88.85  & 59.89 & 87.48  & 48.30 & 88.90  & 81.27 & \multicolumn{1}{c|}{74.82}  & 59.99 & 85.37  \\ 
\multicolumn{1}{c|}{KNN+} & 49.73 & 88.06  & 59.27 & 82.20  & 31.94 & 93.81  & 37.11 & 91.86  & 48.30 & 87.96  & 84.16 & \multicolumn{1}{c|}{71.96}  & 51.75 & 85.98  \\ 
\multicolumn{1}{c|}{Watermark} & 84.95 & 75.04  & 73.15 & 85.74  & 72.95 & 85.79  & 71.95 & 85.47  & 71.95 & 81.82  & 79.25 & \multicolumn{1}{c|}{77.48}  & 75.70 & 81.89  \\ 
\multicolumn{1}{c|}{ASH}& 65.63   & 87.44 & 18.90  & 96.77  & 76.81   & 80.53  & 79.26  & 79.59 & 72.71   & 80.54  & 81.72  & \multicolumn{1}{c|}{74.84}  & 65.84   & 83.29  \\
\midrule[0.6pt]
\multicolumn{15}{c}{OOD Training Strategies} \\ 
\midrule[0.6pt]
\multicolumn{1}{c|}{BoundaryGAN}& 84.50 & 71.00  & 53.55 & 88.38  & 74.80 & 78.87  & 77.90 & 77.60  & 86.00 & 66.80  & 83.55 & \multicolumn{1}{c|}{72.07}  & 76.72 & 75.79  \\
\multicolumn{1}{c|}{ConfGAN}& 88.30 & 72.04  & 39.35 & 92.01  & 77.85 & 80.26  & 79.70 & 79.47  & 79.65 & 71.27  & 84.30 & \multicolumn{1}{c|}{70.99}  & 74.86 & 77.67  \\
\multicolumn{1}{c|}{ManifoldGAN}& 81.65 & 74.51  & 39.95 & 91.52  & 80.20 & 73.16  & 81.80 & 74.07  & 76.35 & 76.31  & 81.30 & \multicolumn{1}{c|}{74.84}  & 73.54 & 77.40  \\
\multicolumn{1}{c|}{CSI} & 62.96 & 84.75  & 61.84  & 85.82  & 96.47 & 49.28  & 95.91 & 52.98  & 78.30 & 71.25  & 85.00 & \multicolumn{1}{c|}{71.45}  & 80.08 & 85.23  \\
\multicolumn{1}{c|}{G2D}& 45.40 & 88.56  & 42.40 & 91.38  & 85.80 & 73.22  & 84.70 & 74.66  & 83.95 & 72.18  & 82.10 & \multicolumn{1}{c|}{74.19}  & 70.73 & 79.03 \\
\multicolumn{1}{c|}{LogitNorm} & 41.37 & 92.78  & 14.57 & 97.20  & 87.96 & 67.88  & 89.15 & 65.74  & 69.66 & 78.87  & 78.00 &\multicolumn{1}{c|}{78.59}  & 63.45 & 80.18  \\ 
\multicolumn{1}{c|}{CMG}& 73.05 & 82.12   & 86.50  & 72.07   & 81.45  & 77.80   & 82.70  & 76.85   & 80.30  & 73.86   & 73.65  & \multicolumn{1}{c|}{82.39 }  & 79.60	 & 77.51  \\
\multicolumn{1}{c|}{VOS} & 78.06 & 82.59  & 40.40 & 92.90  & 83.47 & 70.82  & 85.77 & 70.20  & 82.46 & 77.22  & 82.31 & \multicolumn{1}{c|}{75.47}  & 75.41 & 78.20  \\
\multicolumn{1}{c|}{NPOS} & 66.09 & 87.59  & 31.70 & 95.11  & 63.34 & 80.98  & 62.59 & 80.27  & 74.76 & 80.45  & 77.86 & \multicolumn{1}{c|}{80.65}  & 62.72 & 84.17  \\
\multicolumn{1}{c|}{{CIDER}} & 52.21 & 88.44  & 46.88 & 90.18  & 52.23 & 89.89  & 47.57 & 89.91  & 84.67 & 70.62  & 84.67 & \multicolumn{1}{c|}{71.82}  & 61.37 & 83.48  \\

\midrule[0.6pt]
\multicolumn{1}{c|}{ATOL} & 75.50 & 81.50  & 10.05  & 98.15  & 55.35  & 87.46  & 56.75  & 87.64  & 64.90 & 83.32  & 70.55 & \multicolumn{1}{c|}{79.86}  & 55.52 & 86.32  \\
\multicolumn{1}{c|}{ATOL-S} & 69.95  & 80.95   & 17.00  & 96.89  & 16.35  & 96.95  & 22.75  & 95.11 & 58.55 & 83.35  & 74.45 & \multicolumn{1}{c|}{77.35}  & 43.18 & 88.43  \\
\multicolumn{1}{c|}{ATOL-B} & 54.65 & 89.69  & 43.95  & 91.27  & 7.80  & 98.57  & 9.60  & 98.13  & 37.45 & 89.51  & 66.90 & \multicolumn{1}{c|}{80.82}  & \textbf{36.72}  & \textbf{91.33} \\ 
\bottomrule[1.5pt]
\end{tabular}}
\end{table*}

\subsection{ImageNet Benchmarks}
Table~\ref{tab: imagenet full} lists the detailed experiments on the ImageNet benchmark. The baselines are the same as what we described in Section~\ref{app: full exp}. Our ATOL achieves superior performance on average against all the considered baselines. Further, for the cases with iNaturalist and Places365, which are believed to be the challenging OOD datasets on the ImageNet situation, our ATOL also achieves considerable improvements against all other advanced methods. We highlight that ATOL outperforms the best baseline (i.e., CSI) by $5.55\%$ in FPR$95$, and ATOL is also simpler to use and implement than CSI, which relies on sophisticated data augmentations and ensemble in testing. Overall, it demonstrates that our ATOL can also work well for challenging detection scenarios with extremely large semantic space and complex data patterns. 

\begin{table*}[t]
\centering
\caption{Comparison of ATOL and advanced methods on ImageNet dataset. All methods are trained on ID data only, without using outlier data. $\downarrow$ (or $\uparrow$) indicates smaller (or larger) values are preferred; a bold font indicates the best results. }
\vspace{0.5em}
\label{tab: imagenet full}
\resizebox{0.9\linewidth}{!}{
\begin{tabular}{ccccccccccc}
\toprule[1.5pt]
\multicolumn{1}{c|}{\multirow{2}{*}{Method}} & \multicolumn{2}{c}{iNaturalist} & \multicolumn{2}{c}{SUN}    & \multicolumn{2}{c}{Places365}   & \multicolumn{2}{c|}{Texture}  & \multicolumn{2}{c}{\textbf{Average}}     \\ \cline{2-11} 
\multicolumn{1}{c|}{} & FPR95 $\downarrow$   & AUROC $\uparrow$     & FPR95 $\downarrow$ & AUROC $\uparrow$ & FPR95 $\downarrow$   & AUROC $\uparrow$     & FPR95 $\downarrow$   & \multicolumn{1}{c|}{AUROC $\uparrow$}     & FPR95 $\downarrow$   & AUROC $\uparrow$     \\ \midrule[1pt]
\multicolumn{11}{c}{OOD Scoring Functions}\\ \midrule[0.6pt]
\multicolumn{1}{c|}{MSP}          & 64.25    & 78.49    & 87.47   & 68.61 & 85.56    & 72.29    & 63.60    & \multicolumn{1}{c|}{79.28}    & 75.22    & 74.67    \\
\multicolumn{1}{c|}{ODIN}         & 63.85    & 77.78    & 89.98   & 61.80 & 88.00    & 67.17   & 67.87    & \multicolumn{1}{c|}{77.40}    & 77.43    & 71.04    \\
\multicolumn{1}{c|}{Mahalanobis}  & 95.90    & 60.56    & 95.42   & 45.33 & 98.90    & 44.65    & 55.80    & \multicolumn{1}{c|}{84.60}    & 86.50    & 58.78    \\
\multicolumn{1}{c|}{Free Energy}  & 61.14    & 77.65    & 90.20   & 61.59 & 89.88    & 63.78    & 57.61    & \multicolumn{1}{c|}{77.61}    & 74.71    & 70.16    \\
\multicolumn{1}{c|}{MaxLogit}          & 57.41    & 81.85    & 83.91   & 70.63 & 81.36    & 73.54    & 63.42    & \multicolumn{1}{c|}{76.18}    & 71.53    & 75.55    \\
\multicolumn{1}{c|}{ReAct}        & 55.47    & 81.20    & 66.81   & 82.59 & 63.71    & 85.11    & 46.33    & \multicolumn{1}{c|}{90.30}    & 58.08    & 84.80    \\
\multicolumn{1}{c|}{ViM}        & 85.44    & 77.16    & 89.50   & 75.00 & 87.73    & 78.25    & 38.47   & \multicolumn{1}{c|}{90.55}    & 75.29    & 80.24    \\
\multicolumn{1}{c|}{KNN}          & 65.40    & 83.73    & 75.62   & 77.33 & 79.20    & 74.34    & 40.80    & \multicolumn{1}{c|}{86.45}    & 64.75    & 80.91    \\
\multicolumn{1}{c|}{KNN+}          & 61.48    & 80.35    & 74.04   & 77.33 & 63.53    & 82.95    & 33.48    & \multicolumn{1}{c|}{90.53}    & 58.13    & 82.79    \\
\multicolumn{1}{c|}{Watermark}          & 67.36    & 80.60    & 72.64   & 79.55 & 79.20    & 70.80    & 81.74    & \multicolumn{1}{c|}{83.46}    & 75.24    & 78.61    \\
\multicolumn{1}{c|}{ASH}          & 80.00    & 67.39    & 92.20   & 58.97 & 87.01    & 68.10    & 70.18    & \multicolumn{1}{c|}{69.61}    & 82.35    & 66.02    \\
\midrule[0.6pt]
\multicolumn{11}{c}{OOD Training Strategies} \\ \midrule[0.6pt]
\multicolumn{1}{c|}{BoundaryGAN}          & 83.36    & 68.68    & 90.75   & 63.84 & 87.66    & 67.50    & 80.16    & \multicolumn{1}{c|}{68.12 }    & 85.48    & 66.78    \\
\multicolumn{1}{c|}{ConfGAN}          & 72.67    & 78.29    & 80.73   & 73.88 & 77.40    & 77.24    & 68.74    & \multicolumn{1}{c|}{78.74}    & 74.88    & 77.03    \\
\multicolumn{1}{c|}{ManifoldGAN}          & 72.50    & 78.08    & 84.40   & 72.67 & 82.85    & 74.90    & 50.28    & \multicolumn{1}{c|}{85.28}    & 72.50    & 78.08    \\
\multicolumn{1}{c|}{CSI}          & 64.24    & 85.47    & 53.92   & 95.30 & 58.68    & 93.00    & 52.57    & \multicolumn{1}{c|}{91.13 }    & 57.35    & \textbf{91.22}    \\
\multicolumn{1}{c|}{G2D}         & 73.44    & 78.11    & 81.28   & 74.05 & 77.18    & 77.33    & 67.82    & \multicolumn{1}{c|}{79.17}    & 74.93    & 77.16    \\
\multicolumn{1}{c|}{LogitNorm}          & 76.96    & 77.94    & 75.91   & 79.07 & 72.65    & 81.44    & 71.63    & \multicolumn{1}{c|}{80.15}    & 74.29    & 79.65    \\
\multicolumn{1}{c|}{CMG}         & 71.33    & 78.78    & 80.88   & 73.93 & 75.77    & 77.58    & 63.83    & \multicolumn{1}{c|}{80.57}    & 72.95    & 77.63    \\
\multicolumn{1}{c|}{VOS}          & 87.52    & 72.45    & 83.85   & 74.76 & 79.97    & 77.26    & 72.91    & \multicolumn{1}{c|}{82.71}    & 81.06    & 76.79    \\ 
\multicolumn{1}{c|}{NPOS}          & 74.74    & 77.43    & 83.09   & 73.73 & 78.23    & 76.91    & 56.10    & \multicolumn{1}{c|}{84.37}    & 73.04    & 78.11    \\ 
\multicolumn{1}{c|}{{CIDER}}          & 79.22    & 67.27    & 84.82   & 74.34  & 77.39    & 78.77    & 19.80    & \multicolumn{1}{c|}{95.16}    & 65.31    & 78.89    \\ 
\midrule[0.6pt]
\multicolumn{1}{c|}{ATOL}         & 60.98 & 79.53 & 73.90   & 79.97 & 58.48 & 86.97 & 13.85 & \multicolumn{1}{c|}{96.80} & \textbf{51.80} & 85.82 \\ 
\bottomrule[1.5pt]
\end{tabular}}
\end{table*}

\subsection{Hard OOD Detection}
\label{sec: nearood}
Besides the above test OOD datasets, we also consider hard OOD scenarios~\cite{Tack20CSI}, of which the test OOD data are very similar to that of the ID cases in style. Following the common setup~\cite{knn2022} with the CIFAR-$10$ dataset being the ID case, we evaluate our ATOL on three hard OOD datasets, namely, LSUN-Fix~\cite{yu2015lsun}, ImageNet-Fix~\cite{deng2009imagenet}, and CIFAR-$100$. We compare our ATOL with several works reported performing well in hard OOD detection, including KNN~\cite{knn2022}, ASH~\cite{djurisic2022ash} and CSI~\cite{Tack20CSI}, where the results are summarized in Table~\ref{tab: cifar hard}. As we can see, our ATOL can beat these advanced methods across all the considered datasets, even for the challenging CIFAR-$10$ vs. CIFAR-$100$ setting. 

\begin{table}[htbp]
\caption{Comparison of ATOL and advanced methods in hard OOD detection. $\downarrow$ (or $\uparrow$) indicates smaller (or larger) values are preferred; a bold font indicates the best results in a column.} \label{tab: cifar hard}
\centering
\small
\begin{tabular}{c|cc|cc|cc}
\specialrule{0em}{1pt}{1pt}
\toprule[1.2pt]
\multirow{2}{*}{Methods} & \multicolumn{2}{c|}{LSUN-Fix}& \multicolumn{2}{c|}{ImageNet-Fix}       & \multicolumn{2}{c}{CIFAR-100}\\ 
\cline{2-7}\specialrule{0em}{1pt}{0pt}
& FPR$95$ $\downarrow$ & AUROC $\uparrow$ & FPR$95$ $\downarrow$ & AUROC $\uparrow$ & FPR$95$ $\downarrow$ & AUROC $\uparrow$ \\ 
\midrule[0.6pt]
CSI & 39.79 & 93.63 & 37.47 & 93.93 & 45.64 & 87.64 \\
KNN & 42.01 & 91.98 & 40.41 & 92.33 & 49.22 & 89.91 \\
ASH& 45.12 & 89.72 & 42.56 & 89.99 & 50.45 & 87.09 \\ 
\midrule[0.6pt]
ATOL & \textbf{22.25} & \textbf{95.89} & \textbf{24.20} & \textbf{94.99} & \textbf{42.80} & \textbf{91.19}\\ 
\bottomrule[1.2pt]
\end{tabular}
\end{table}

\subsection{Comparison of ATOL and Outlier Exposure}
This section compares ATOL and OE on CIFAR and ImageNet benchmarks in Tables~\ref{tab: oe cifar}-\ref{tab: oe imagenet}. As we can see, our ATOL shows comparable performance with outlier exposure, meanwhile eliminating the reliance on real OOD data. Such outstanding performances are sufficient to verify that ATOL can make the predictor learn from the auxiliary OOD detection task as effectively as learning from the real OOD data. Note that, on the ImageNet benchmark, OE reveals inferior performance compared to ATOL since the real OOD data collected from the realistic scenarios inevitably overlap with the large-scale ID data to some extent, while getting pure and efficient OOD data is labor-intensive and inflexible. The results demonstrate the drawbacks of the methods using real OOD data. In contrast, our ATOL does not suffer from this issue and can benefit from the auxiliary OOD detection task for real OOD detection. Moreover, the better results of our ATOL over OE verify the effectiveness of OOD learning only based on ID data, which can draw more attention from the community toward OOD learning with ID data.

\begin{table}[htbp]
\centering
\caption{Comparison of ATOL and Outlier Exposure on CIFAR-$10$ and CIFAR-$100$. $\downarrow$ (or $\uparrow$) indicates smaller (or larger) values are preferred; a bold font indicates the best results in a column.}
\label{tab: oe cifar}
\resizebox{\linewidth}{!}{
\begin{tabular}{ccccccccccccccc}
\toprule[1.5pt]
\multicolumn{1}{c|}{\multirow{2}{*}{Scores}} & \multicolumn{2}{c}{SVHN}              & \multicolumn{2}{c}{LSUN-Crop}         & \multicolumn{2}{c}{LSUN-Resize}       & \multicolumn{2}{c}{iSUN}              & \multicolumn{2}{c}{Texture}           & \multicolumn{2}{c|}{Places365}                             & \multicolumn{2}{c}{\textbf{Average}}                                       \\ \cline{2-15} 
\multicolumn{1}{c|}{}                        & FPR95 $\downarrow$ & AUROC $\uparrow$ & FPR95 $\downarrow$ & AUROC $\uparrow$ & FPR95 $\downarrow$ & AUROC $\uparrow$ & FPR95 $\downarrow$ & AUROC $\uparrow$ & FPR95 $\downarrow$ & AUROC $\uparrow$ & FPR95 $\downarrow$ & \multicolumn{1}{c|}{AUROC $\uparrow$} & FPR95 $\downarrow$              & AUROC $\uparrow$                \\ \midrule[1pt]
\multicolumn{15}{c}{CIFAR-10} \\ \midrule[1pt]
\multicolumn{1}{c|}{ATOL} & 20.60 & 96.03  & 1.45  & 99.59  & 5.20  & 98.78  & 5.00  & 98.76  & 26.05 & 95.03  & 27.55 & \multicolumn{1}{c|}{94.33}  & 14.31 & 97.09  \\
\multicolumn{1}{c|}{ATOL-S} & 21.40 & 94.66  & 1.95  & 99.43  & 2.20  & 99.44  & 2.15  & 99.45  & 23.85 & 93.96  & 30.35 & \multicolumn{1}{c|}{92.84}  & 13.65 & 96.63  \\
\multicolumn{1}{c|}{ATOL-B} & 12.75 & 96.92  & 4.60  & 98.92  & 0.65  & 99.78  & 0.55  & 99.83  & 10.25 & 97.12  & 22.85 & \multicolumn{1}{c|}{94.80}  & \textbf{8.61}  & \textbf{97.90} \\
\multicolumn{1}{c|}{\cellcolor{greyL}OE}  & \cellcolor{greyL}20.35 & \cellcolor{greyL}96.75  & \cellcolor{greyL}2.20  & \cellcolor{greyL}99.52  & \cellcolor{greyL}0.95  & \cellcolor{greyL}99.72  & \cellcolor{greyL}0.85  & \cellcolor{greyL}99.76  & \cellcolor{greyL}17.15 & \cellcolor{greyL}96.79  & \cellcolor{greyL}23.05 & \multicolumn{1}{c|}{\cellcolor{greyL}{95.64}}  & \cellcolor{greyL}10.76 & \cellcolor{greyL}97.03  \\
\midrule[1pt]
\multicolumn{15}{c}{CIFAR-100}    \\ \midrule[1pt]
\multicolumn{1}{c|}{ATOL} & 75.50 & 81.50  & 10.05  & 98.15  & 55.35  & 87.46  & 56.75  & 87.64  & 64.90 & 83.32  & 70.55 & \multicolumn{1}{c|}{79.86}  & 55.52 & 86.32  \\
\multicolumn{1}{c|}{ATOL-S} & 69.95  & 80.95   & 17.00  & 96.89  & 16.35  & 96.95  & 22.75  & 95.11 & 58.55 & 83.35  & 74.45 & \multicolumn{1}{c|}{77.35}  & 43.18 & 88.43  \\
\multicolumn{1}{c|}{ATOL-B} & 54.65 & 89.69  & 43.95  & 91.27  & 7.80  & 98.57  & 9.60  & 98.13  & 37.45 & 89.51  & 66.90 & \multicolumn{1}{c|}{80.82}  & \textbf{36.72}  & \textbf{91.33} \\  
\multicolumn{1}{c|}{\cellcolor{greyL}OE}  & \cellcolor{greyL}75.10 & \cellcolor{greyL}80.69  & \cellcolor{greyL}24.95  & \cellcolor{greyL}95.11  & \cellcolor{greyL}20.05  & \cellcolor{greyL}95.36  & \cellcolor{greyL}25.45  & \cellcolor{greyL}93.94  & \cellcolor{greyL}48.15 & \cellcolor{greyL}87.94  & \cellcolor{greyL}43.25 & \multicolumn{1}{c|}{\cellcolor{greyL}{88.55}}  & \cellcolor{greyL}39.49 & \cellcolor{greyL}90.26  \\
\bottomrule[1.5pt]
\end{tabular}}
\end{table}

\begin{table}[htbp]
\centering
\caption{Comparison of ATOL and Outlier Exposure on ImageNet. $\downarrow$ (or $\uparrow$) indicates smaller (or larger) values are preferred; a bold font indicates the best results in a column.}
\label{tab: oe imagenet}
\resizebox{0.9\linewidth}{!}{
\begin{tabular}{ccccccccccc}
\toprule[1.5pt]
\multicolumn{1}{c|}{\multirow{2}{*}{Scores}} & \multicolumn{2}{c}{iNaturalist} & \multicolumn{2}{c}{SUN}    & \multicolumn{2}{c}{Places365}   & \multicolumn{2}{c|}{Texture}  & \multicolumn{2}{c}{\textbf{Average}}     \\ \cline{2-11} 
\multicolumn{1}{c|}{} & FPR95 $\downarrow$   & AUROC $\uparrow$     & FPR95 $\downarrow$ & AUROC $\uparrow$ & FPR95 $\downarrow$   & AUROC $\uparrow$     & FPR95 $\downarrow$   & \multicolumn{1}{c|}{AUROC $\uparrow$}     & FPR95 $\downarrow$   & AUROC $\uparrow$     \\ \midrule[1pt]
\multicolumn{1}{c|}{ATOL}         & 60.98 & 79.53 & 73.90   & 79.97 & 58.48 & 86.97 & 13.85 & \multicolumn{1}{c|}{96.80} & \textbf{51.80} & \textbf{85.82} \\ 
\multicolumn{1}{c|}{\cellcolor{greyL}OE} & \cellcolor{greyL}78.31    & \cellcolor{greyL}75.23    & \cellcolor{greyL}80.10   & \cellcolor{greyL}76.55 & \cellcolor{greyL}70.41    & \cellcolor{greyL}81.78    & \cellcolor{greyL}66.38    & \multicolumn{1}{c|}{\cellcolor{greyL}82.04}    & \cellcolor{greyL}73.80    & \cellcolor{greyL}78.90    \\
\bottomrule[1.5pt]
\end{tabular}}
\end{table}

\subsection{Model Trained from Scratch}
In this section, we provide the implementation details and the experimental results for ATOL trained from scratch. We train the WRN-40-2 model on CIFAR-$10$. For training, ATOL is run for 100 epochs with an initial learning rate of 0.1 and cosine decay. The batch size is $64$ for ID cases and $256$ for OOD cases. Going beyond fine-tuning with the pre-trained model, we show that ATOL is also applicable and effective when training from scratch. Here, we further introduce the \emph{area under the precision-recall curve} (AUPR) to evaluate the OOD detection performance. The table~\ref{tab: cifar10 scratch} showcases the performance of ATOL trained on the CIFAR-$10$ dataset, where the promising performance of ATOL still holds.
\begin{table}[htbp]
\centering
\caption{Performance of ATOL from scratch on CIFAR-$10$}
\label{tab: cifar10 scratch}
\small
\begin{tabular}{c|ccc}
\toprule[1.2pt]
ATOL from scratch & FPR$95$ $\downarrow$ & AUROC $\uparrow$ & AUPR $\uparrow$ \\ \midrule[0.6pt]
SVHN             & 18.85      & 96.63      & 99.30     \\
LSUN-Crop           & 2.20      & 99.51      & 99.90     \\
LSUN-Resize           & 6.70      & 97.88      & 99.56     \\
iSUN             & 7.25      & 97.83      & 99.55     \\
Texture          & 26.40      & 94.77      & 98.62     \\
Places365        & 30.35      & 92.53      & 97.95     \\ \midrule[0.6pt]
Average          & 15.29      & 96.52      & 99.15     \\ 
\bottomrule[1.2pt]
\end{tabular}
\end{table}

\subsection{ATOL with different scoring functions}
\label{app: different scoring}
To further verify the generality and the effectiveness of ATOL, we test ATOL with three representative OOD scoring functions, namely, MSP~\cite{hendrycks2016baseline}, Free energy~\cite{energy2020}, and MaxLogit~\cite{Hendrycks22maxlogit}. Regarding all the cases with different scoring functions, our ATOL always achieves good performance, demonstrating that our proposal can genuinely make the predictor learn from OOD knowledge for OOD detection. Further, comparing the results across different scoring strategies, we observe that using the MaxLogit scoring leads to better results than the MSP scoring. Therefore, we choose the MaxLogit in ATOL.

\begin{table}[htbp]
\centering
\caption{Performance with different OOD scoring functions on CIFAR-$10$ and CIFAR-$100$}
\label{tab: score}
\resizebox{\linewidth}{!}{
\begin{tabular}{ccccccccccccccc}
\toprule[1.5pt]
\multicolumn{1}{c|}{\multirow{2}{*}{Scores}} & \multicolumn{2}{c}{SVHN}              & \multicolumn{2}{c}{LSUN-Crop}         & \multicolumn{2}{c}{LSUN-Resize}       & \multicolumn{2}{c}{iSUN}              & \multicolumn{2}{c}{Texture}           & \multicolumn{2}{c|}{Places365}                             & \multicolumn{2}{c}{\textbf{Average}}                                       \\ \cline{2-15} 
\multicolumn{1}{c|}{}                        & FPR95 $\downarrow$ & AUROC $\uparrow$ & FPR95 $\downarrow$ & AUROC $\uparrow$ & FPR95 $\downarrow$ & AUROC $\uparrow$ & FPR95 $\downarrow$ & AUROC $\uparrow$ & FPR95 $\downarrow$ & AUROC $\uparrow$ & FPR95 $\downarrow$ & \multicolumn{1}{c|}{AUROC $\uparrow$} & FPR95 $\downarrow$              & AUROC $\uparrow$                \\ \midrule[1pt]
\multicolumn{15}{c}{CIFAR-10} \\ \midrule[1pt]
\multicolumn{1}{c|}{MSP} & 29.10              & 94.57            & 1.45               & 99.58            & 6.95               & 98.60            & 6.75               & 98.63            & 22.45              & 94.78            & 27.30              & \multicolumn{1}{c|}{93.27}            & 15.67       & 96.57       \\
\multicolumn{1}{c|}{Energy}                  & 16.45              & 96.27            & 1.25               & 99.60            & 3.35               & 99.29            & 3.70               & 99.22            & 26.00              & 93.75            & 35.70              & \multicolumn{1}{c|}{92.14}            & 14.41       & 96.71       \\
\multicolumn{1}{c|}{MaxLogit}                & 21.15              & 95.97            & 1.20               & 99.61            & 4.40               & 98.87            & 5.40               & 98.75            & 24.75              & 95.18            & 25.40              & \multicolumn{1}{c|}{94.74}            & \textbf{13.72} & \textbf{97.19} \\ \midrule[1pt]
\multicolumn{15}{c}{CIFAR-100}    \\ \midrule[1pt]
\multicolumn{1}{c|}{MSP} & 69.70              & 83.88            & 16.75              & 96.77            & 56.60              & 88.55            & 58.95              & 87.55            & 64.60              & 82.14            & 77.00              & \multicolumn{1}{c|}{78.09}            & 57.27       & 86.16       \\
\multicolumn{1}{c|}{Energy}                  & 36.00              & 94.05            & 8.85               & 98.12            & 38.15              & 89.07            & 42.25              & 87.77            & 66.65              & 81.37            & 75.40              & \multicolumn{1}{c|}{77.01}            & \textbf{44.55}       & \textbf{87.90}       \\
\multicolumn{1}{c|}{MaxLogit}                & 70.85              & 84.70            & 13.45              & 97.52            & 51.85              & 90.12            & 55.80              & 89.02            & 63.10              & 83.37            & 75.30              & \multicolumn{1}{c|}{78.86}            & 55.06       & 87.26       \\ \bottomrule[1.5pt]
\end{tabular}}
\end{table}

\begin{table}[htbp]
\centering
\caption{Performance with different OOD scoring functions on ImageNet. }
\label{tab: imagenet score}
\resizebox{0.9\linewidth}{!}{
\begin{tabular}{ccccccccccc}
\toprule[1.5pt]
\multicolumn{1}{c|}{\multirow{2}{*}{Scores}} & \multicolumn{2}{c}{iNaturalist} & \multicolumn{2}{c}{SUN}    & \multicolumn{2}{c}{Places365}   & \multicolumn{2}{c|}{Texture}  & \multicolumn{2}{c}{\textbf{Average}}     \\ \cline{2-11} 
\multicolumn{1}{c|}{} & FPR95 $\downarrow$   & AUROC $\uparrow$     & FPR95 $\downarrow$ & AUROC $\uparrow$ & FPR95 $\downarrow$   & AUROC $\uparrow$     & FPR95 $\downarrow$   & \multicolumn{1}{c|}{AUROC $\uparrow$}     & FPR95 $\downarrow$   & AUROC $\uparrow$     \\ \midrule[1pt]
\multicolumn{1}{c|}{MSP}          & 61.16    & 80.38    & 73.37   & 77.65 & 61.73    & 83.63    & 22.91    & \multicolumn{1}{c|}{93.55}    & 54.79    & 83.80    \\
\multicolumn{1}{c|}{Energy}         & 62.15    & 80.49    & 72.13   & 82.15 & 62.69    & 85.78   & 31.10    & \multicolumn{1}{c|}{92.73}    & 57.02    & 85.29    \\
\multicolumn{1}{c|}{MaxLogit}         & 60.98 & 79.53 & 73.90   & 79.97 & 58.48 & 86.97 & 13.85 & \multicolumn{1}{c|}{96.80} & \textbf{51.80} & \textbf{85.82} \\ 
\bottomrule[1.5pt]
\end{tabular}}
\end{table}

\subsection{{{Performance and Efficiency Comparison on Advanced Approaches}}}
\label{app: efficient}
{In the main context, our primary focus is on comparing data generation-based approaches. Moreover, we also compare the training time for a set of representative methods on CIFAR-$100$ (similar results on CIFAR-$10$), summarizing the results in the following table. As we can see, our method demonstrates promising performance improvement over other methods with acceptable computational resources.}

\begin{table}[htbp]
\centering
\caption{Performance and efficiency comparison with advanced approaches on CIFAR-$100$. We report the per epoch training time (measured by seconds)}
\label{tab: cifar100 efficiency}
\small
\begin{tabular}{c|cc|c}
\toprule[1.2pt]
Methods & FPR$95$ $\downarrow$ & AUROC $\uparrow$ & Training Time (s)$\downarrow$ \\ \midrule[0.6pt]
CSI             & 80.08      & 85.23      & 98.88     \\
LogitNorm           & 63.45      & 80.18      & \textbf{25.16}     \\
VOS           & 75.41      & 78.20      & 38.97     \\
NPOS             & 62.72      & 84.17      & 61.54     \\
ATOL          & 55.22      & 87.24      & 58.33     \\
ATOL-S        & 43.18      & 88.43      & 70.28     \\ 
ATOL-B          & 	\textbf{36.72}      & \textbf{91.33}      & 65.33     \\ 
\bottomrule[1.2pt]
\end{tabular}
\end{table}

\subsection{Mean and Standard Deviation}

This section demonstrates the effectiveness of our ATOL by validating the experiments using five individual trails (random seeds) on the CIFAR benchmarks. Along with the individual findings, we also summarize the mean performance and standard deviation for all of the trials for CIFAR-10, CIFAR-100 and ImageNet. The experimental results are summarized in Tables~\ref{tab: cifar mean+std}-\ref{tab: imagenet mean+std}. As we can see, our ATOL can result in better OOD detection performance and more stable performance over various ID dataset options.

\begin{table}[t]
\centering
\caption{Performance of ATOL on CIFAR with 5 individual trails. $\downarrow$ (or $\uparrow$) indicates smaller (or larger) values are preferred; and a bold font indicates the best results in the corresponding column. } \label{tab: cifar mean+std}
\resizebox{\linewidth}{!}{
\begin{tabular}{c|cccccccccccc|cc}
\toprule[1.5pt]
\multicolumn{1}{c|}{\multirow{2}{*}{Scores}} & \multicolumn{2}{c}{SVHN}              & \multicolumn{2}{c}{LSUN-Crop}         & \multicolumn{2}{c}{LSUN-Resize}       & \multicolumn{2}{c}{iSUN}              & \multicolumn{2}{c}{Texture}           & \multicolumn{2}{c|}{Places365}                             & \multicolumn{2}{c}{\textbf{Average}}                                       \\ \cline{2-15} 
\multicolumn{1}{c|}{}                        & FPR95 $\downarrow$ & AUROC $\uparrow$ & FPR95 $\downarrow$ & AUROC $\uparrow$ & FPR95 $\downarrow$ & AUROC $\uparrow$ & FPR95 $\downarrow$ & AUROC $\uparrow$ & FPR95 $\downarrow$ & AUROC $\uparrow$ & FPR95 $\downarrow$ & \multicolumn{1}{c|}{AUROC $\uparrow$} & FPR95 $\downarrow$              & AUROC $\uparrow$                \\ \midrule[1pt]
\multicolumn{15}{c}{CIFAR-10} \\ \midrule[1pt]
\#1        & 21.55 & 96.10 & 0.95 & 99.62 & 5.00 & 98.79 & 4.55 & 98.90 & 26.30 & 95.04 & 28.90 & 93.60 & 14.54 & 97.01  \\
\#2        & 21.65 & 95.74 & 1.15 & 99.60 & 5.40 & 98.79 & 3.80 & 98.93 & 26.25 & 94.91 & 27.75 & 94.19 & 14.33 & 97.03  \\
\#3        & 18.05 & 96.27 & 1.45 & 99.58 & 4.35 & 98.88 & 4.45 & 98.86 & 20.90 & 95.70 & 25.85 & 94.39 & 12.51 & 97.28 \\
\#4        & 20.85 & 96.11 & 1.60 & 99.59 & 5.05 & 98.76 & 5.75  & 98.73 & 26.50 & 94.80 & 28.20 & 94.32  & 14.66 & 97.05 \\
\#5        & 21.40 & 94.66 & 1.95 & 99.43 & 2.20 & 99.44 & 2.15 & 99.45 & 23.85 & 93.96 & 30.35 & 92.84 & 13.65 & 96.63 \\
\makecell{mean \\ $\pm$ std} 
           & \makecell{\textbf{20.70} \\ $\pm$ \textbf{1.35}} & \makecell{\textbf{95.77} \\ $\pm$ \textbf{0.58}} & \makecell{\textbf{1.42} \\ $\pm$ \textbf{0.34}} & \makecell{\textbf{99.56} \\ $\pm$ \textbf{0.06}} & \makecell{\textbf{4.4} \\ $\pm$ \textbf{1.15}} & \makecell{\textbf{98.93} \\ $\pm$ \textbf{0.25}} & \makecell{\textbf{4.14} \\ $\pm$ \textbf{1.17}} & \makecell{\textbf{98.97} \\ $\pm$ \textbf{0.24}} & \makecell{\textbf{24.75} \\ $\pm$ \textbf{2.16}} & \makecell{\textbf{94.88} \\ $\pm$ \textbf{0.55}} & \makecell{\textbf{28.21} \\ $\pm$ \textbf{ 1.47}} & \makecell{\textbf{93.86} \\ $\pm$ \textbf{0.58}} & \makecell{\textbf{13.93} \\ $\pm$ \textbf{0.79}} & \makecell{\textbf{97.00} \\ $\pm$ \textbf{0.20}}  \\

\midrule[1pt]
\multicolumn{15}{c}{CIFAR-100}    \\ \midrule[1pt]
\#1        & 68.45 & 85.62 & 14.10 & 97.41 & 54.70 & 88.63 & 59.55 & 87.45 & 61.60 & 83.56 & 74.70 & 78.76  & 55.52 & 86.90\\
\#2        & 70.05 & 86.34 & 14.90 & 97.27 & 51.60 & 89.37 & 55.55 & 88.34 & 63.30 & 83.37 & 75.95 & 78.76  & 55.22 & 87.24 \\
\#3        & 70.85 & 84.70 & 13.45 & 97.52 & 51.85 & 90.12 & 55.80 & 89.02 & 63.10 & 83.37 & 75.30 & 78.86  & 55.06 & 87.26\\
\#4        & 65.55 & 86.13 & 14.55 & 97.37 & 53.00 & 89.31 & 55.65 & 88.81 & 67.50 & 82.39 & 75.35 & 77.76  & 55.27 & 86.96\\
\#5        & 69.45 & 84.80 & 14.35 & 97.67 & 55.45 & 89.08 & 58.10 & 88.30 & 61.70 & 84.64 & 74.70 & 78.16  & 55.63 & 87.11\\
\hline
\makecell{mean \\ $\pm$ std} 
           & \makecell{\textbf{68.86} \\ $\pm$ \textbf{1.83}} & \makecell{\textbf{85.52} \\ $\pm$ \textbf{0.67}} & \makecell{\textbf{14.27} \\ $\pm$ \textbf{0.48}} & \makecell{\textbf{97.44} \\ $\pm$ \textbf{0.13}} & \makecell{\textbf{53.32} \\ $\pm$ \textbf{1.52}} & \makecell{\textbf{89.30} \\ $\pm$ \textbf{0.48}} & \makecell{\textbf{56.92} \\ $\pm$ \textbf{1.61}} & \makecell{\textbf{88.38} \\ $\pm$ \textbf{0.54}} & \makecell{\textbf{63.44} \\ $\pm$ \textbf{2.14}} & \makecell{\textbf{83.46} \\ $\pm$ \textbf{0.71}} & \makecell{\textbf{75.19} \\ $\pm$ \textbf{0.46}} & \makecell{\textbf{78.46} \\ $\pm$ \textbf{0.42}} & \makecell{\textbf{55.34} \\ $\pm$ \textbf{0.20}} & \makecell{\textbf{87.09} \\ $\pm$ \textbf{0.14}}  \\
\bottomrule[1.5pt]
\end{tabular}}
\end{table}

\begin{table}[t]
\centering
\caption{Performance of ATOL on ImageNet with 5 individual trails. $\downarrow$ (or $\uparrow$) indicates smaller (or larger) values are preferred; and a bold font indicates the best results in the corresponding column. }
\label{tab: imagenet mean+std}
\resizebox{0.9\linewidth}{!}{
\begin{tabular}{c|cccccccc|cc}
\toprule[1.5pt]
\multicolumn{1}{c|}{\multirow{2}{*}{Scores}} & \multicolumn{2}{c}{iNaturalist} & \multicolumn{2}{c}{SUN}    & \multicolumn{2}{c}{Places365}   & \multicolumn{2}{c|}{Texture}  & \multicolumn{2}{c}{\textbf{Average}}     \\ \cline{2-11} 
\multicolumn{1}{c|}{} & FPR95 $\downarrow$   & AUROC $\uparrow$     & FPR95 $\downarrow$ & AUROC $\uparrow$ & FPR95 $\downarrow$   & AUROC $\uparrow$     & FPR95 $\downarrow$   & \multicolumn{1}{c|}{AUROC $\uparrow$}     & FPR95 $\downarrow$   & AUROC $\uparrow$     \\ \midrule[1pt]
\#1        & 60.98 & 79.53 & 73.90 & 79.97 & 58.48 & 86.97 & 13.85 & 96.80 & 51.80 & 85.82 \\
\#2        & 62.87 & 78.82 & 73.95 & 80.48 & 58.67 & 87.14 & 13.95 & 96.79 & 52.36 & 85.80 \\
\#3        & 61.50 & 78.99 & 73.53 & 80.28 & 58.73  & 87.09  & 13.95  & 96.79  & 51.93  & 85.79 \\
\#4        & 58.07 & 80.10 & 72.32 & 81.09 & 59.41 & 86.70  & 15.20 & 96.56 & 51.25 & 86.12  \\
\#5        & 61.70 & 78.77 & 73.06 & 80.11  & 58.80 & 87.20 & 13.95 & 96.79 & 51.88 & 85.72 \\
\hline
\makecell{mean \\ $\pm$ std} 
           & \makecell{\textbf{61.02} \\ $\pm$ \textbf{1.60}} & \makecell{\textbf{79.24} \\ $\pm$ \textbf{0.50}} & \makecell{\textbf{73.35} \\ $\pm$ \textbf{0.60}} & \makecell{\textbf{80.38} \\ $\pm$ \textbf{0.39}} & \makecell{\textbf{58.81} \\ $\pm$ \textbf{0.31}} & \makecell{\textbf{87.02} \\ $\pm$ \textbf{0.17}} & \makecell{\textbf{14.18} \\ $\pm$ \textbf{0.51}} & \makecell{\textbf{96.74} \\ $\pm$ \textbf{0.09}} & \makecell{\textbf{51.84} \\ $\pm$ \textbf{0.35}} & \makecell{\textbf{85.85} \\ $\pm$ \textbf{0.13}} \\
\bottomrule[1.5pt]
\end{tabular}}
\end{table}

\subsection{Visualization of generated images}
\label{app: generated images}
\begin{figure}[t]
    \centering    
    \subfigure{				
    \includegraphics[width=0.18\textwidth]{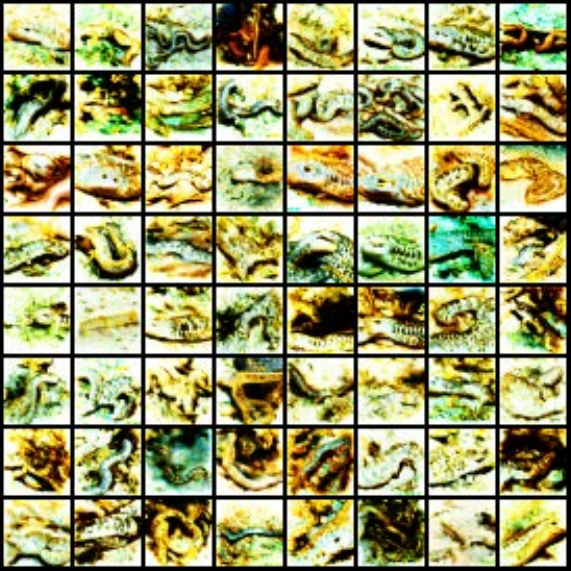}}
    \subfigure{				
    \includegraphics[width=0.18\textwidth]{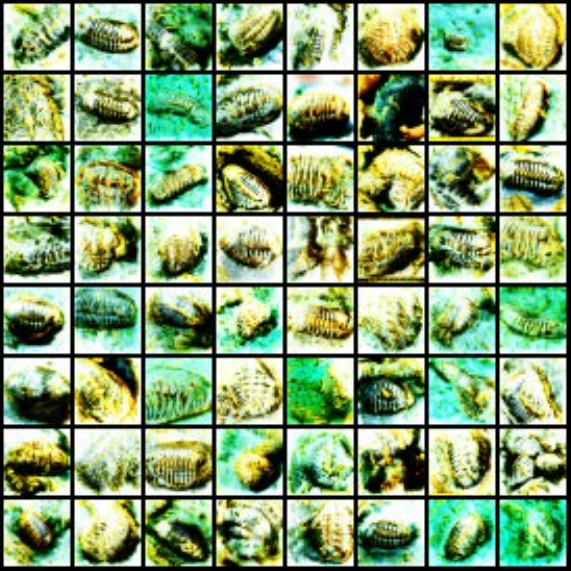}}
    \subfigure{				
    \includegraphics[width=0.18\textwidth]{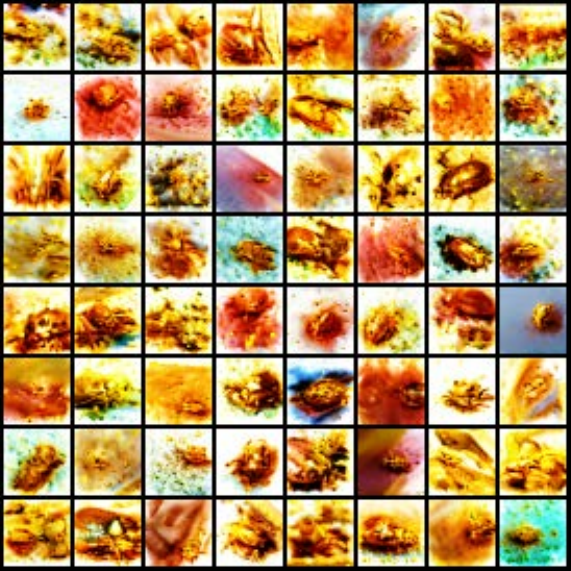}}
    \subfigure{				
    \includegraphics[width=0.18\textwidth]{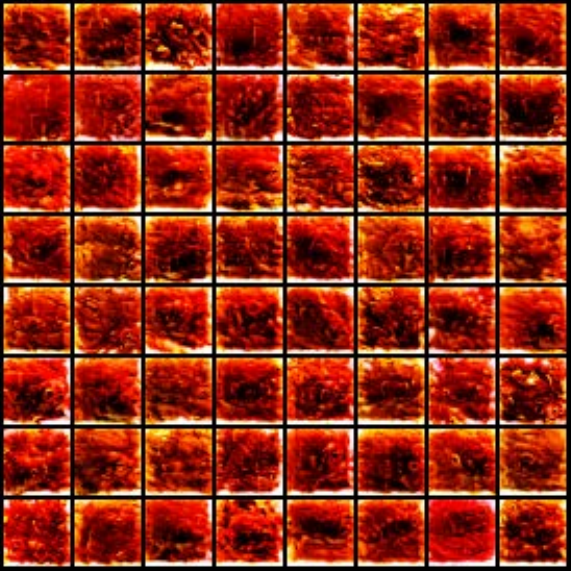}}
    \subfigure{				
    \includegraphics[width=0.18\textwidth]{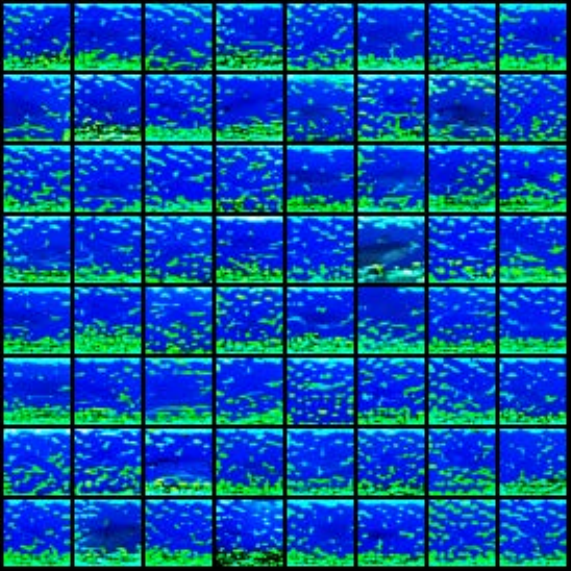}}
    \subfigure{				
    \includegraphics[width=0.18\textwidth]{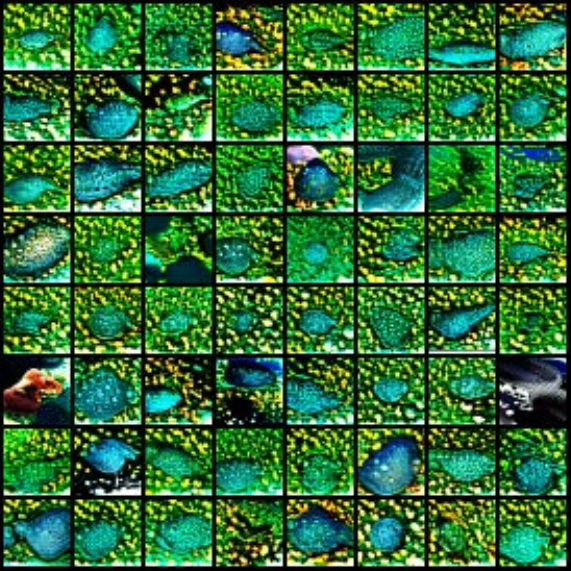}}
    \subfigure{				
    \includegraphics[width=0.18\textwidth]{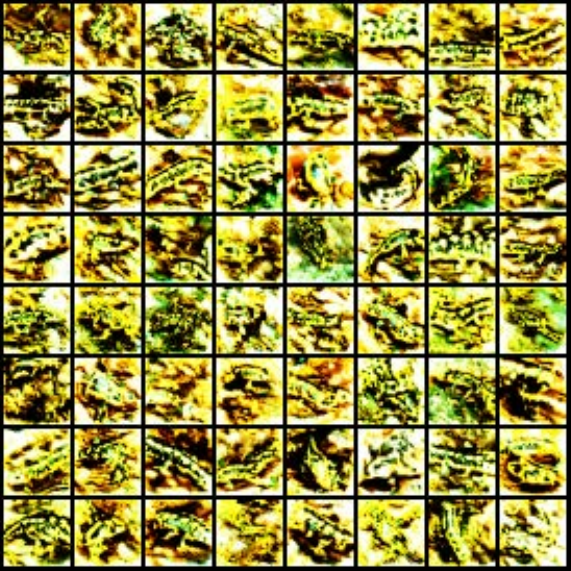}}
    \subfigure{				
    \includegraphics[width=0.18\textwidth]{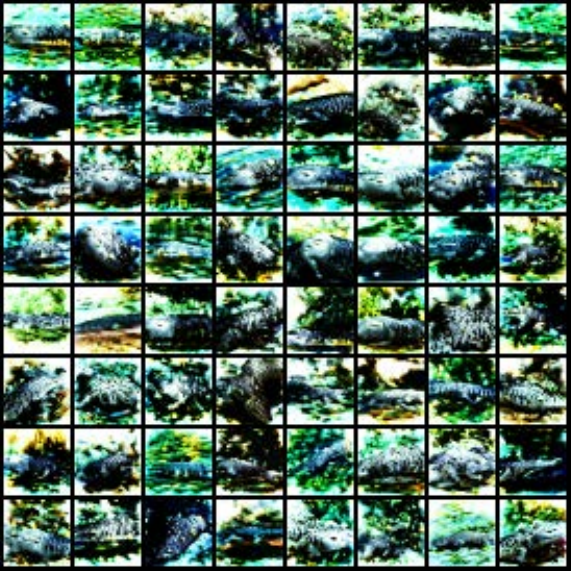}}
    \subfigure{				
    \includegraphics[width=0.18\textwidth]{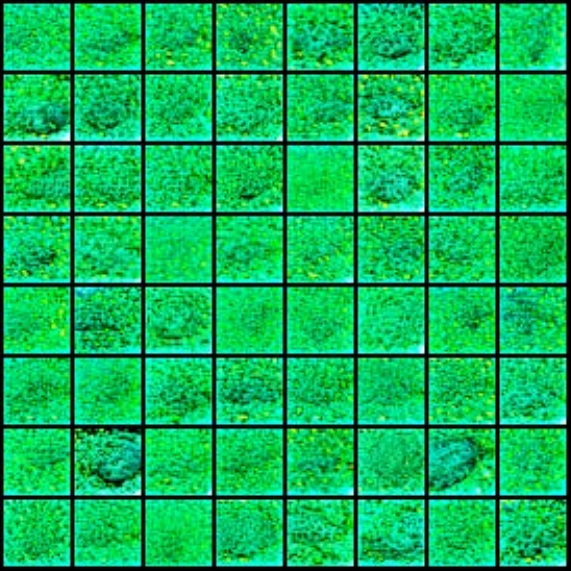}}
    \subfigure{				
    \includegraphics[width=0.18\textwidth]{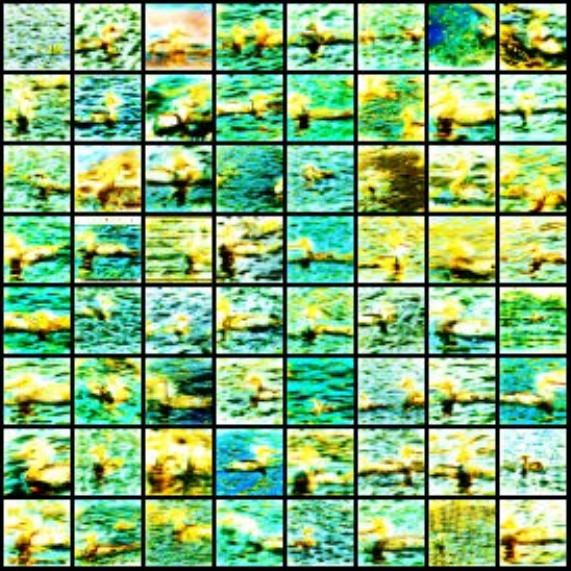}}
    \caption{Generated auxiliary ID data on CIFAR-10 dataset. Each figure contains 64 images of each class.}	
    \label{fig:cifar10p agentID}				
\end{figure}

\begin{figure}[t]
    \centering
    \includegraphics[width=.9\textwidth]{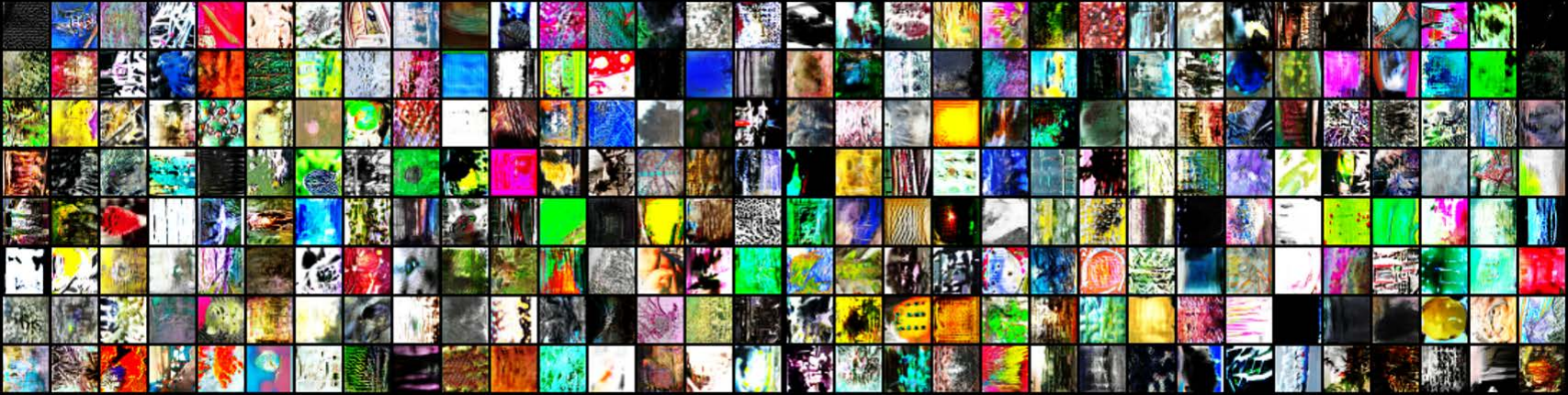} 	
    \caption{Generated auxiliary OOD data on CIFAR-10 dataset.}	
    \label{fig:cifar10p agentOOD}						
\end{figure}

\begin{figure}[t]
    \centering    
    \subfigure{				
    \includegraphics[width=0.18\textwidth]{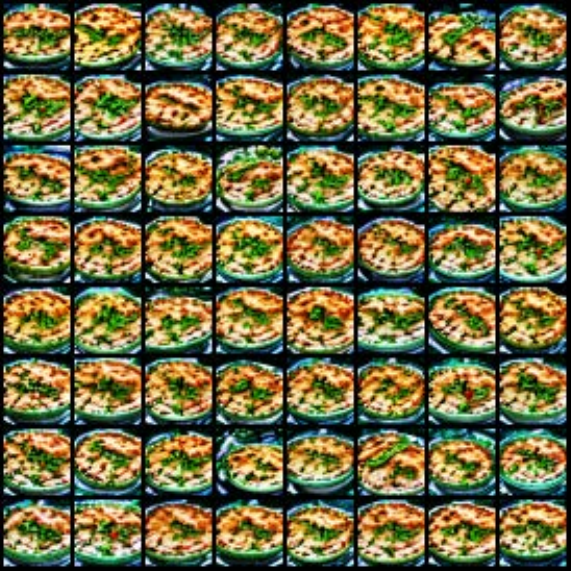}}
    \subfigure{				
    \includegraphics[width=0.18\textwidth]{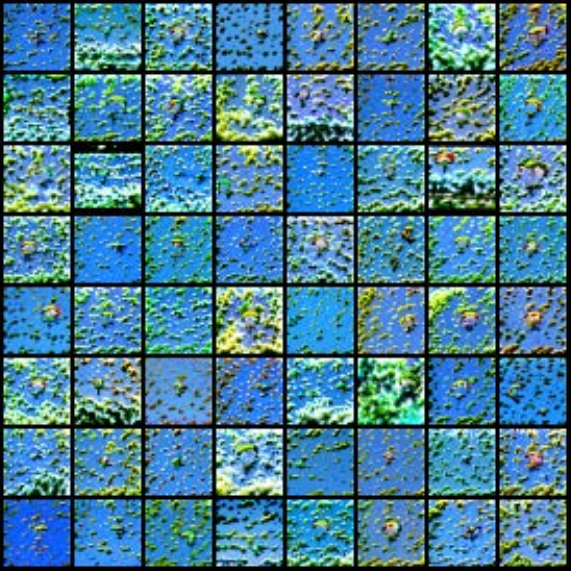}}
    \subfigure{				
    \includegraphics[width=0.18\textwidth]{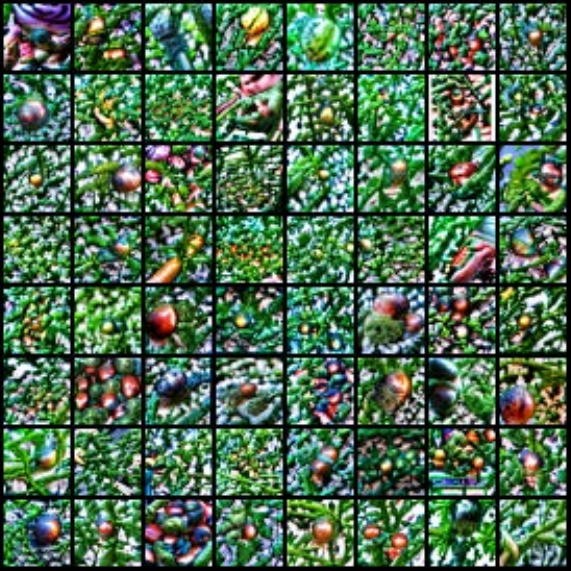}}
    \subfigure{				
    \includegraphics[width=0.18\textwidth]{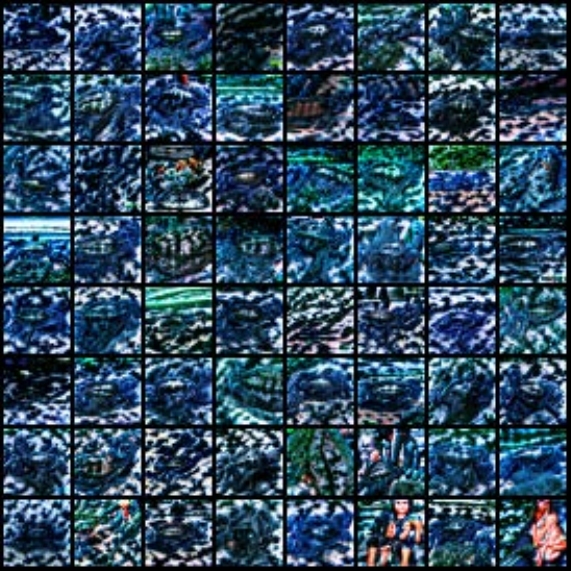}}
    \subfigure{				
    \includegraphics[width=0.18\textwidth]{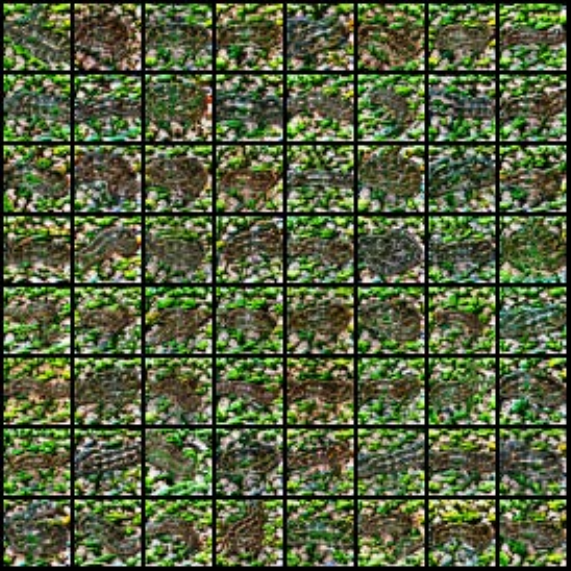}}
    \subfigure{				
    \includegraphics[width=0.18\textwidth]{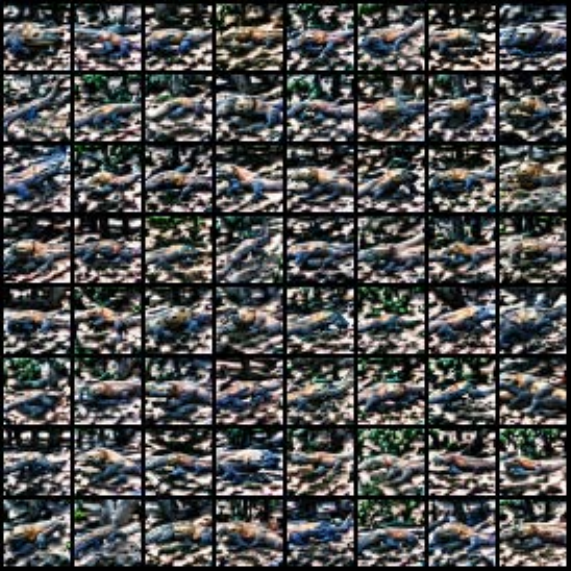}}
    \subfigure{				
    \includegraphics[width=0.18\textwidth]{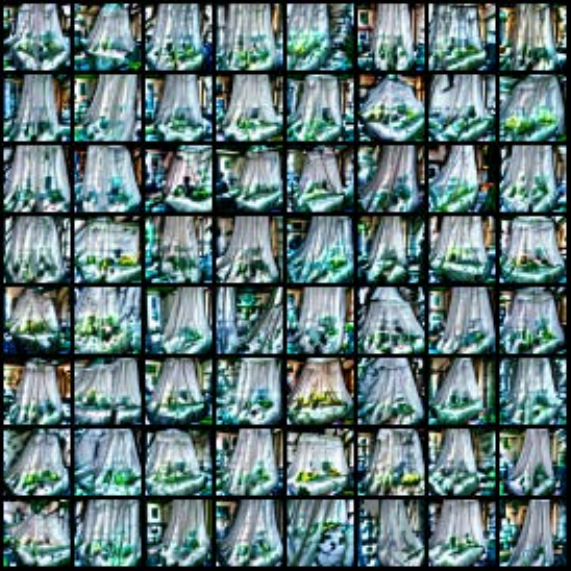}}
    \subfigure{				
    \includegraphics[width=0.18\textwidth]{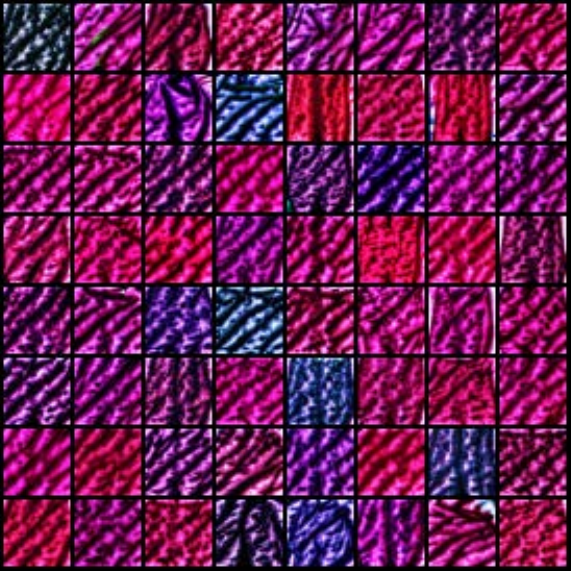}}
    \subfigure{				
    \includegraphics[width=0.18\textwidth]{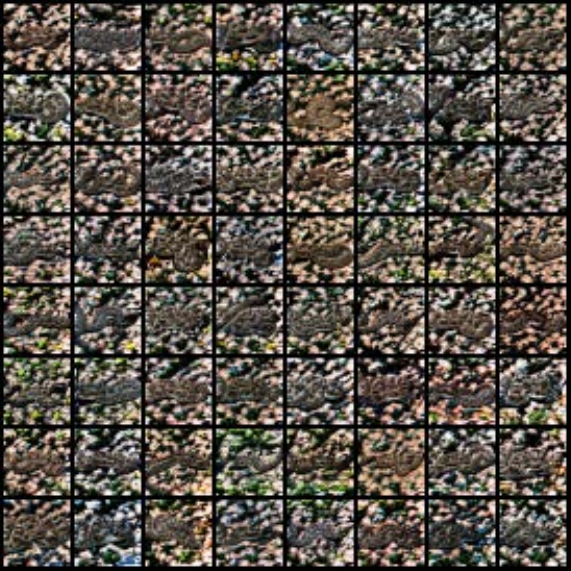}}
    \subfigure{				
    \includegraphics[width=0.18\textwidth]{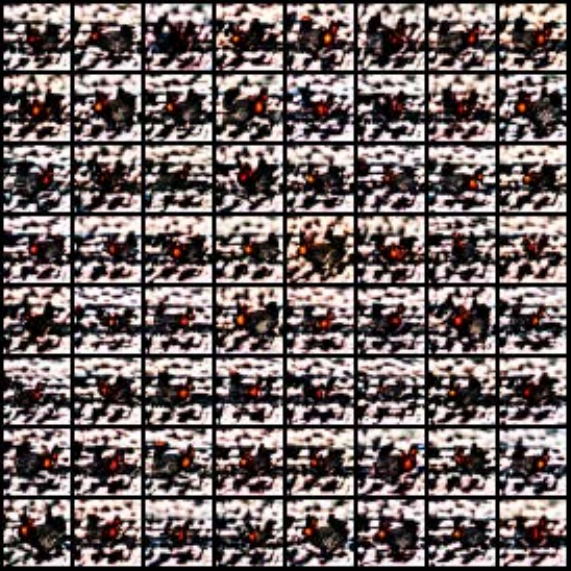}}
    \caption{Generated auxiliary ID data on CIFAR-100 dataset. Each figure contains 64 images of each class. Due to space limitations, we only show 10 out of 100 classes.}		
    \label{fig:cifar100p agentID}			
\end{figure}

\begin{figure}[t]
    \centering
    \includegraphics[width=.9\textwidth]{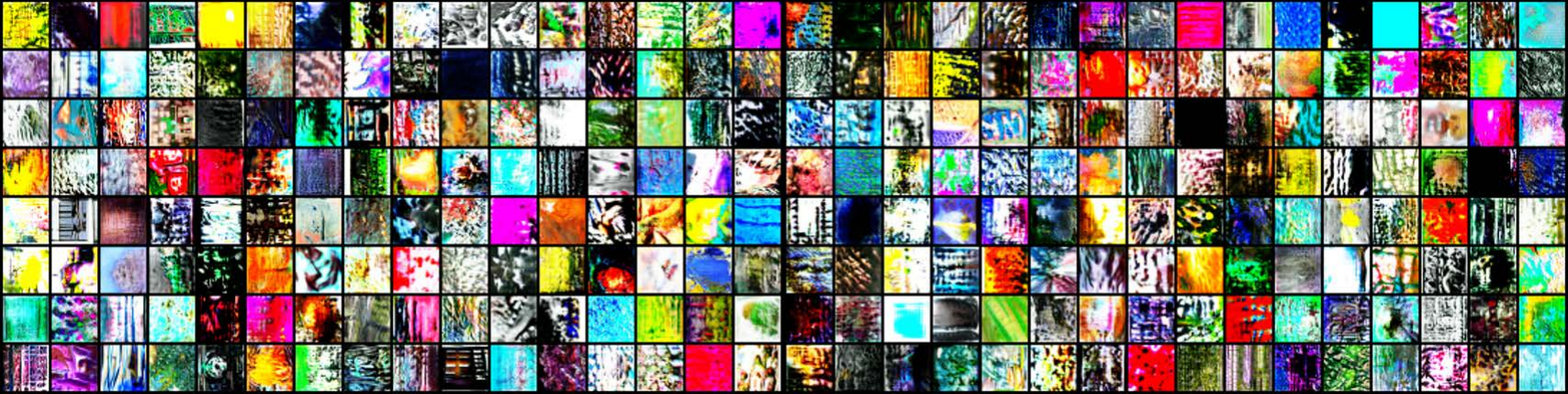} 	
    \caption{Generated auxiliary OOD data on CIFAR-100 dataset.}	 
    \label{fig:cifar100p agentOOD}							
\end{figure}

\begin{figure}[t]
    \centering    
    \subfigure{				
    \includegraphics[width=0.18\textwidth]{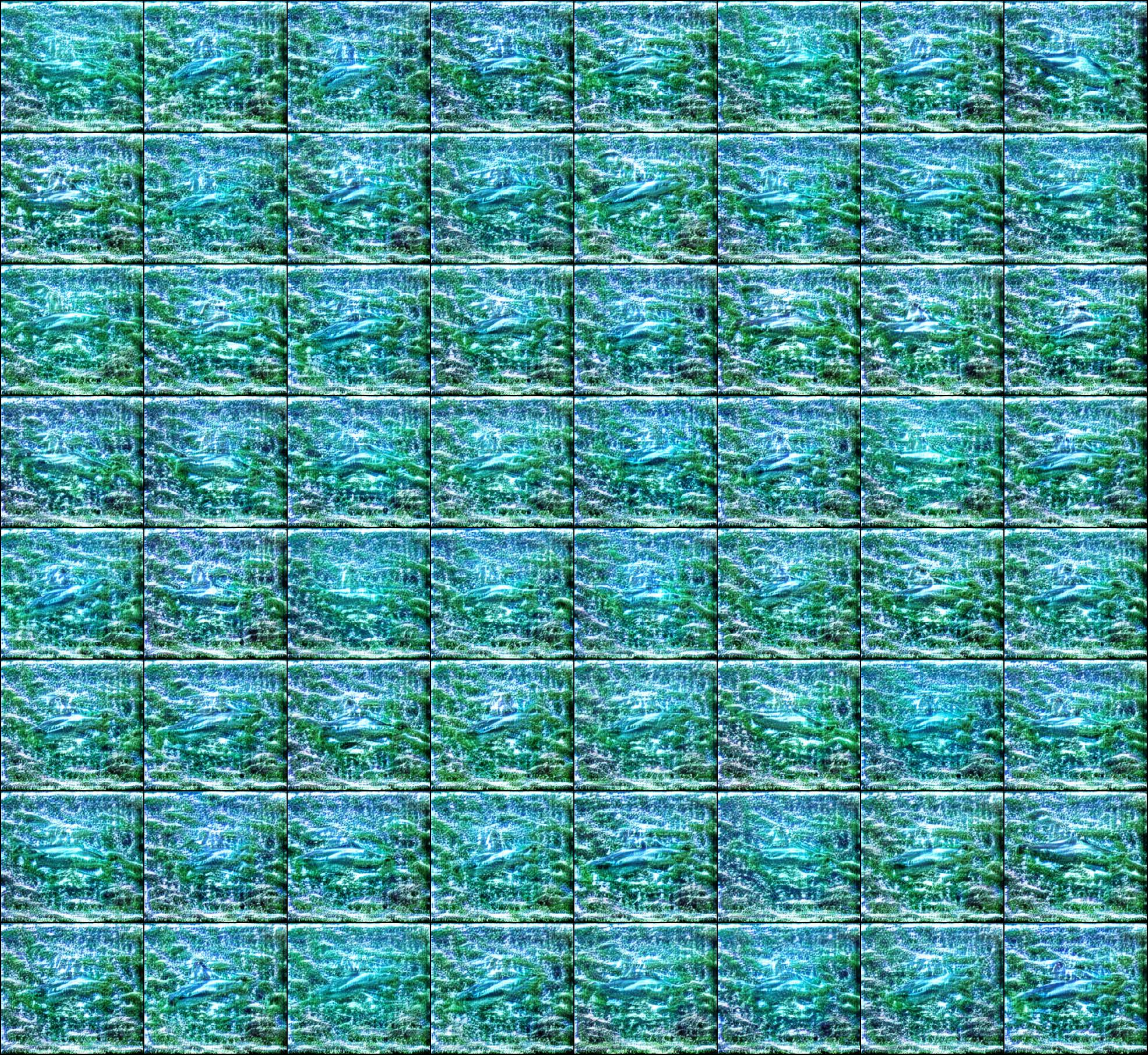}}
    \subfigure{				
    \includegraphics[width=0.18\textwidth]{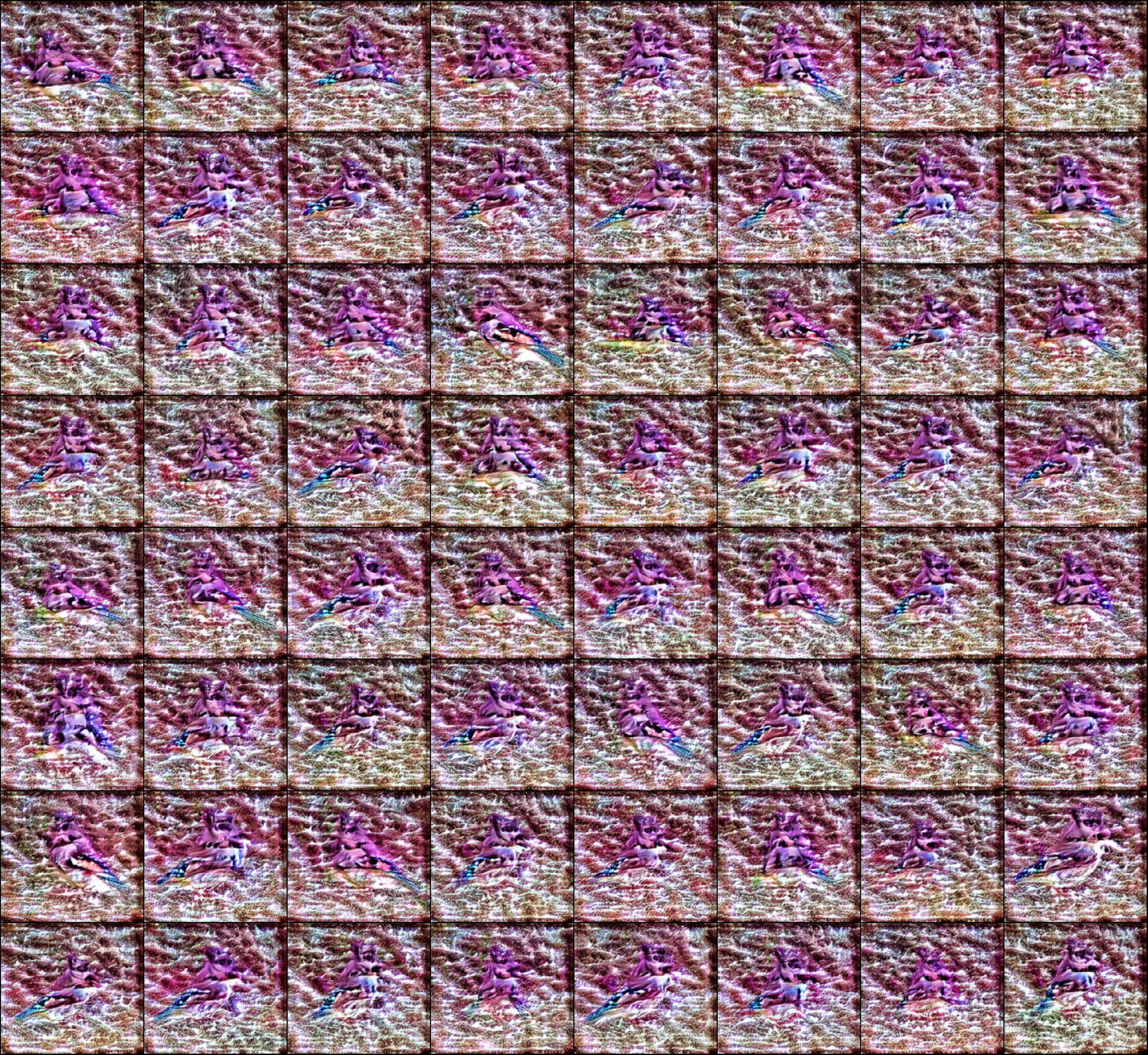}}
    \subfigure{				
    \includegraphics[width=0.18\textwidth]{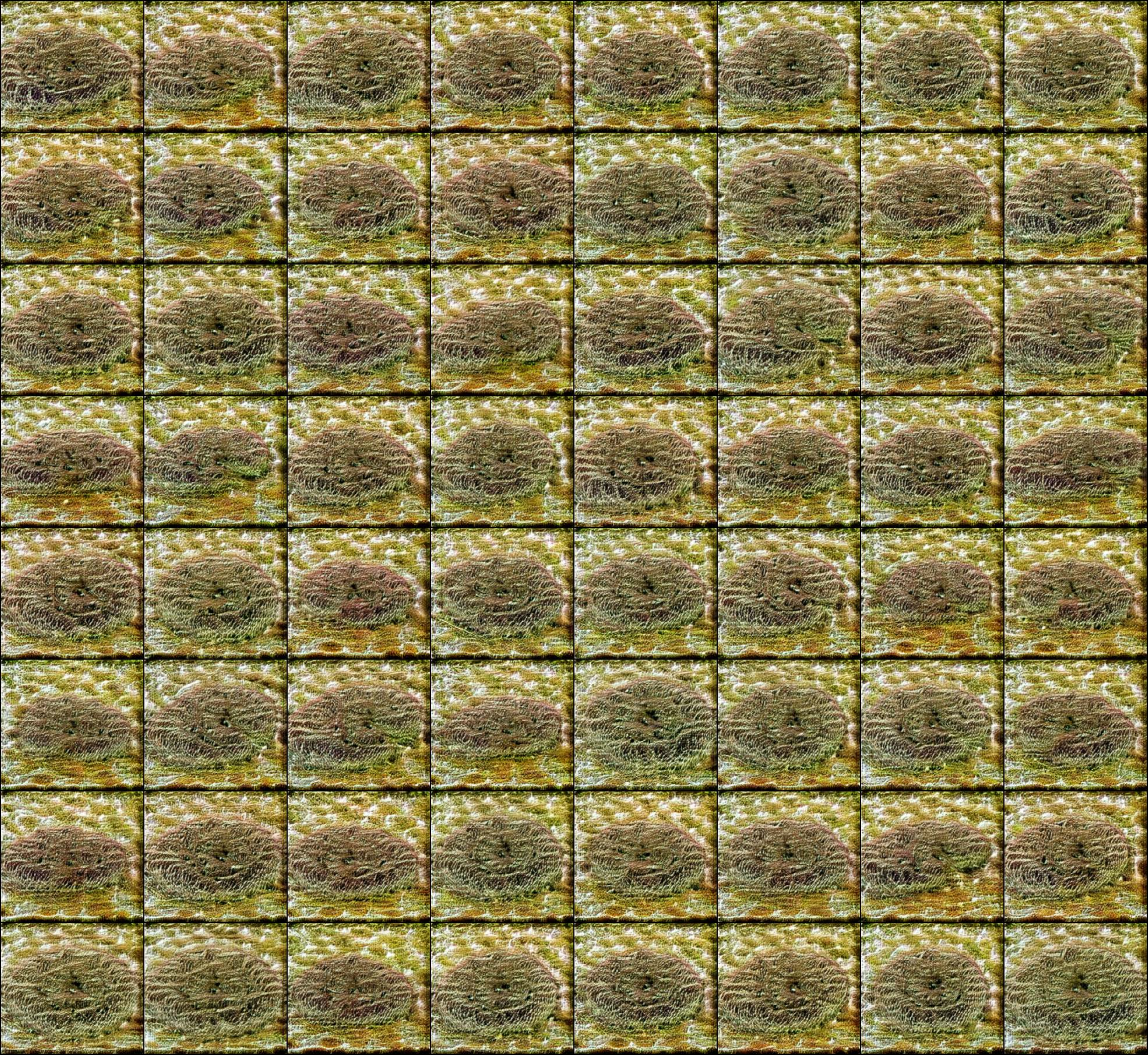}}
    \subfigure{				
    \includegraphics[width=0.18\textwidth]{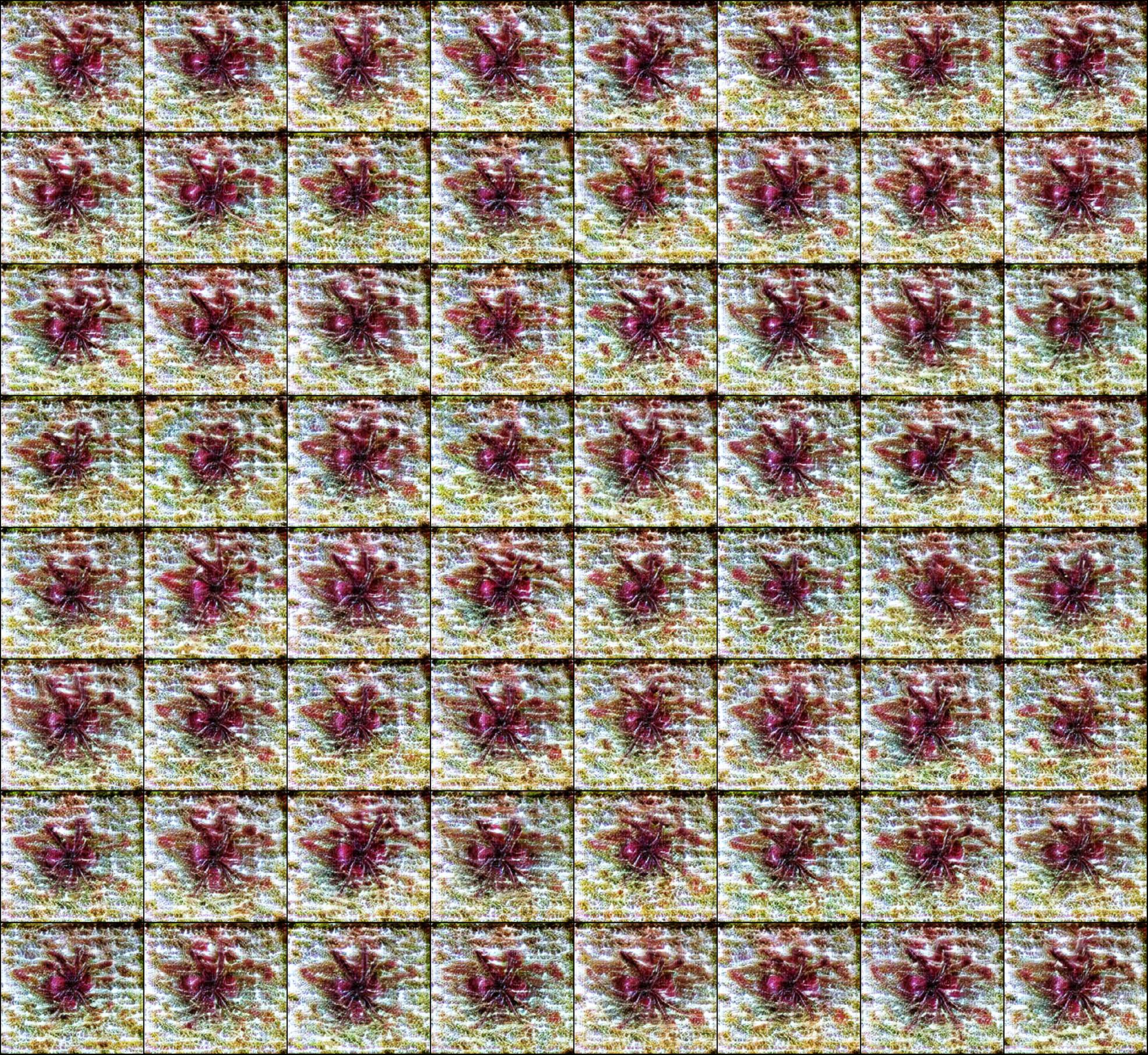}}
    \subfigure{				
    \includegraphics[width=0.18\textwidth]{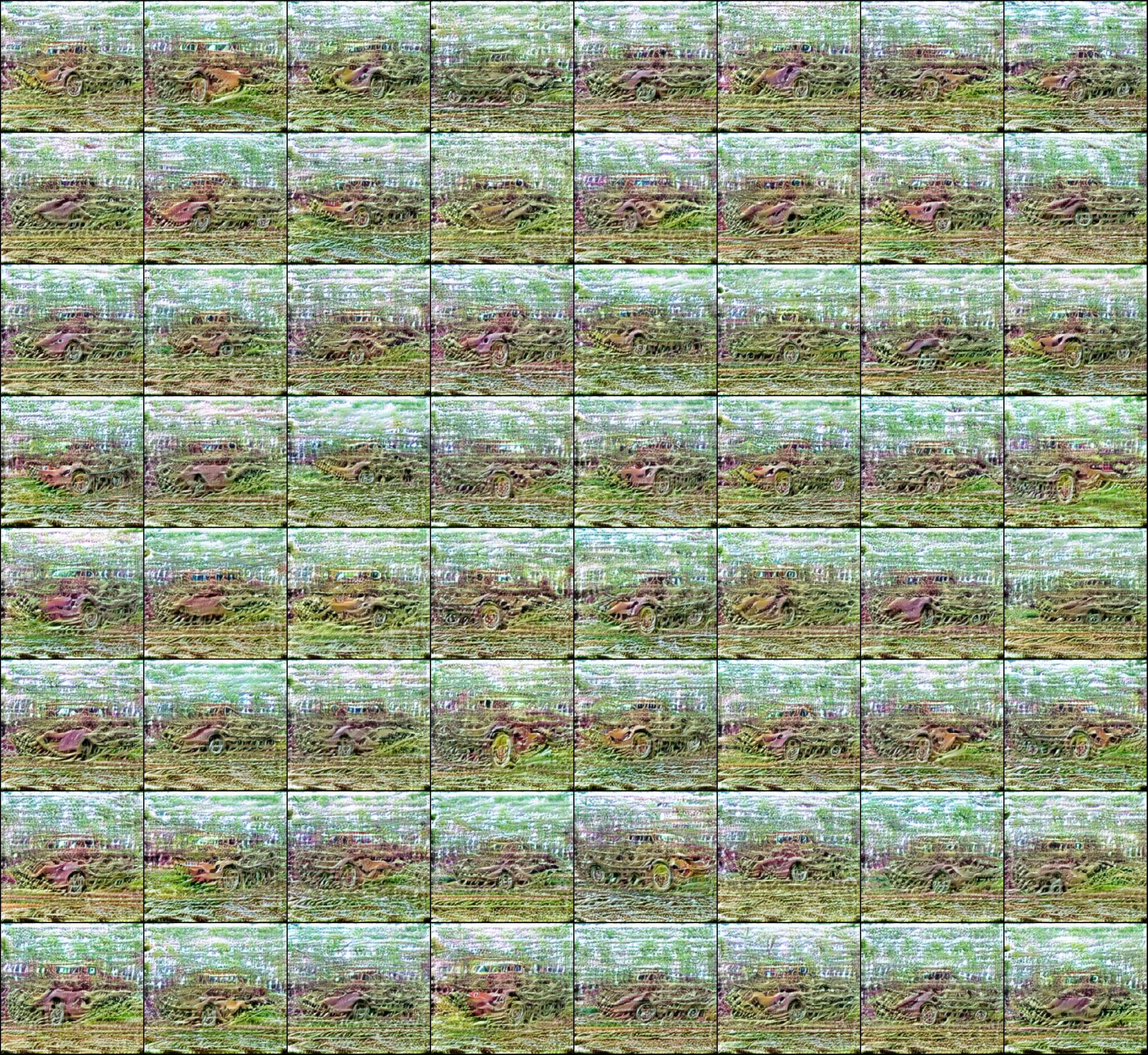}}
    \subfigure{				
    \includegraphics[width=0.18\textwidth]{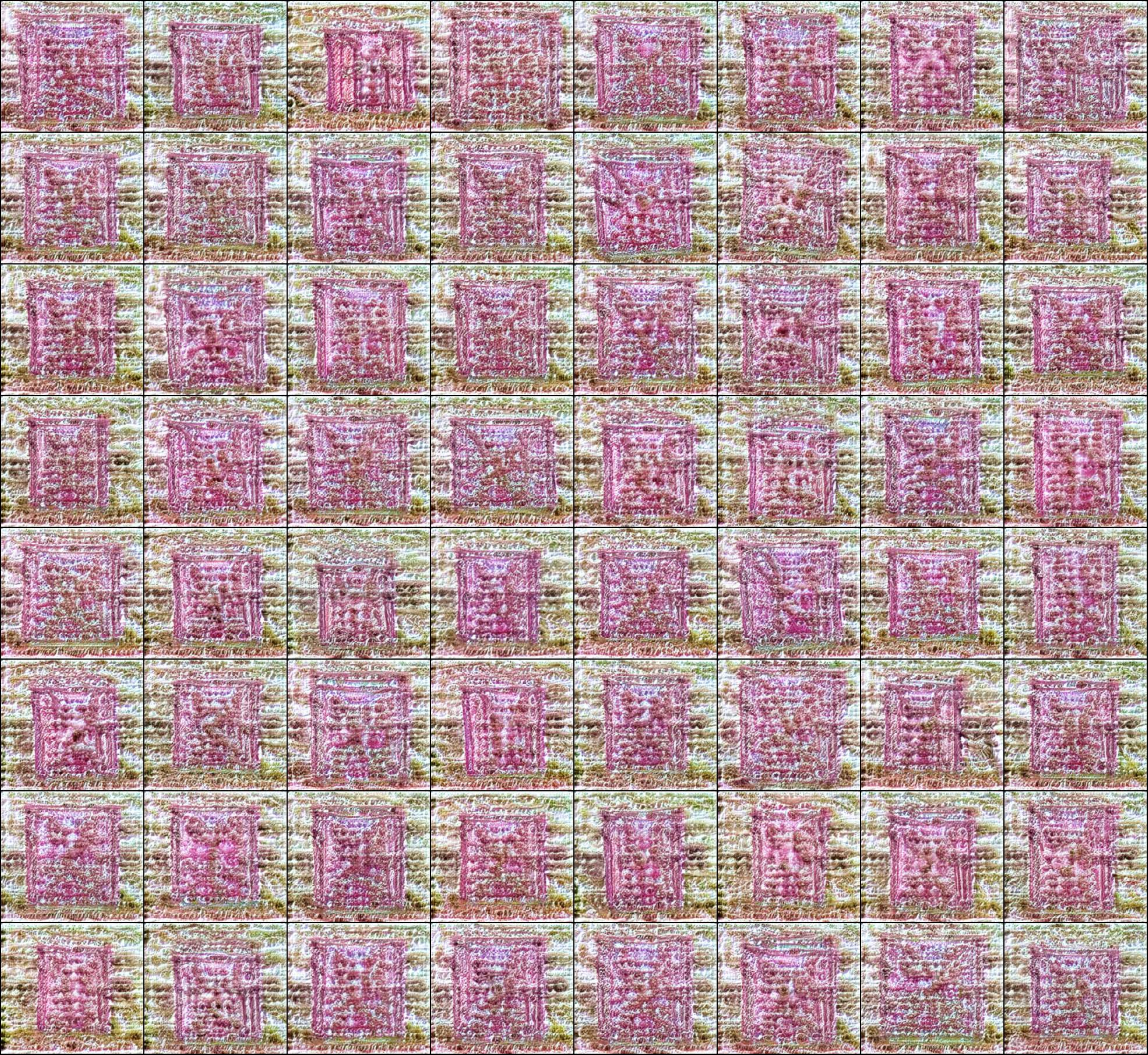}}
    \subfigure{				
    \includegraphics[width=0.18\textwidth]{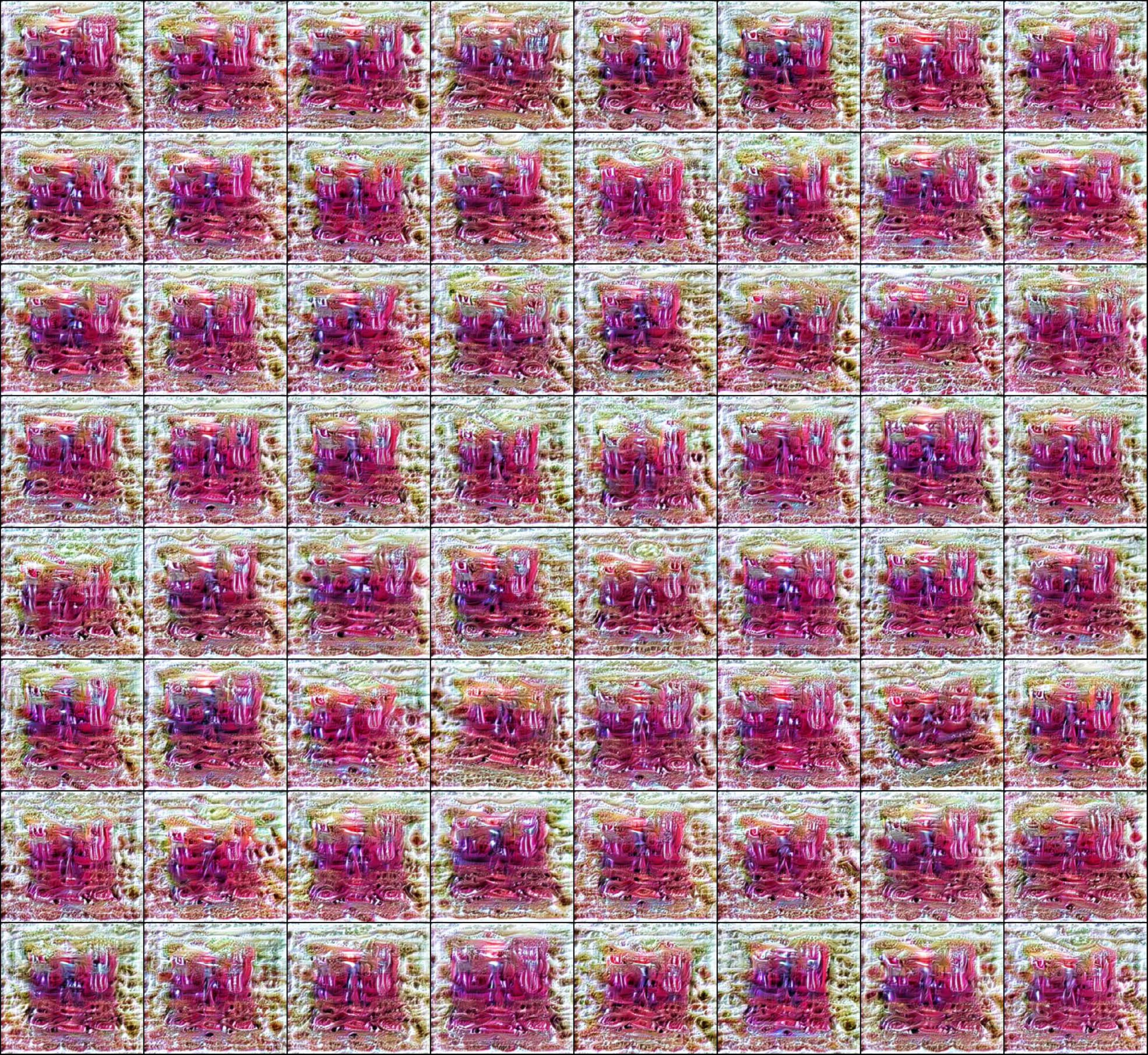}}
    \subfigure{				
    \includegraphics[width=0.18\textwidth]{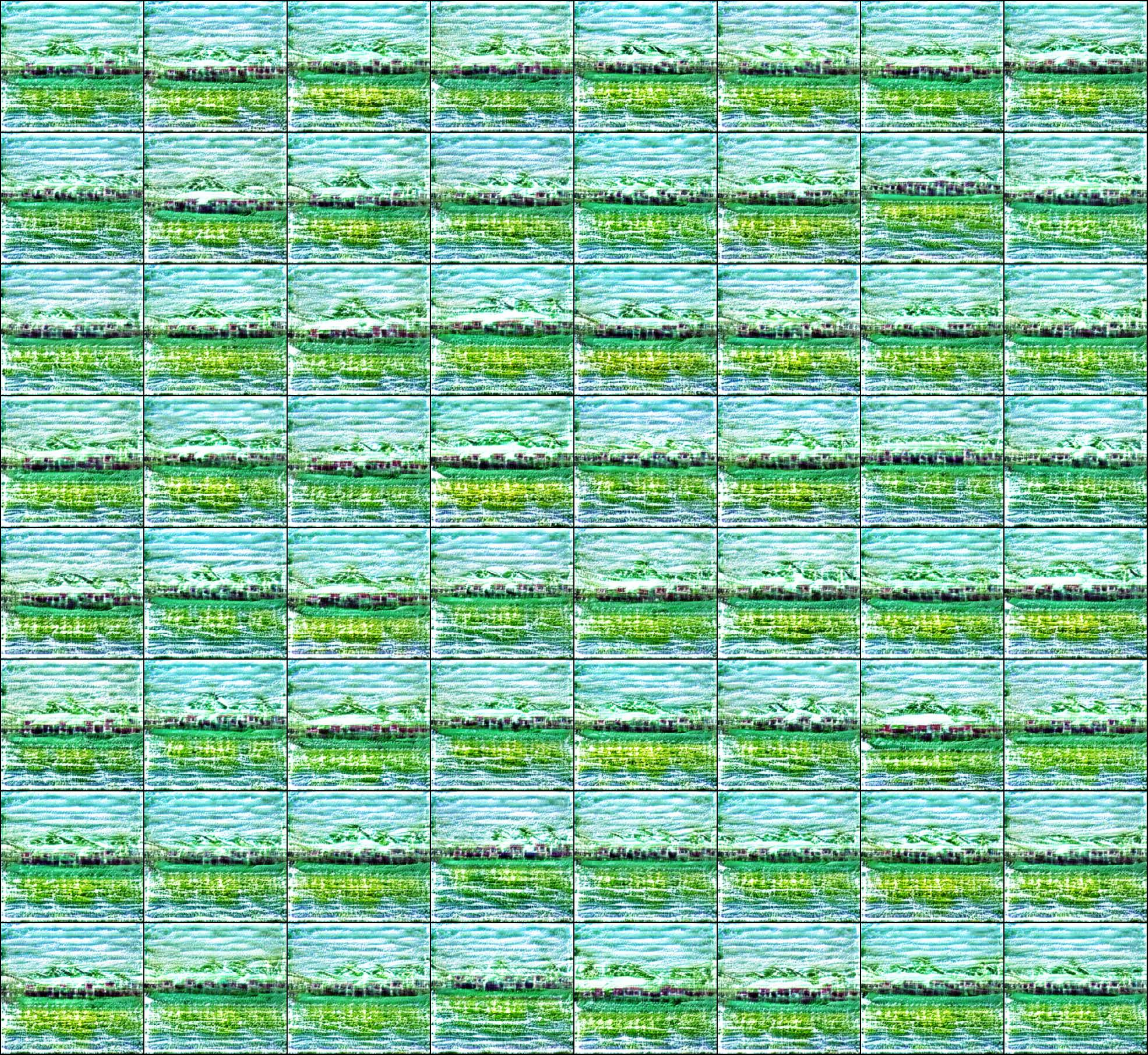}}
    \subfigure{				
    \includegraphics[width=0.18\textwidth]{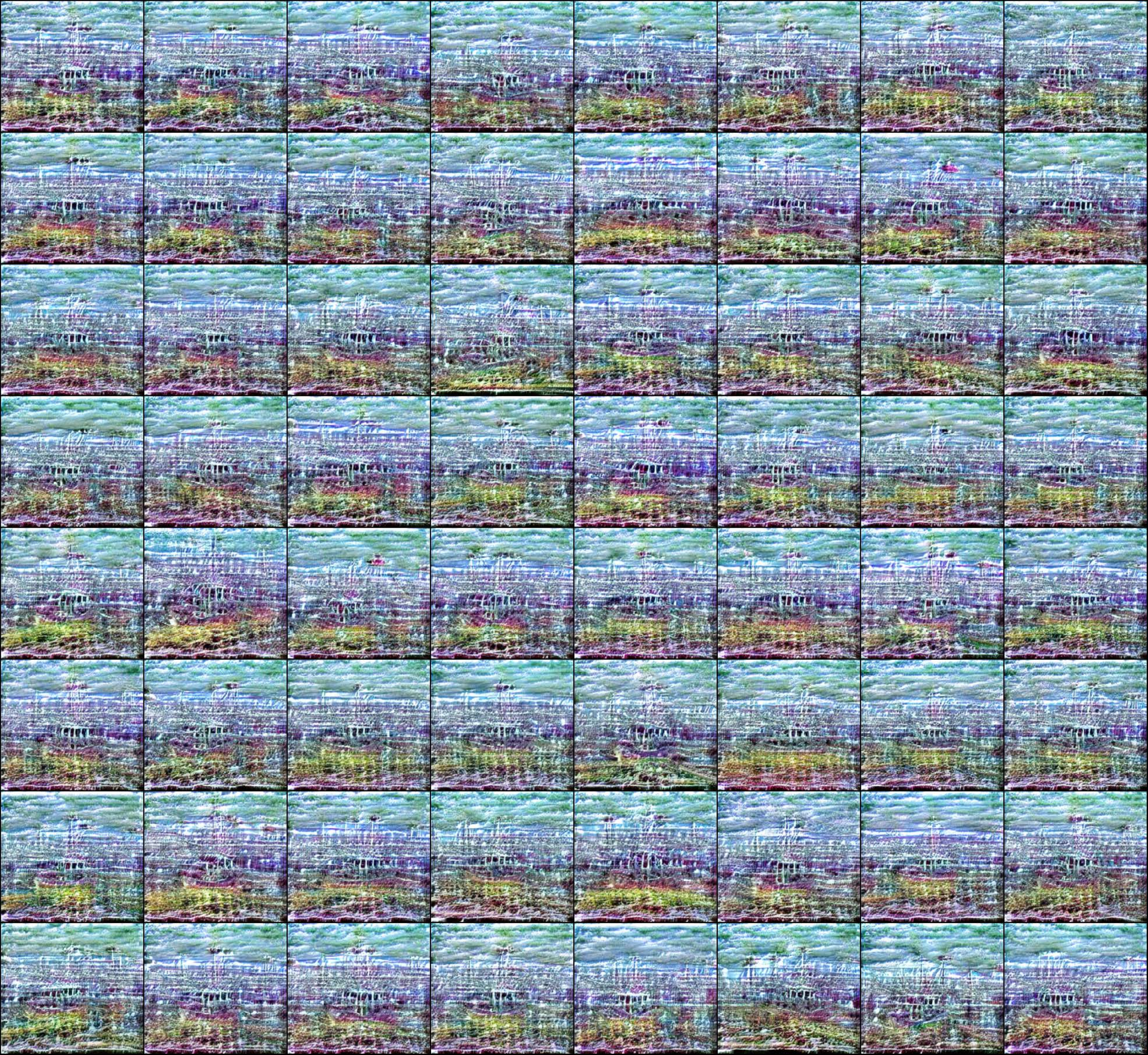}}
    \subfigure{				
    \includegraphics[width=0.18\textwidth]{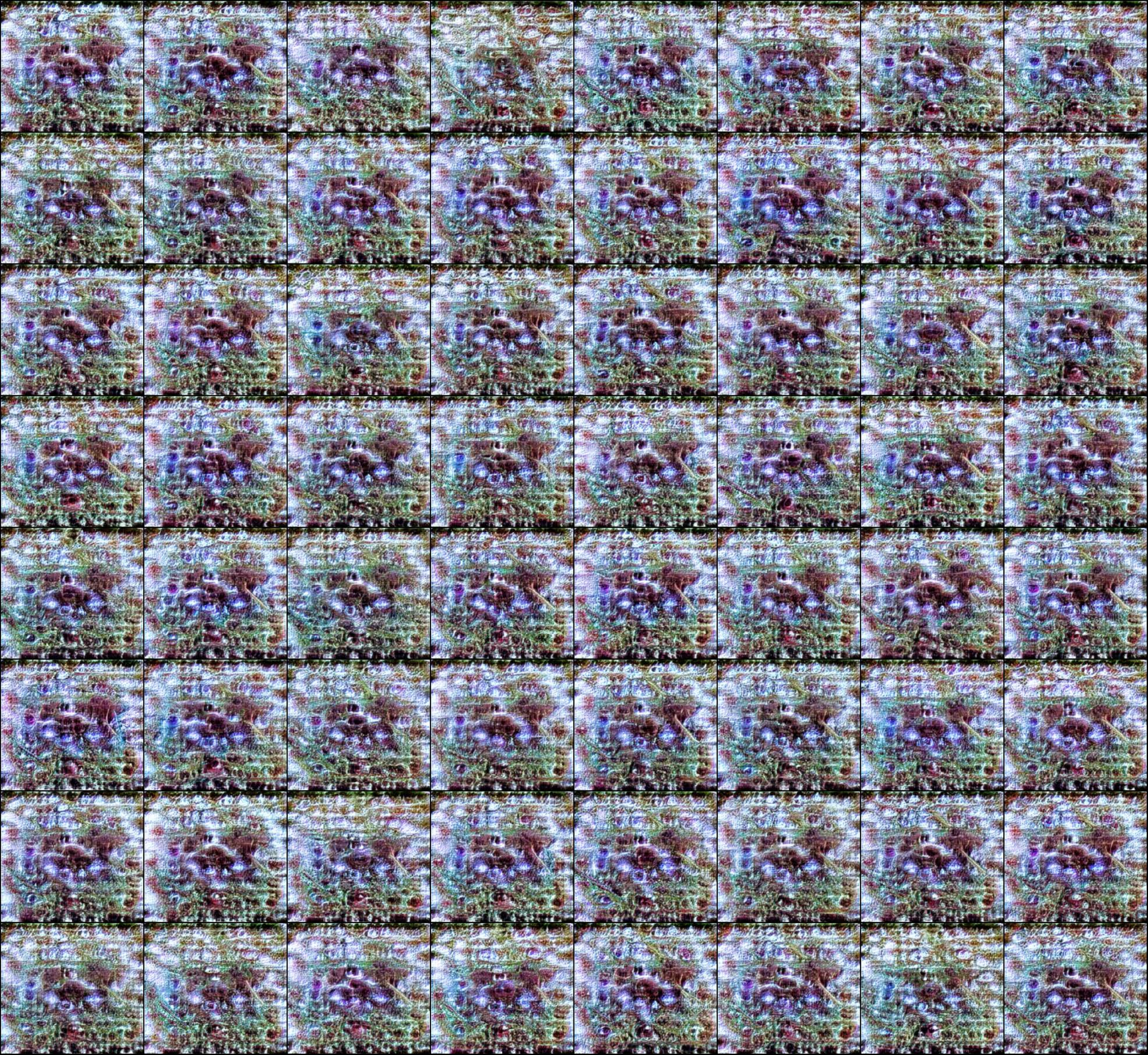}}
    \caption{Generated auxiliary ID data on ImageNet dataset. Each figure contains 64 images of each class. Due to space limitations, we only show 10 out of 1000 classes.}	
    \label{fig:imagenet agentID}						
\end{figure}

\begin{figure}[t]
    \centering
    \includegraphics[width=.9\textwidth]{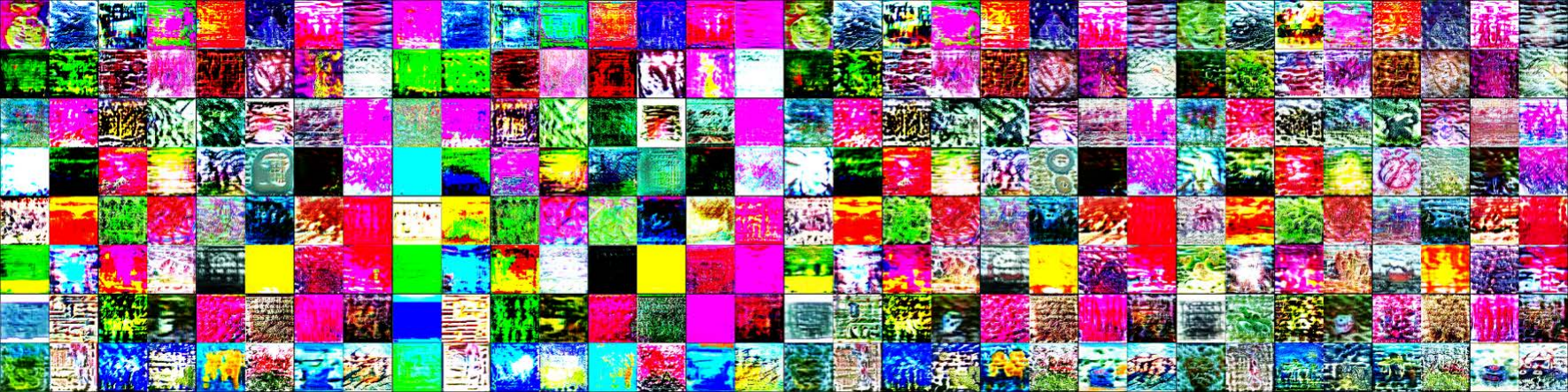} 	
    \caption{Generated auxiliary OOD data on ImageNet dataset.}	
    \label{fig:imagenet agentOOD}						
\end{figure}

In this section, we visualize some synthesized examples for intuitive demonstration. As demonstrated in the main paper, ATOL performs surprisingly well on CIFAR-$10$, CIFAR-$100$, and ImageNet. ATOL also enables us to generate visual results for intuitive inspection. For auxiliary ID cases, we sample noise from different Gaussian distributions from MoG with different means but the same standard deviation in latent space, representing different classes of the auxiliary ID data. For auxiliary OOD cases, we sample noise from a uniform distribution in latent space except for the region with high MoG density. The generated auxiliary data are visualized in Figure~\ref{fig:cifar10p agentID}-\ref{fig:imagenet agentOOD}.

\begin{figure}[t]
    \centering    
    \subfigure{				
    \includegraphics[width=0.18\textwidth]{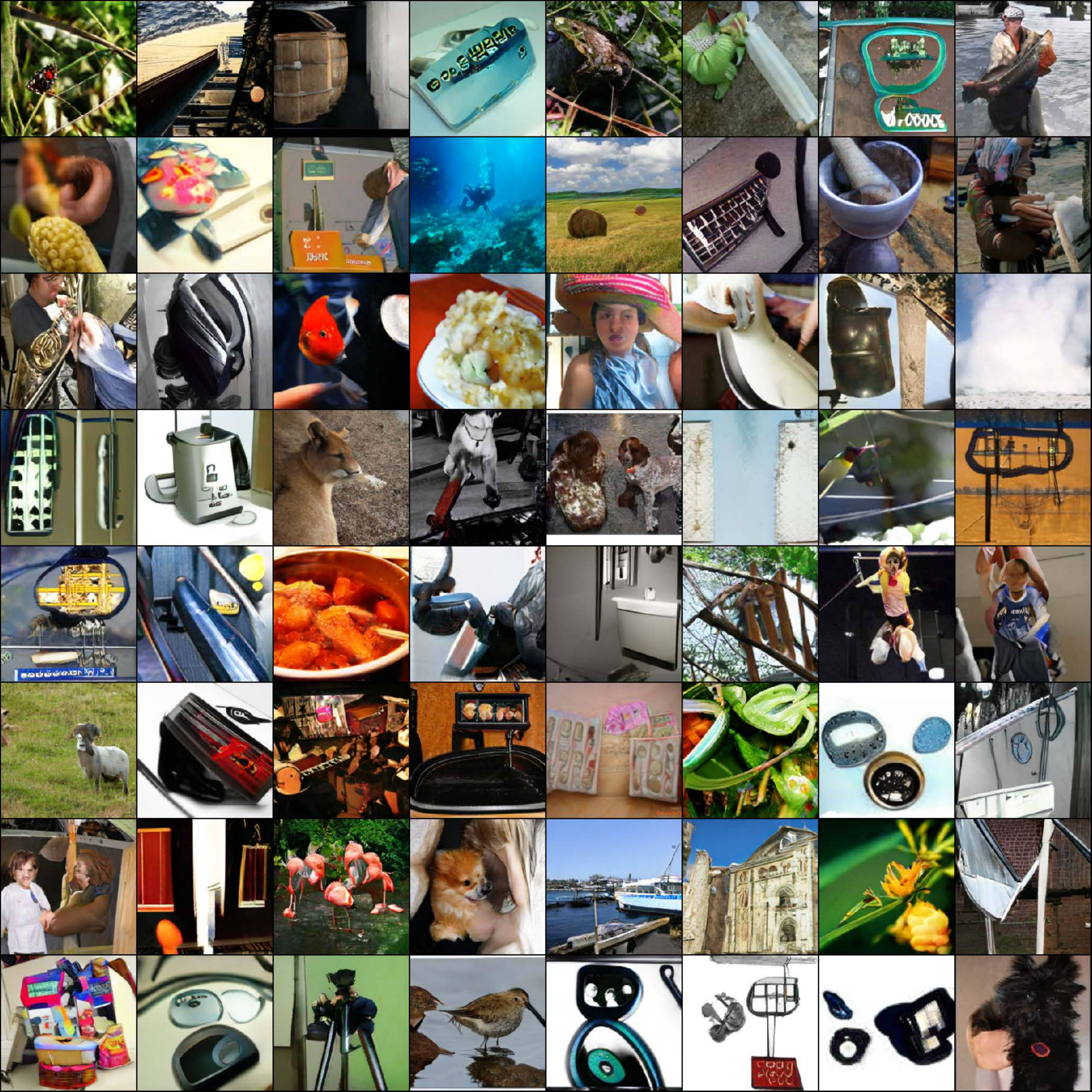}}
    \subfigure{				
    \includegraphics[width=0.18\textwidth]{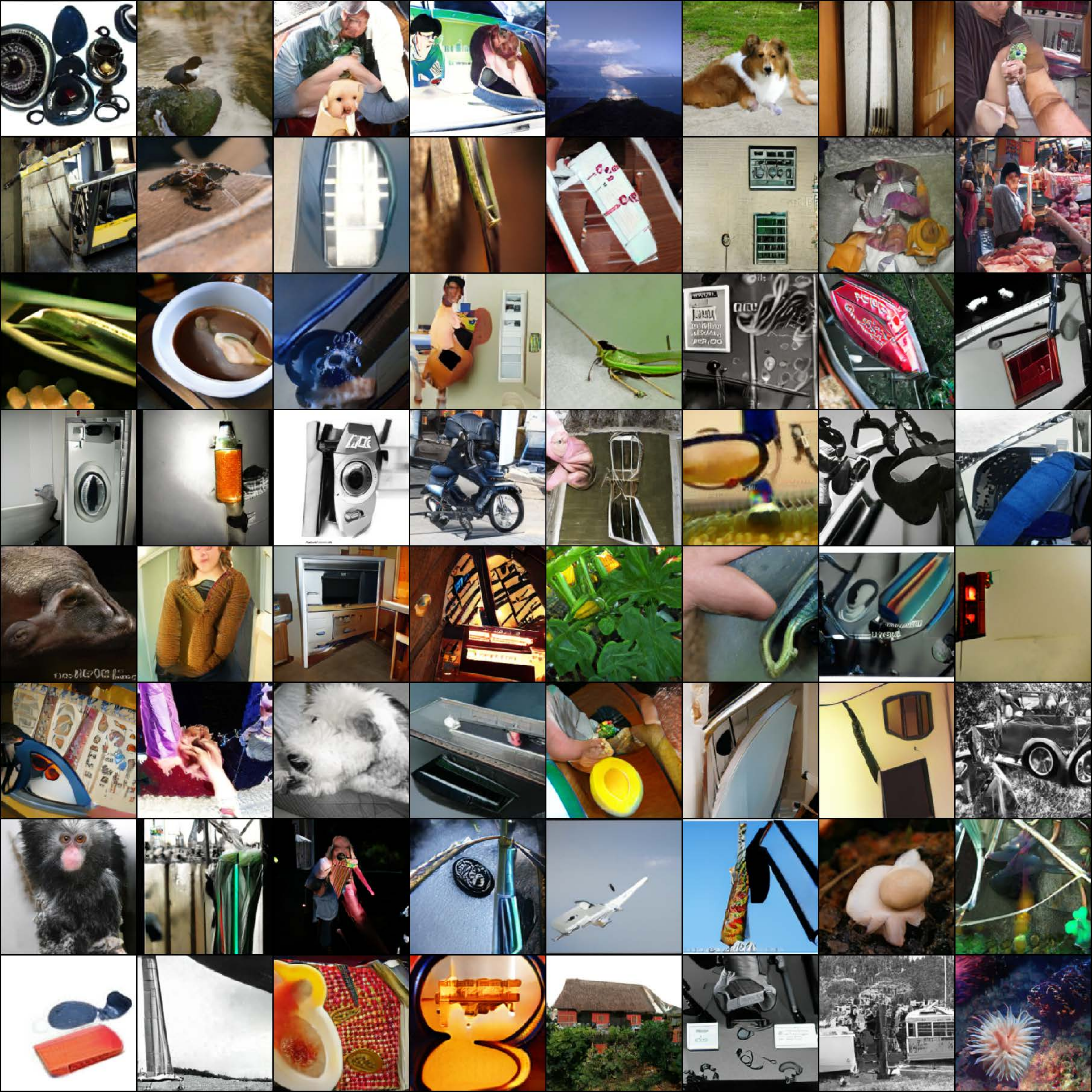}}
    \subfigure{				
    \includegraphics[width=0.18\textwidth]{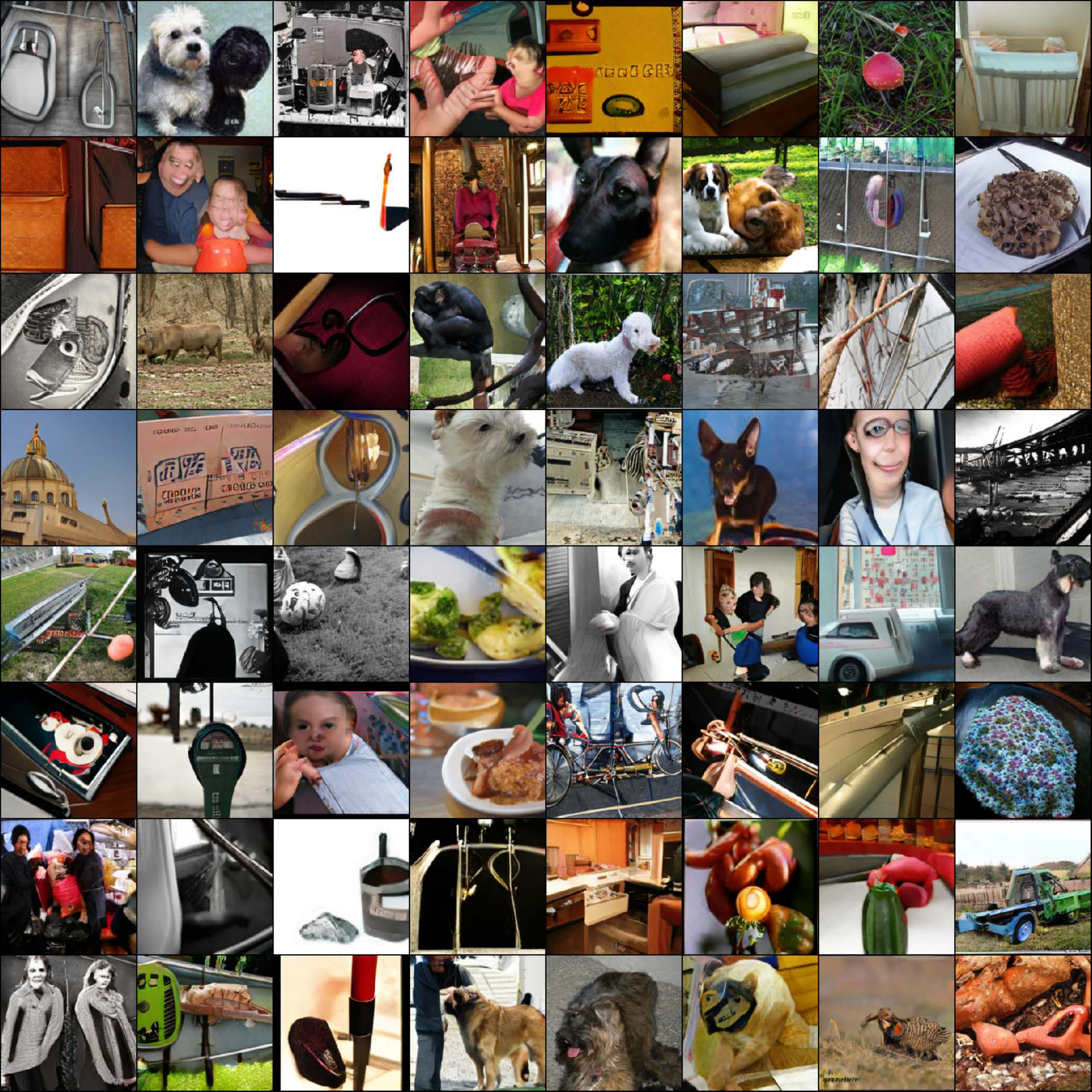}}
    \subfigure{				
    \includegraphics[width=0.18\textwidth]{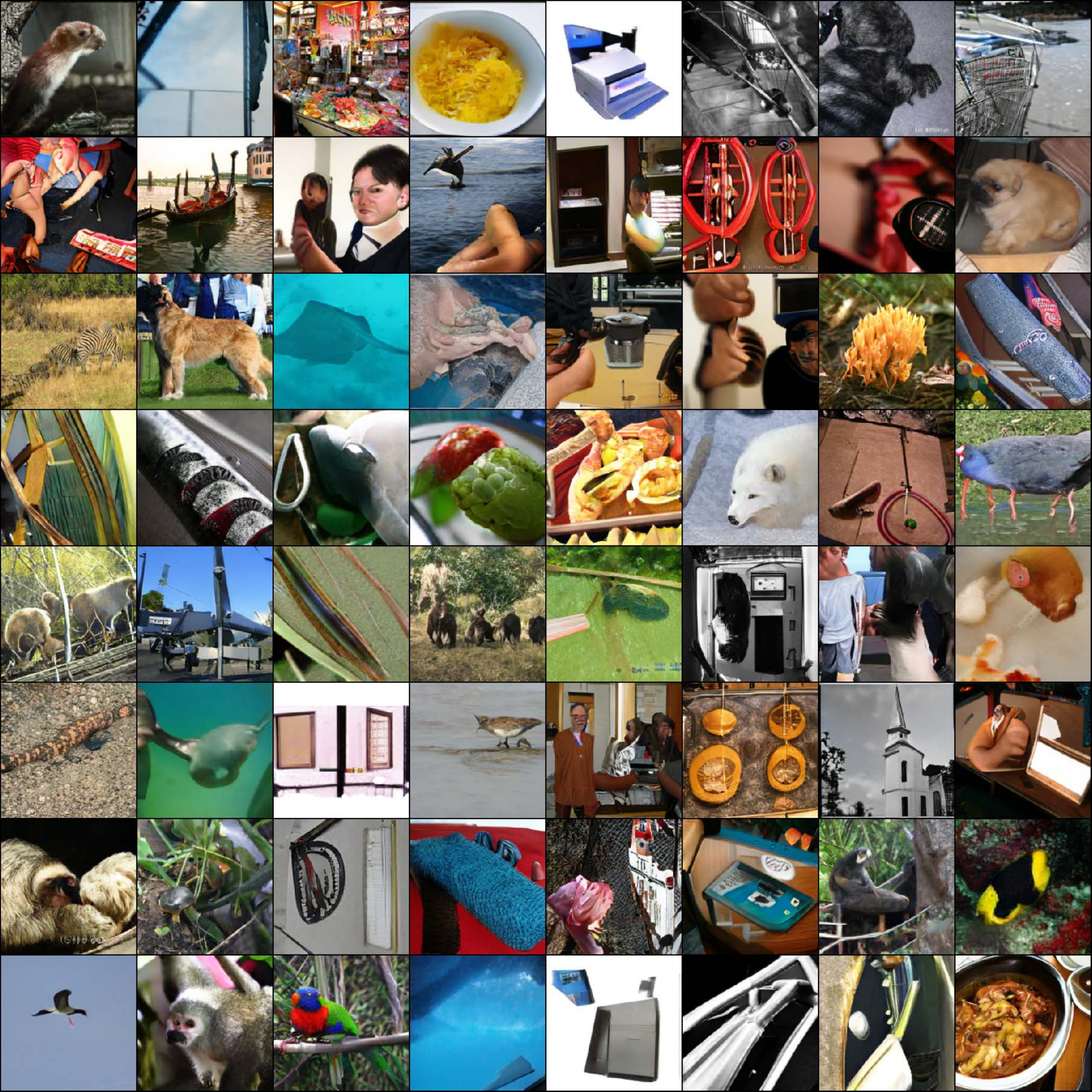}}
    \subfigure{				
    \includegraphics[width=0.18\textwidth]{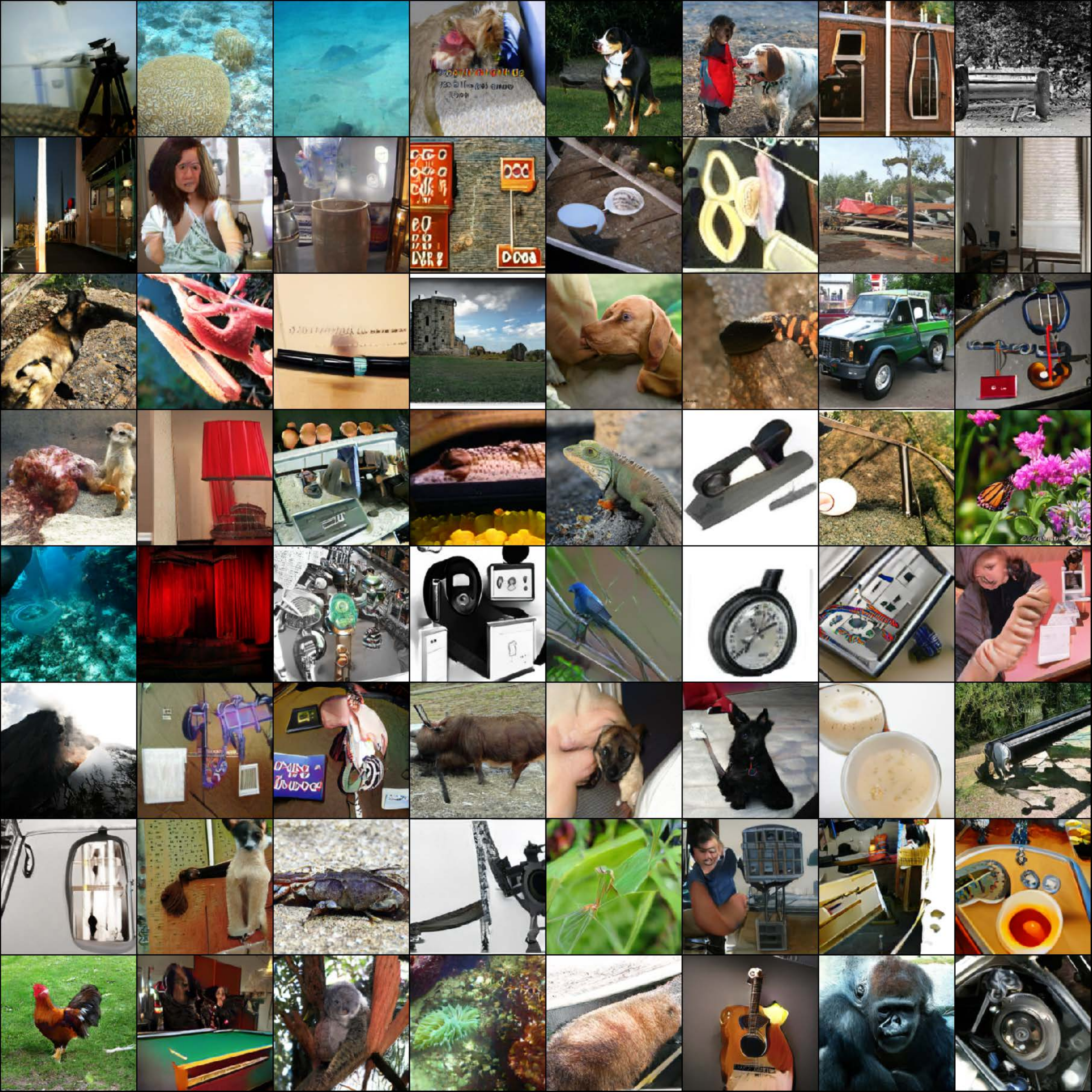}}
    \subfigure{				
    \includegraphics[width=0.18\textwidth]{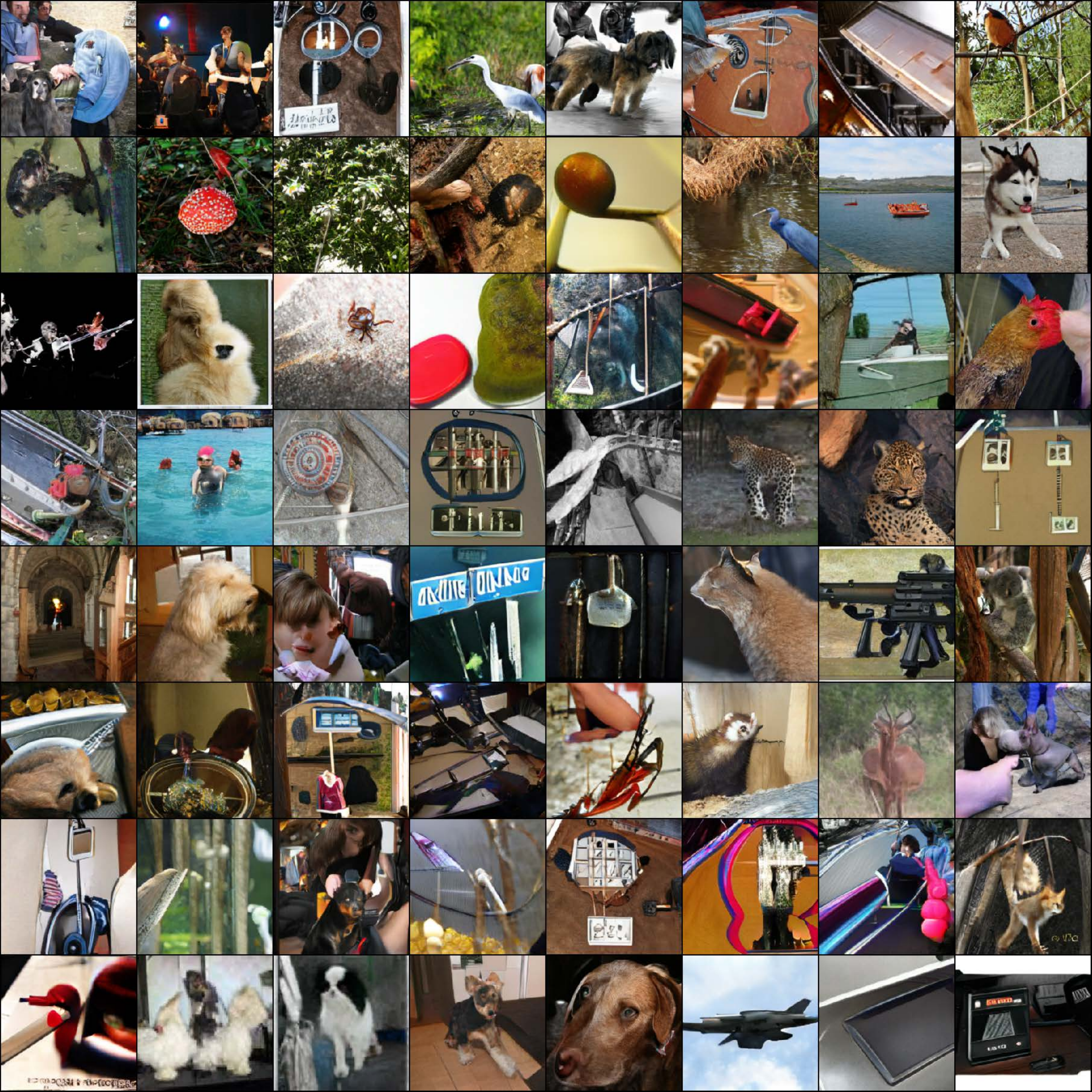}}
    \subfigure{				
    \includegraphics[width=0.18\textwidth]{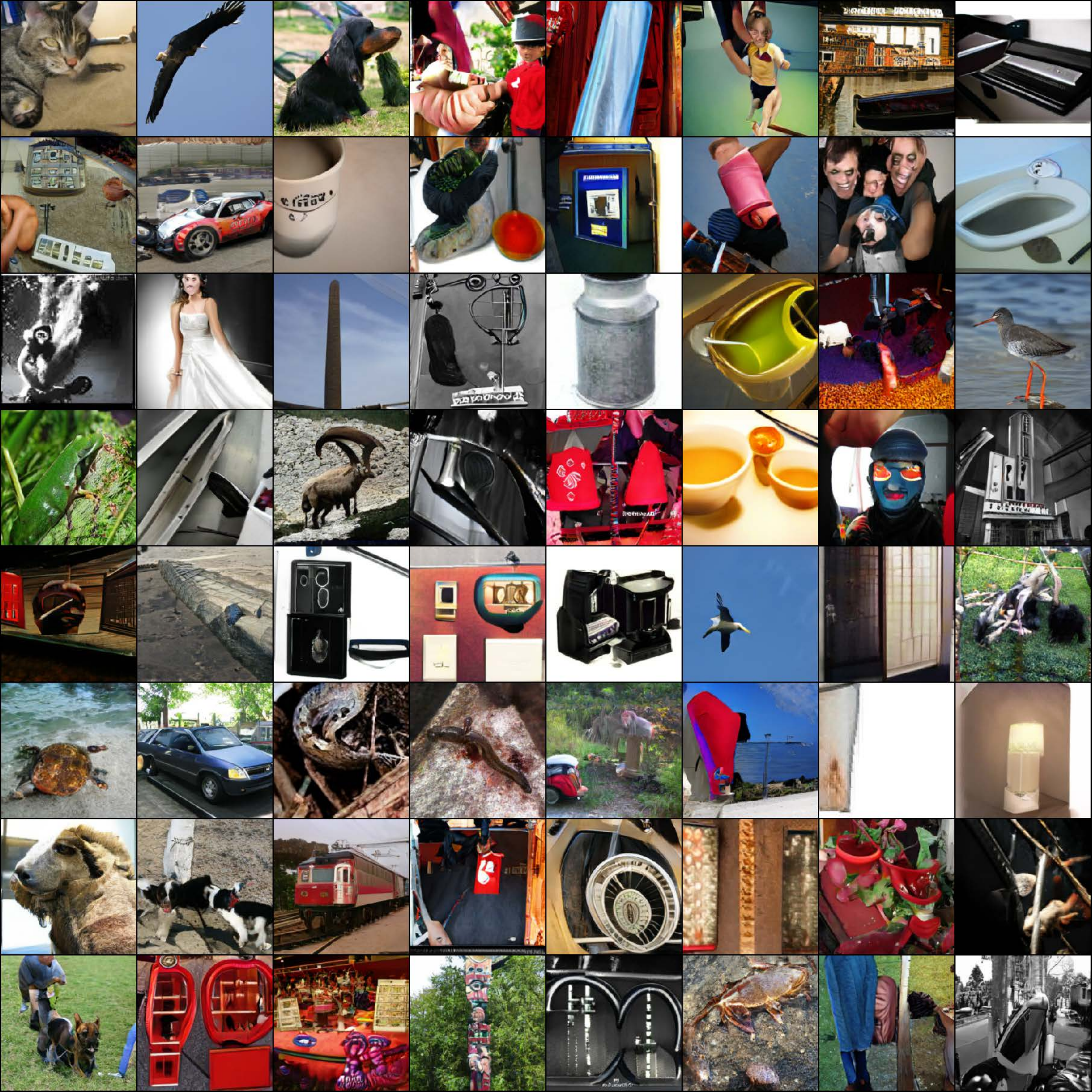}}
    \subfigure{				
    \includegraphics[width=0.18\textwidth]{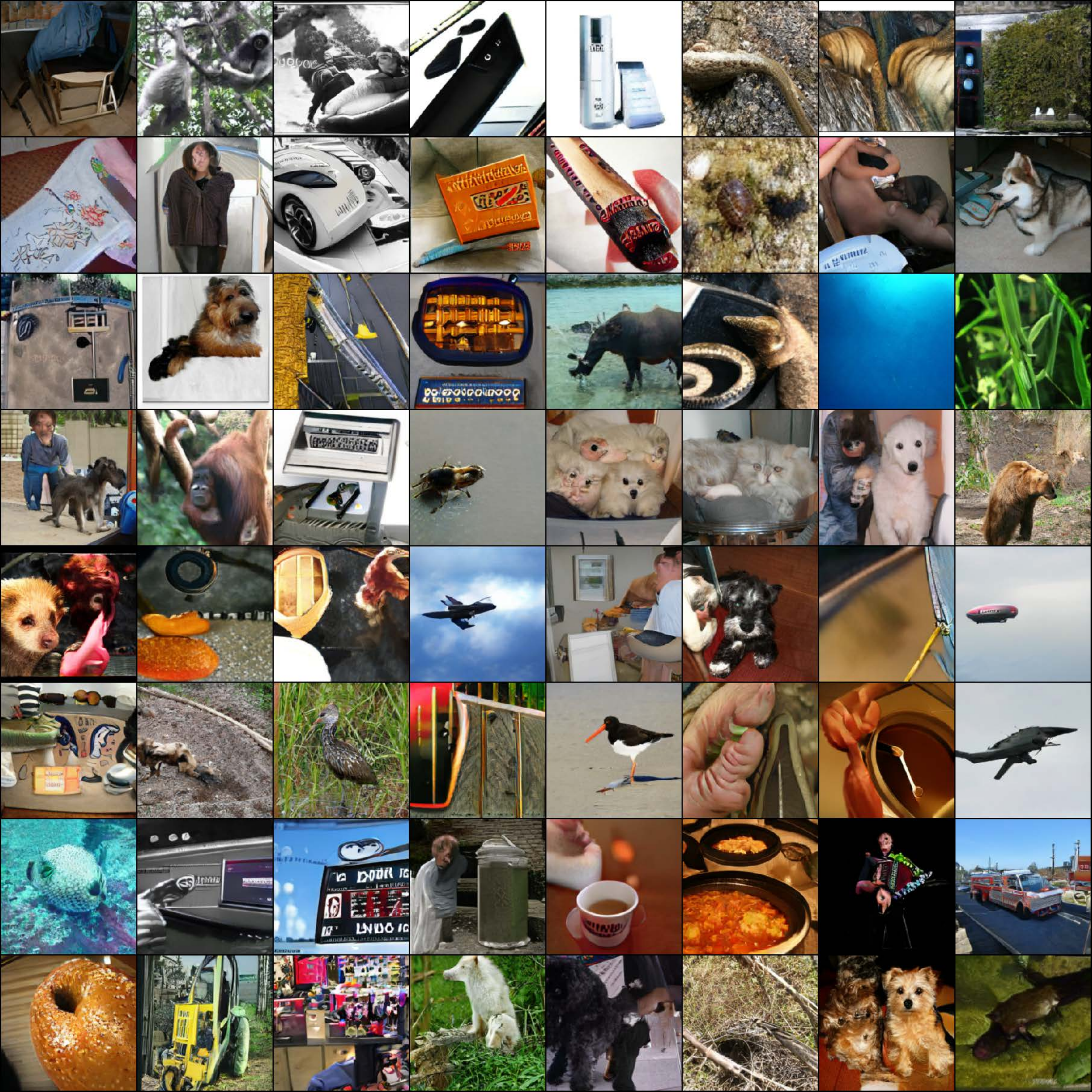}}
    \subfigure{				
    \includegraphics[width=0.18\textwidth]{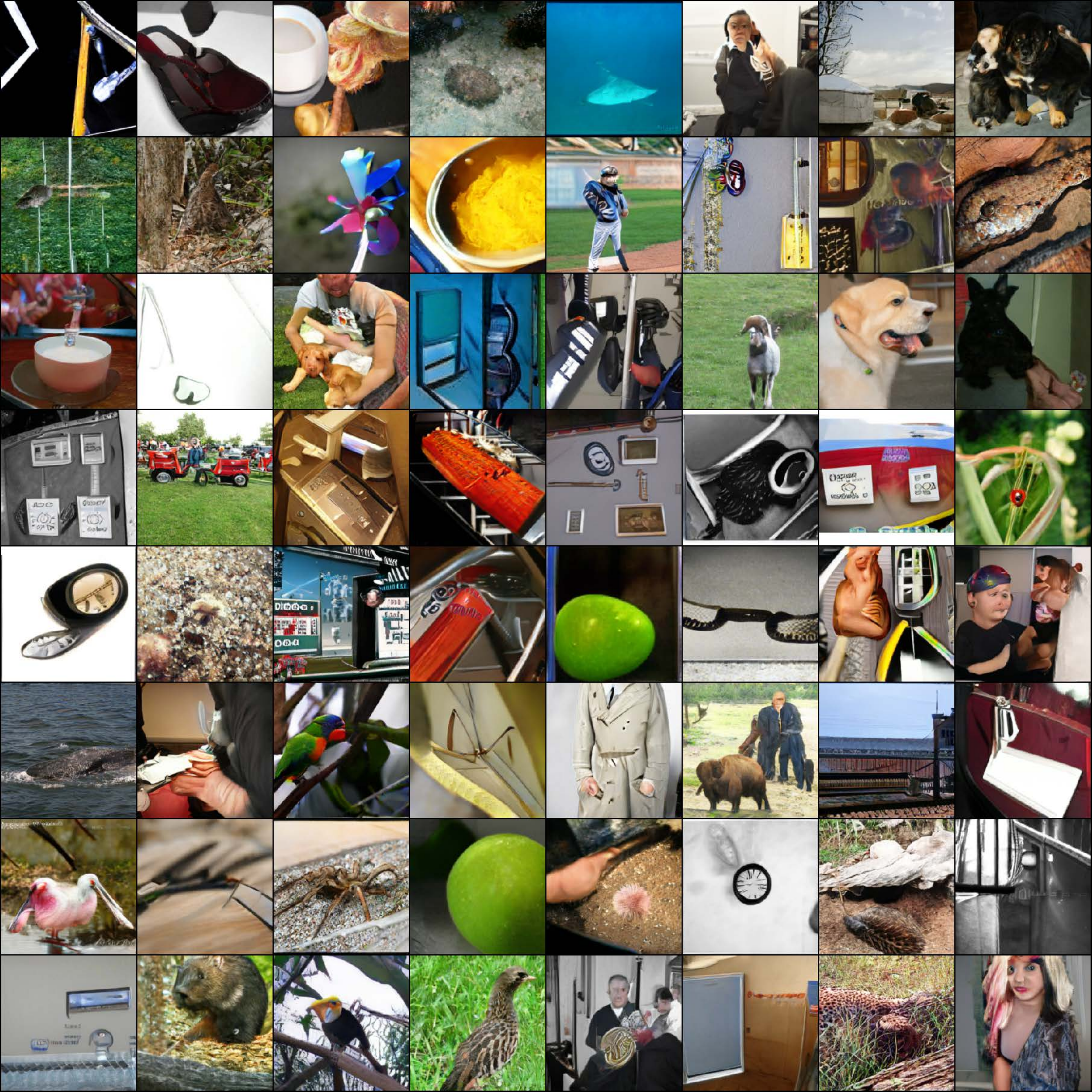}}
    \subfigure{				
    \includegraphics[width=0.18\textwidth]{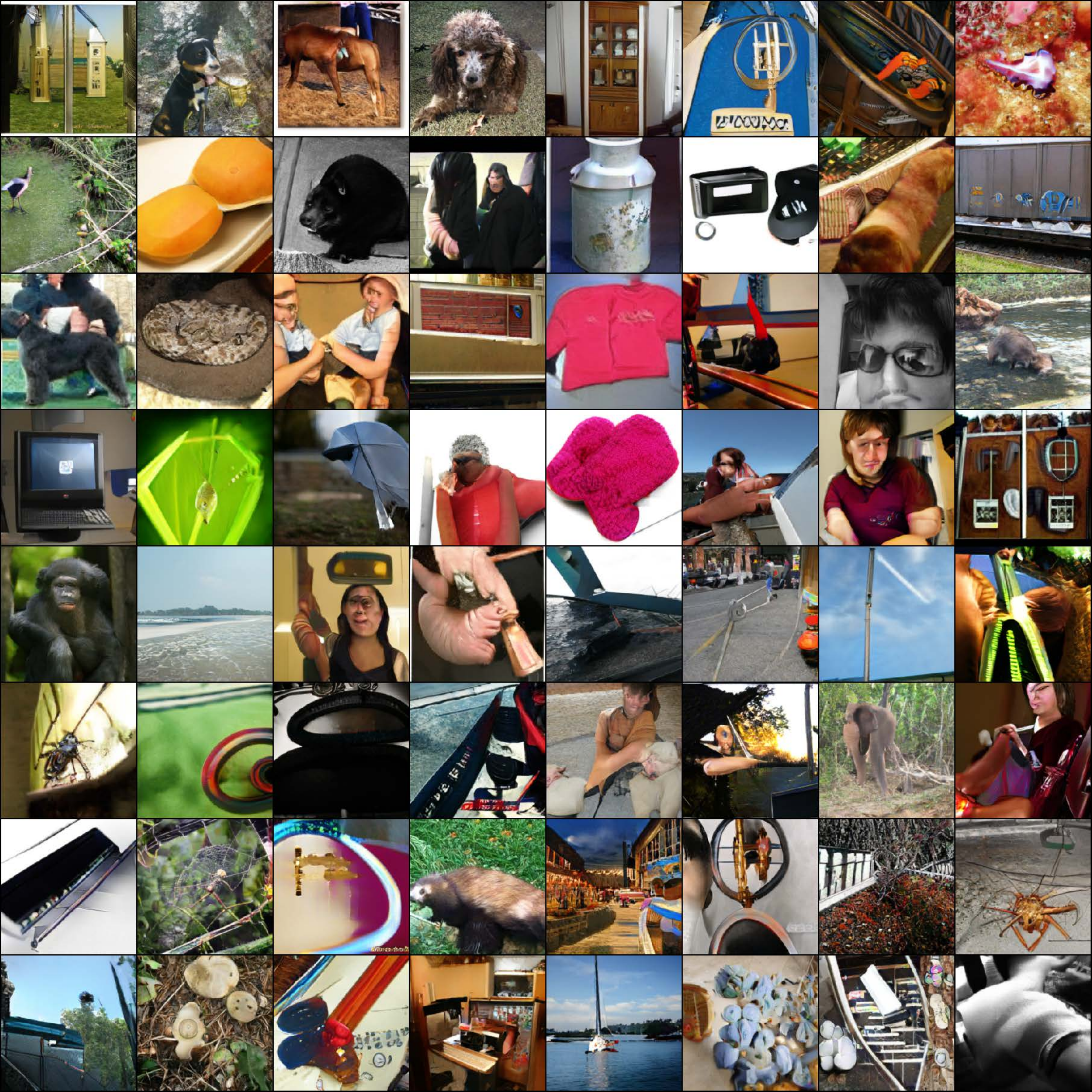}}
    \caption{Mistaken OOD generation on ImageNet dataset. Each figure contains 64 images.}		
    \label{fig:imagenet mistaken}								
\end{figure}

Except for the mistaken OOD generation on CIFAR datasets, we further visualize the mistaken OOD generation data of the advanced data generation-based methods on ImageNet dataset. Since the generators are trained on ID data and these selection procedures can make mistakes, one may wrongly select data with ID semantics as OOD cases (cf., Figure~\ref{fig:imagenet mistaken}).
As we can see, the increased difficulty in searching for the proper OOD-like data in ImageNet leads to more critical mistaken OOD generation.

\section{Broader Impacts and Limitations}
\textbf{Broader impacts. } This paper pioneers work on the problem of mistaken OOD generation in OOD detection, which is significant for the safety-critical applications of models with the rapid development of machine learning. Our method is proposed for the problem, relieving the this problem by a large margin and achieving superior performance. Due to data privacy and security, access of data is often challenging. In our method, we propose an auxiliary OOD detection task built upon the generator with ID knowledge to make the predictor learn to discern ID and OOD cases. Note that, the idea of the auxiliary task can be extended to other domains beyond OOD detection, which may result in the particular applications of other techniques.

\textbf{Limitations. } First, our current realization of ATOL is relatively intricate, requiring training constraints on both the generator and the predictor. Further studies will explore more advanced conditions that can ease our realization and further reduce computing costs. Second, we observe that the diversity of generated data is closely related to the final performance (cf., Appendix~\ref{app: generators}). However, in our current version, we do not consider diversity for the generator in either theories or algorithms, which will motivate our following exploration.

\end{document}